\pdfoutput=1
\documentclass{article}

\usepackage{iclr2027_conference,times}
\iclrfinalcopy   %

\usepackage[utf8]{inputenc}
\usepackage[T1]{fontenc}
\usepackage{amsmath,amssymb,amsfonts}
\usepackage{graphicx}
\usepackage{booktabs}
\usepackage{arydshln}    %
\usepackage{nicematrix}  %
\usepackage{multirow}
\usepackage{wrapfig}     %
\usepackage{needspace}  %
\usepackage{placeins}    %
\usepackage{float}       %
\usepackage{hyperref}
\usepackage{url}
\usepackage{algorithm}
\usepackage{algorithmic}
\usepackage{xcolor}
\usepackage{subcaption}
\usepackage{bbm}
\usepackage{nicefrac}
\usepackage{microtype}
\usepackage{pifont}      %

\newcommand{\amor}{\textsc{Amor}}

\newcommand{\entropy}{\mathcal{H}}
\newcommand{\gate}{g}
\newcommand{\amormamba}{\textsc{Amor}-Mamba2}
\newcommand{\amorgdn}{\textsc{Amor}-Gated DeltaNet}

\newcommand{\amorheart}{\,\textcolor{red}{\scalebox{0.8}{\ding{170}}}}

\usepackage[normalem]{ulem}
\usepackage{framed}
\usepackage{colortbl}
\definecolor{iclrReviewOrange}{HTML}{FFF4E8}
\definecolor{iclrReviewInk}{HTML}{80552D}
\definecolor{iclrChangeOrange}{HTML}{CC7000}
\newcommand{\changed}[1]{{\color{iclrChangeOrange}#1}}

\newenvironment{iclrproposal}[1]{%
  \par\begingroup
  \colorlet{shadecolor}{iclrReviewOrange}%
  \setlength{\FrameSep}{5pt}%
  \begin{shaded*}
  \noindent{\sffamily\scriptsize\color{iclrReviewInk}PROPOSED \textbf{#1}}\par\nobreak\smallskip
  \noindent
}{\end{shaded*}\endgroup\par}

\definecolor{iclrReviewBlue}{HTML}{E8F1FB}
\definecolor{iclrReviewBlueInk}{HTML}{1F4E79}
\definecolor{iclrChangeBlue}{HTML}{1A5FB4}
\newcommand{\changedB}[1]{{\color{iclrChangeBlue}#1}}
\definecolor{iclrChangeLightBlue}{HTML}{4A90C2}
\newcommand{\changedC}[1]{{\color{iclrChangeLightBlue}#1}}

\newenvironment{iclrproposalB}[1]{%
  \par\begingroup
  \colorlet{shadecolor}{iclrReviewBlue}%
  \setlength{\FrameSep}{5pt}%
  \begin{shaded*}
  \noindent{\sffamily\scriptsize\color{iclrReviewBlueInk}PROPOSED \textbf{#1}}\par\nobreak\smallskip
  \noindent
}{\end{shaded*}\endgroup\par}

\renewenvironment{wrapfigure}[2]{%
  \begin{figure}[htbp]\centering\begin{minipage}{#2}%
  \renewcommand{\vspace}[1]{}%
}{\end{minipage}\end{figure}}

\definecolor{S125EditGreen}{HTML}{0057B8}
\definecolor{S125AdditionPurple}{HTML}{7B2CBF}
\DeclareRobustCommand{\reviewedit}[1]{{\color{S125EditGreen}#1}}
\DeclareRobustCommand{\reviewaddition}[1]{{\color{S125AdditionPurple}#1}}
\DeclareRobustCommand{\reviewmoved}[1]{{\color{S125EditGreen}#1}}
\title{When to Think Fast and Slow? AMOR:\\Adaptive Entropy Gate for Hybrid Models}

\author{%
  Haoran Zheng\\
  The University of Chicago\\
  \texttt{haoranzheng@uchicago.edu}
  \And
  Chen Shani\\
  Stanford University\\
  \texttt{cshani@stanford.edu}
}

\begin{document}

\maketitle
\lhead{Preprint.}

\raggedbottom
\setlength{\textfloatsep}{12pt}
\setlength{\intextsep}{12pt}
\setlength{\floatsep}{12pt}

\begin{abstract}

Recurrent-attention hybrids aim to combine the efficiency of recurrence with the contextual recall %
 of attention, but existing approaches typically apply attention uniformly across all positions, even when the recurrent state alone is sufficient for accurate prediction.
We introduce \amor{} (\emph{Adaptive Metacognitive Output Router}), a post-hoc hybrid architecture that selectively invokes attention based on predictive uncertainty. A recurrent backbone is augmented with entropy-gated attention blocks that activate only when the model's output entropy exceeds a dynamic threshold derived from a running batch median and scaled standard deviation. The resulting binary gate requires no learned routing parameters. %
Pretrained from scratch on FineWeb-Edu and with attention invoked on only ${\sim}40\%$ of positions, one of the \amor{} variants (Mamba2 or Gated DeltaNet backbones) achieves the highest eight-task common-sense reasoning average at each scale among pure recurrent, pure attention, and fixed-schedule hybrid models. %
 \amor{} also improves retrieval performance over pure recurrent %
 models while remaining competitive against fixed-schedule hybrids. Additionally, \amor{} retains the long-context robustness of its recurrent backbones, where the Transformer and other hybrid architectures degrade under distribution shift. %
These results suggest that \emph{when} attention is applied matters as much as \emph{how much}: selectively allocating attention based on predictive uncertainty improves accuracy, robustness, and efficiency, offering a simple alternative to uniform or fixed routing strategies.

\end{abstract}

\vspace*{-1em}
\begin{figure}[H]
  \centering
  \includegraphics[width=\textwidth]{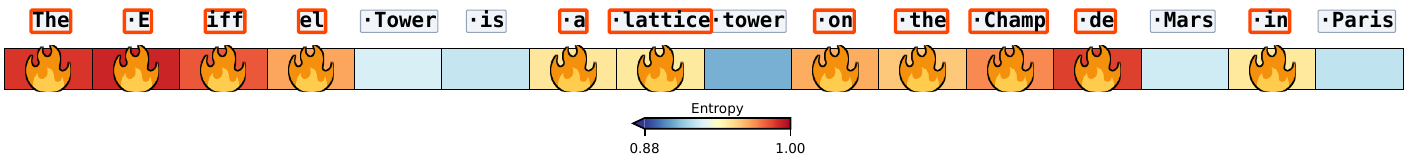}
\caption{\textbf{Entropy gate firing pattern of \amor{}} on the input sentence ``The Eiffel Tower is a lattice tower on the Champ de Mars in Paris.'' The example illustrates how \amor{} combines the strengths of recurrent models and attention: the gate fires early in the sequence, when uncertainty over possible continuations is high, and again on rarer or information-dense tokens such as \emph{lattice} and \emph{Champ}. In contrast, the gate remains inactive on highly predictable tokens that can be inferred from local context or world knowledge (e.g., \emph{Tower}, \emph{Paris}, conditioned on \emph{Eiffel}). This suggests that \textbf{\amor{} selectively deploys attention when next-token prediction becomes difficult}, while relying on the lightweight recurrent backbone for easier continuations.}
  \label{fig:gate_s0_teaser}
\end{figure}

\section{Introduction}
\label{sec:intro}

Since GPT-3~\citep{gpt3}, attention~\citep{transformer} has been the de facto sequence mixer
for language models: every token attends to all preceding tokens, enabling fully parallel training. This uniform pairwise coupling, however, incurs $O(N^2)$ training cost and requires a key-value cache at inference that grows linearly with sequence length, making long-context generation increasingly expensive. To address this, linear recurrent models replace quadratic attention with a fixed-size state: each layer selectively projects the input into a bounded representation, enabling constant memory and $O(1)$ time per token at decode \citep{mamba3, gdn}. However, this efficiency comes with a structural limitation. \textbf{A finite state cannot faithfully store an unbounded set of distinct key-value associations}, making in-context retrieval inherently lossy and leading to degradation on exact-copy stress tests\footnote{Exact-copy stress tests evaluate a model's ability to reproduce a target span verbatim from context, probing lossless storage and retrieval rather than generalization.}~\citep{based, jelassi_repeat_after_me, just_read_twice}.

Hybrid architectures combine the two: attention handles recall while the recurrent backbone keeps long-context decode cheap. Open hybrid models such as Kimi~K3~\citep{kimi_k3} and Nemotron~3~Ultra~\citep{nemotron_3_ultra} are among the strongest open-weight systems (Kimi~K3 is also competitive with closed frontier models on standard benchmarks).
However, these hybrid models still rely on predetermined layer schedules, applying attention uniformly over all tokens, incurring \textbf{full quadratic cost even when the recurrent state alone is sufficient to predict the next token}.

In this work, we aim to \textbf{resolve the tension between computationally heavy attention and compact recurrent models by selectively allocating expensive computation}. We propose \amor{} (Adaptive Metacognitive Output Router; see Fig.~\ref{fig:gate_s0_teaser}), a \emph{post-hoc} hybrid that introduces a lightweight, human-inspired gating mechanism. Motivated by dual-process accounts of cognition~\citep{kahneman_thinking_fast_and_slow}\footnote{{``System 1 operates automatically and quickly, with little or no effort and no sense of voluntary control. System 2 allocates attention to the effortful mental activities that demand it, including complex computations''.}}, where fast automatic responses are complemented by slower deliberation when uncertainty is high, \amor{} preserves a complete recurrent backbone and appends $K$ entropy-gated attention blocks. At each block, \amor{} computes the normalized prediction entropy of each token in the input and activates the block only when entropy exceeds an adaptive threshold. The threshold is defined as an exponential moving average of the batch median with a small standard-deviation-scaled offset, allowing it to track shifts in the backbone’s uncertainty distribution during training. This produces an intuitive and stable routing rule that requires only the model’s own predictive uncertainty and no additional learned routing parameters.

We pretrain eight architectures at three scales (180M, 440M, and 1.5B) under a uniform Chinchilla-optimal token budget. Across scales, \amor{} attends to $\sim40\%$ of positions during training. At inference, firing rates vary by domain but remain stable across context lengths, even beyond those seen in training (Table~\ref{tab:ablation_amor_fire}).

Despite sparse attention, an \amor{} variant achieves the best average common-sense reasoning score at all scales (Fig.~\ref{fig:iclr-common-a}), and outperforms its recurrent backbone on retrieval while remaining competitive with fixed-schedule hybrid baselines (Fig.~\ref{fig:iclr-retrieval-a}). Furthermore, \amor{} preserves its backbone's robustness on long-context tasks, where both Transformer and other hybrids degrade under out-of-distribution vanilla RoPE~\citep{rope} ({Fig.}~\ref{fig:iclr-long-a}). Overall, by routing attention only to uncertain positions, \amor{} augments its recurrent backbone. Our results suggest that efficiency and expressivity are not opposing forces, but a routing problem of \textbf{where and when to allocate compute}.\footnote{Code: \url{https://github.com/HaoranZhengRaul/AMOR}; model-weight links: Appendix~\ref{app:model-weights}.}

\section{Related Work}
\label{sec:related}

\paragraph{Recurrent models.} Linear recurrent models offer efficient alternatives to attention, including S4~\citep{s4}, Mamba~\citep{mamba1}, Mamba2~\citep{mamba2}, Mamba3~\citep{mamba3}, DeltaNet~\citep{deltanet}, Gated DeltaNet~\citep{gdn}, and RWKV~\citep{rwkv}. \amor{} is backbone-agnostic: any recurrent mixer can serve as the unmodified backbone before the entropy gates. We study Mamba2 and Gated DeltaNet.

By compressing past context into a fixed-size state, recurrent models support linear-time training and $O(1)$ updates per decoded token~\citep{mamba2,gdn}, versus $O(N)$ for full attention over $N$ tokens. However, this fixed capacity limits how much information they can retain for later retrieval.
\citet{jelassi_repeat_after_me} show that recurrent models with fixed-size states fail to copy sequences whose information content exceeds their state size, and that pretrained Mamba underperforms Pythia Transformers on synthetic recall tasks such as copying and phone-book lookup. Consistent with this, \citet{based} report a substantial recall-accuracy gap between attention-based models and Mamba on real-world, recall-intensive benchmarks, and \citet{just_read_twice} extend these findings to large-scale retrieval settings, where Transformers retain a clear advantage even at a billion-parameter scale.
Together, these results suggest that \textbf{while recurrent models offer compelling efficiency gains, they struggle to match the flexible, content-addressable memory afforded by attention.}

\paragraph{Recurrent-attention hybrids.} To address the limitations of attention and recurrent models, recent work has explored hybrid architectures, which we divide into \emph{serial} and \emph{fused} hybrids.

Serial hybrids replace the recurrent mixer with attention at a \textit{fixed subset} of layers. Mamba-based hybrids typically use sparse schedules: Jamba~\citep{jamba} uses a 1:7 attention-to-Mamba ratio, Bamba~\citep{bamba} $\sim$1:10, and Nemotron~3~Super~\citep{nemotron_3_super} 1:5. Delta-rule-based hybrids tend toward denser schedules, with Qwen3.5~\citep{qwen3_5}, Kimi Linear~\citep{kimi_linear}, and Olmo Hybrid~\citep{olmo_hybrid} near 1:3. Other examples include Samba~\citep{samba}, Griffin~\citep{griffin}, Zamba2~\citep{zamba2}, Granite\,4.0~\citep{granite4}, and Jet-Nemotron~\citep{jet_nemotron}. Fused hybrids instead run attention \emph{in parallel} with the recurrent mixer: Hymba~\citep{hymba} averages attention and Mamba heads with learned per-channel scales, while Falcon-H1~\citep{falcon_h1} concatenates them at a fixed channel ratio.

Prior hybrids thus focus primarily on \emph{depth}: determining \textbf{where} and in what proportion to place attention. \citet{bae_placement} find that attention distributed throughout the network improves fused hybrids, while serial hybrids favor an attention-to-Mamba ratio of $\sim1{:}5$. Similarly, \citet{wang_survey} find that full-attention-to-linear ratios of $1{:}3$--$1{:}6$ recover Transformer-level recall.

In both serial and fused hybrids, attention placement is fixed at design time. \amor{} instead acts on \textbf{when} attention fires, not just \textbf{where} or \textbf{how much}. It appends \(K\) post-hoc attention blocks after the recurrent backbone and selectively activates them at each token via normalized-entropy gating.

\paragraph{Conditional Compute.} Soft methods such as ACT~\citep{act} and PonderNet~\citep{pondernet} use differentiable halting, while hard-routing methods make discrete compute decisions. Mixture-of-Depths (MoD)~\citep{mod} routes a fixed token budget through attention and MLP blocks, while DeepSeek Sparse Attention~\citep{deepseek_v32}, Native Sparse Attention~\citep{nsa}, and MoBA~\citep{moba} prune attention via top-$k$ routing.

Most conditional-compute methods route based on \emph{input} representations, such as MoD's router signals or MoBA's query-key similarity. A smaller line of work conditions on the model's \emph{output}; CALM~\citep{calm}, for example, uses the top-1 vs.\ top-2 probability margin for layer-wise early exit. \amor{} instead uses the full next-token distribution's entropy for per-position gating \emph{within} attention sublayers, selectively invoking attention rather than skipping layers.

\section{AMOR}
\label{sec:method}

We now describe \amor{} in detail, starting with the general architecture (Section~\ref{sec:method:arch}), and focusing on our entropy gate (Section~\ref{sec:method:gate}) and each block's flow (Section~\ref{sec:method:fwd}) during training (Section~\ref{sec:method:train}) and inference (Section~\ref{sec:method:decode}). An optional distilled approximation to the entropy gate is evaluated separately as a deployment ablation in Section~\ref{sec:exp:inference}; its definition is in Appendix~\ref{app:iclr-router}.

\subsection{Architecture}
\label{sec:method:arch}

We instantiate \amor{} within a standard recurrent-attention hybrid skeleton:
\begin{equation}
\resizebox{0.8\textwidth}{!}{$\displaystyle
  \mathrm{Embed} \to
  \big[\mathrm{Recurrent\,Mixer}_{i} + \mathrm{SwiGLU\text{-}MLP}_{i}\big]_{i=1}^{N}
  \to
  \big[\mathrm{AMOR\,Block}_{\ell}\big]_{\ell=0}^{K-1}
  \to \mathrm{final\,norm} \to \mathrm{LM\,Head}.
$}
\label{eq:skeleton}
\end{equation}
{Here $N$ is the backbone depth, counting recurrent-mixer--MLP layers indexed by $i$. We index AMOR blocks by $\ell$, use $K$ for their total count, and reserve $\kappa$ for the entropy-threshold offset multiplier.}

\begin{figure}[t]
  \centering
  \includegraphics[width=0.9\textwidth]{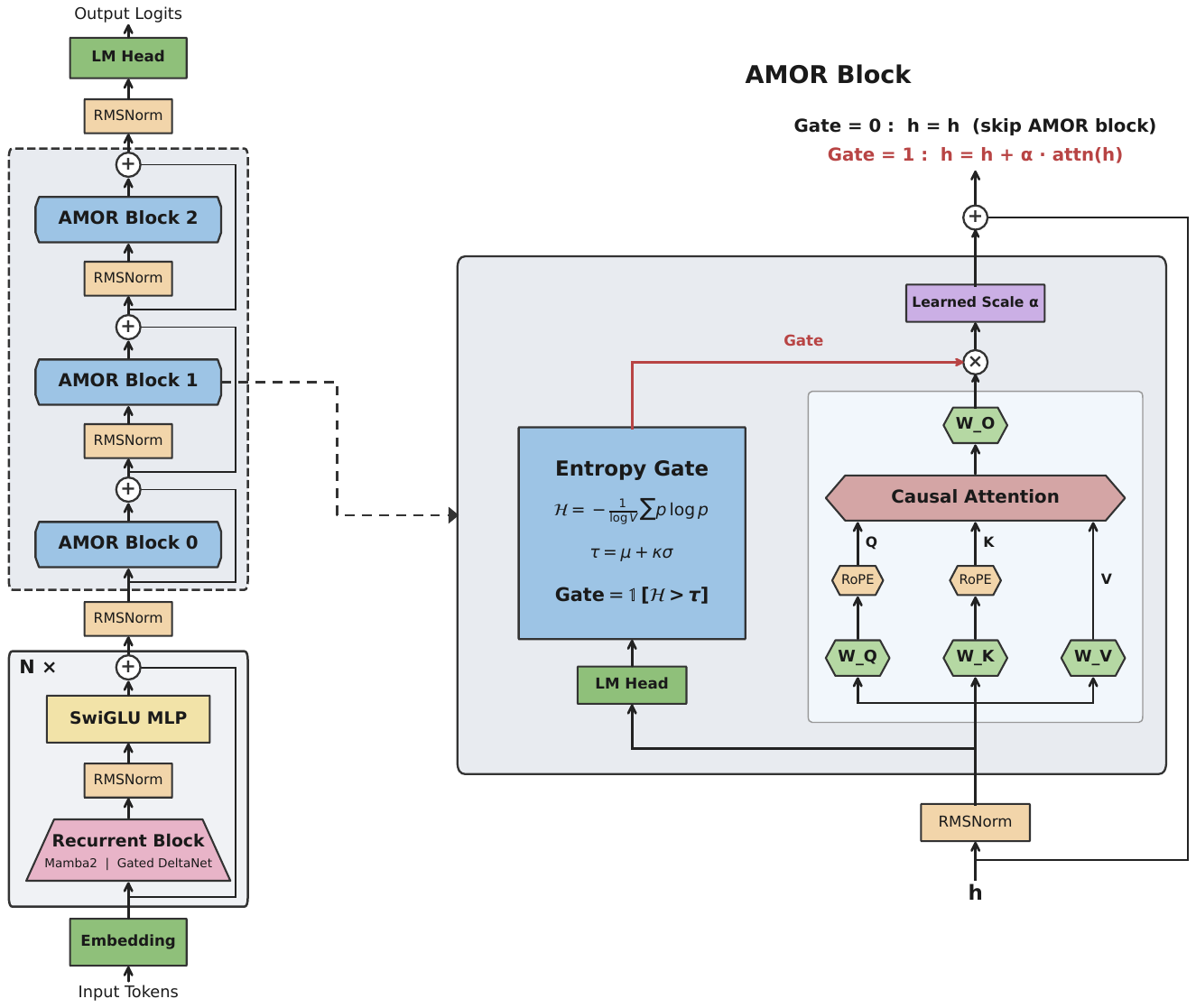}
\caption{\textbf{\amor{} architecture.} A recurrent backbone (Mamba2 or Gated DeltaNet, with SwiGLU MLPs) feeds three \amor{} blocks. Each block queries the tied LM head, gates causal RoPE attention via normalized entropy, and adds a per-channel $\boldsymbol{\alpha}$-scaled residual update. Block~0 consumes the backbone's terminal $\mathrm{norm}_f$ output (no pre-norm); Blocks~1--2 use RMSNorm. $\mathbf{W}_{\!O}$ is zero-initialized, so \amor{} matches the backbone at initialization.}
\label{fig:architecture}
\end{figure}

We study two variants that differ only in the recurrent mixer: \amormamba{} (based on Mamba2) and \amorgdn{} (based on Gated DeltaNet). Both insert $K{=}3$ entropy-gated \amor{} blocks between the backbone and the final LM head (Fig.~\ref{fig:architecture}).\footnote{$K{=}3$ matches the attention budget of our serial and fused hybrid baselines: three attention layers in a 24-layer model, following Jamba's~\citep{jamba} recommended 1:7 ratio.}

With entropy gating, each model reuses a single LM head, weight-tied to the embedding, invoked $K{+}1$ times per forward pass: once on each \amor{} block's normalized input to compute its gate (Section~\ref{sec:method:gate}), and once on the final normalized residual to produce output logits.

Within each \amor{} block, attention is implemented as causal SDPA over $\mathbf{Q}, \mathbf{K}, \mathbf{V}$ projections of the normalized input, with RoPE~\citep{rope} applied to $\mathbf{Q}$ and $\mathbf{K}$.  For model dimension $D{}=d_{\text{model}}$, each block introduces $\sim\!4D^2$ parameters ($\mathbf{W}_{\!Q}, \mathbf{W}_{\!K}, \mathbf{W}_{\!V}, \mathbf{W}_{\!O}$ plus a per-channel scaling vector $\boldsymbol{\alpha}$).

A key design choice is that the three \amor{} blocks are intentionally \textit{asymmetric}. Block~0 operates directly on the backbone’s final $\mathrm{norm}_f$-normalized residual (the LM-ready representation) and applies attention refinement on this post-$\mathrm{norm}_f$ stream. In contrast, Blocks~1 and~2 follow a standard pre-norm formulation: each first applies its own RMSNorm~\citep{rmsnorm} before computing the entropy gate and attention, then updates the (unnormalized) residual. This distinction is important because the residual is no longer unit-RMS scaled after Block~0's intervention. This asymmetry enforces a key invariant: the recurrent backbone remains a complete, standalone LM, rather than the first stage of a deeper stack. \amor{} thus acts strictly as a conditional refinement layer on top of an already valid LM representation, aligning with the dual-process framing in Section~\ref{sec:intro}. We also considered a symmetric alternative in which $\mathrm{norm}_f$ is repurposed as Block~0's pre-norm and the residual remains unnormalized throughout the \amor{} stack (Appendix~\ref{app:ablations}).

Lastly, each \amor{} block adds a per-channel $\boldsymbol{\alpha}$-scaled attention update to the residual. The per-channel scaling vector $\boldsymbol{\alpha} \in (0,1)^D$ allows each dimension to control its own contribution, rather than relying on a single global factor. After accumulating $K$ such updates, a single $\mathrm{final\,norm}$ rescales the stream before the LM head.
To preserve the backbone's initial behavior, $\mathbf{W}_{\!O}$ is zero-initialized, making \amor{} functionally identical to the recurrent backbone at initialization; deviations are learned progressively as $\mathbf{W}_{\!O}$ updates. Finally, \amor{} blocks contain no MLP (Appendix~\ref{app:ablations}), isolating the effect of attention-based refinement.

\subsection{Entropy gate}
\label{sec:method:gate}

Each \amor{} block decides where to engage attention based on the backbone's predictive uncertainty. At token position $t$, given the block's normalized input $\tilde{h}_t$, the shared LM head produces logits $\mathbf{z}_t = \mathrm{lm\_head}(\tilde{h}_t) \in \mathbb{R}^V$ over the $V{=}128{,}256$-token Llama-3.1 vocabulary. From these logits, we compute the normalized output entropy:
\begin{equation}
  \entropy_t \;=\; -\frac{1}{\log V}\sum_{v=1}^V p_{t,v}\,\log p_{t,v}\;\in [0,1],
  \qquad \mathbf{p}_t = \mathrm{softmax}(\mathbf{z}_t).
  \label{eq:entropy}
\end{equation}
{Here $p_{t,v}$ is the predicted probability of vocabulary token $v$ at position $t$.}

Logits are detached before entropy computation: $\entropy_t$ drives the gating decision and does not backpropagate into the backbone, keeping the gate purely diagnostic.
At each training {forward pass}, the block updates two non-learnable EMA buffers from its entropies $\entropy^{\text{batch}}$ over all positions in the batch,
\begin{equation}
  \mu \,\leftarrow\, (1{-}\eta)\,\mu \,+\, \eta\,\mathrm{median}(\entropy^{\text{batch}}),
  \qquad
  \sigma \,\leftarrow\, (1{-}\eta)\,\sigma \,+\, \eta\,\mathrm{std}(\entropy^{\text{batch}}),
  \label{eq:ema}
\end{equation}
with $\eta{=}0.01$ (momentum $0.99$), and forms the threshold
\begin{equation}
  \tau \;=\; \mu \,+\, \kappa\,\sigma, \qquad \kappa = 0.2.
  \label{eq:threshold}
\end{equation}
The gate is binary: $\gate_t = \mathbf{1}[\entropy_t > \tau] \in \{0,1\}$. The statistics $\mu$ and $\sigma$ are maintained as non-learnable EMA buffers and frozen at inference; accordingly, the threshold $\tau$ adapts to the entropy distribution during training and remains fixed at evaluation.
We set $\tau = \mu + \kappa \sigma$, where using the median-based $\mu$ provides robustness to per-batch outliers, and scaling by $\sigma$ stabilizes the firing rate across model sizes. In contrast, a fixed offset leads to drift, as $\sigma$ typically shrinks with improved calibration.
In practice, we target a per-block firing rate of $\sim\!40\%$. A single choice of $\kappa{=}0.2$ achieves this consistently across three model scales and two backbones (Mamba2 and Gated DeltaNet), with firing rates remaining stable throughout training (Appendix~\ref{app:gate-stability}).

\subsection{Block forward and gradient flow}
\label{sec:method:fwd}

Within block $\ell$, let $h_t^{(\ell)}$ be the input residual and $\tilde{h}_t^{(\ell)}$ its normalized copy used for gating and attention. We set $\tilde{h}_t^{(0)} = h_t^{(0)}$ (the backbone’s $\mathrm{norm}_f$ output), while for $\ell{\geq}1$, $\tilde{h}_t^{(\ell)} = \mathrm{RMSNorm}_\ell\!\left(h_t^{(\ell)}\right)$, with $h_t^{(\ell)}$ the unnormalized residual from block $\ell{-}1$. Here $\tilde h$ denotes the whole normalized input sequence. Training uses dense attention with a post-mask ($\odot$ denotes elementwise multiplication):
\begin{align}
  \mathbf{Q},\mathbf{K},\mathbf{V}
    &\,=\, \mathbf{W}_{\!Q}\,\tilde{h}, \;\; \mathbf{W}_{\!K}\,\tilde{h}, \;\; \mathbf{W}_{\!V}\,\tilde{h}, \\
  (\mathbf{Q},\mathbf{K}) &\,\leftarrow\, \mathrm{RoPE}(\mathbf{Q},\mathbf{K}), \\
  \mathbf{a}_t
    &\,=\, \mathbf{W}_{\!O}\, \mathrm{SDPA}(\mathbf{Q}, \mathbf{K}, \mathbf{V})_t, \\
  h^{(\ell+1)}_t
    &\,=\, h^{(\ell)}_t \;+\; \gate_t \cdot \boldsymbol{\alpha} \odot \mathbf{a}_t.
  \label{eq:residual}
\end{align}
We parameterize the per-channel scaling vector as $\boldsymbol{\alpha} = \mathrm{sigmoid}(\tilde{\boldsymbol{\alpha}})$ with $\tilde{\boldsymbol{\alpha}}_{\text{init}}{=}\mathbf{0}$, yielding $\boldsymbol{\alpha}_{\text{init}} = 0.5$ in every channel.\footnote{As a pretraining safeguard, we clamp pre-sigmoid parameters at $-1$ ($\boldsymbol{\alpha}\ge\mathrm{sigmoid}(-1)\approx0.269$). Across inspected final checkpoints, all channels in Blocks~0--1 remained above the floor; only 5/2{,}048 channels (0.2\%) in 1.5B \amorgdn{}'s Block~2 were at it.} The gate is binary and non-differentiable ($\partial \gate_t / \partial \entropy_t = 0$ almost everywhere), ensuring that no gradients flow through the threshold or back into the backbone via $\entropy$.

Gradient flow is position-dependent (Fig.~\ref{fig:decode}). At firing positions ($\gate_t = 1$), all parameters $\mathbf{W}_{\!Q}, \mathbf{W}_{\!K}, \mathbf{W}_{\!V}, \mathbf{W}_{\!O}, \boldsymbol{\alpha}$ receive gradients as in standard attention. At non-firing positions ($\gate_t = 0$), gradients to $\mathbf{W}_{\!Q}, \mathbf{W}_{\!O}$, and $\boldsymbol{\alpha}$ vanish at $t$, while $\mathbf{W}_{\!K}$ and $\mathbf{W}_{\!V}$ still receive gradients via causal attention from later firing positions $j > t$. This asymmetry is intentional: $\mathbf{Q}$ is only needed where the gate fires, whereas $\mathbf{K}$ and $\mathbf{V}$ must represent all positions for future retrieval.

\subsection{Training mode}
\label{sec:method:train}

We train in a dense $\mathrm{full\_with\_mask}$ mode (causal SDPA at all positions, masked by the gate), which is mathematically equivalent to the $\mathrm{true\_sparse}$ inference implementation; we use the dense form for efficiency via fused FlashAttention-2 kernels~\citep{flashattention2} (Appendix~\ref{app:method:train}).

\Needspace{5\baselineskip}
\subsection{Inference: conditional skip and KV continuity}
\label{sec:method:decode}

\begin{iclrwrapfigure}[32]{R}{0.50\textwidth}
\centering
\includegraphics[width=\linewidth]{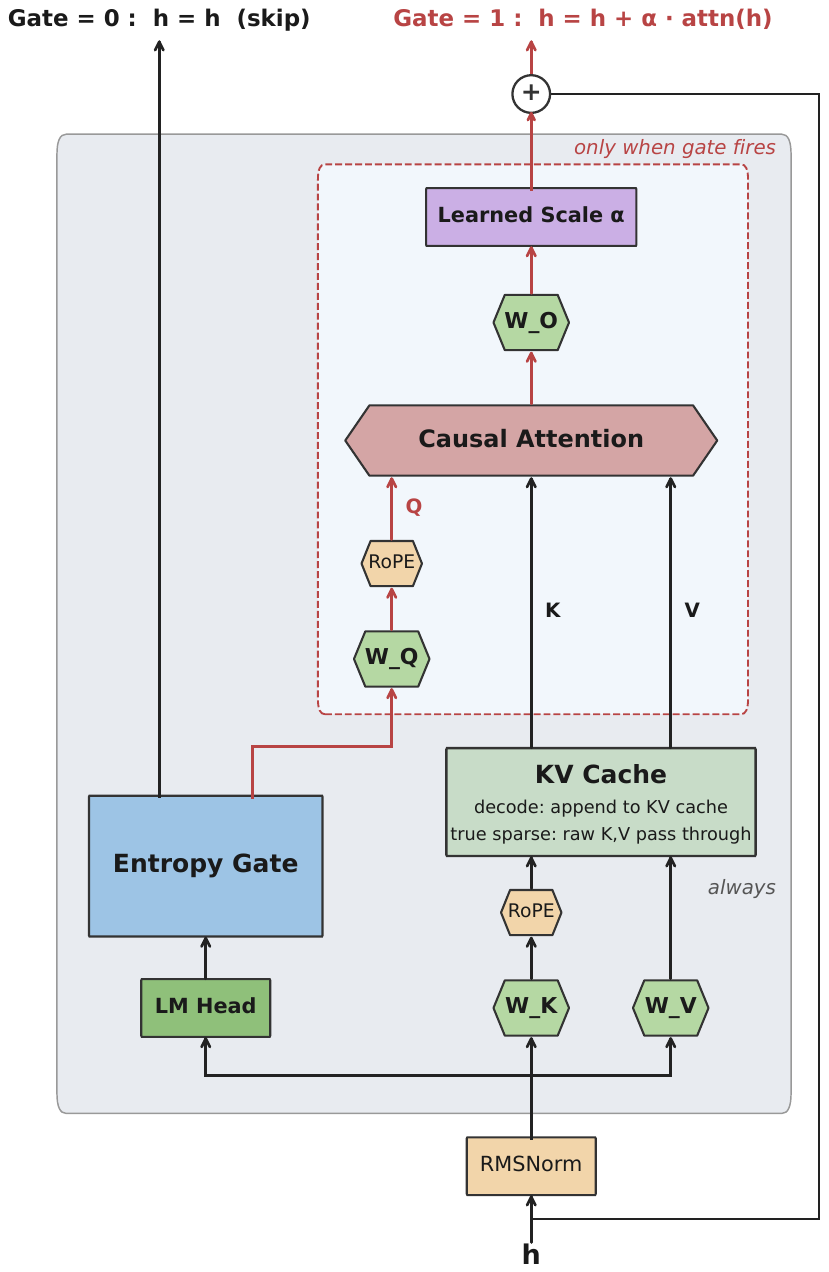}
\caption{{\textbf{\amor{} decode path.} When $\gate_t{=}0$, the block skips $\mathbf{W}_{\!Q}$, RoPE on $\mathbf{Q}$, SDPA, and $\mathbf{W}_{\!O}$. $\mathbf{W}_{\!K}$, $\mathbf{W}_{\!V}$, RoPE on $\mathbf{K}$, and KV-cache updates always run.}}
\label{fig:decode}
\end{iclrwrapfigure}

At decode time, the gate is evaluated per token and induces a conditional skip: when $\gate_t = 0$, the $\mathbf{W}_{\!Q}$ projection, RoPE on $\mathbf{Q}$, SDPA, and $\mathbf{W}_{\!O}$ are bypassed. In all cases, $\mathbf{W}_{\!K}, \mathbf{W}_{\!V}$, RoPE on $\mathbf{K}$, and KV cache updates are performed, ensuring that future firing queries can attend to all past positions, including non-firing ones.

With the threshold frozen, \amor{} yields \emph{difficulty-adaptive compute}: \textbf{attention is invoked more frequently when the backbone is uncertain and less when it is confident}.

\leavevmode For the native entropy gate, counting attention matmul savings only, gating reduces matrix-multiply FLOPs relative to otherwise identical always-on attention when

\begin{equation}
  (1 - f) \cdot C_{\text{attn}}(L) \;>\; C_{\text{lm\_head}},
  \label{eq:decode_speedup}
\end{equation}

\noindent where $f$= firing rate and $L$=attended cache length including the current token. For one decoded token in one AMOR block, \(C_{\text{attn}}(L)=4DL\) and \(C_{\text{lm\_head}}=2VD\). This is an arithmetic condition, not a latency guarantee, as memory traffic, kernel execution, and synchronization also affect runtime. Appendix~\ref{app:bench:breakeven} derives the entropy gate's arithmetic threshold and compares routing costs; Section~\ref{sec:exp:inference} reports measured latency and memory at 1.5B.

\section{Experiments}
\label{sec:experiments}

We compare \amor{} with pure recurrent, pure attention, and fixed-schedule hybrid baselines to assess whether selective attention allocation improves downstream performance. The eight architectures all use sequence mixers and SwiGLU MLPs, with differences in mixer type and attention placement.

\paragraph{Baselines.} \textbf{Backbone models} include pure recurrent mixers (Mamba2~\citep{mamba2}, Gated DeltaNet~\citep{gdn}) and a pure attention baseline (a Llama-style Transformer~\citep{llama3} with full RoPE~\citep{rope}). \textbf{Fixed-schedule hybrids} combine recurrence with attention at predetermined layers. We evaluate two serial hybrids (Mamba2, Gated DeltaNet), which replace recurrent blocks with attention at layers $\{4,8\}$ for $N{=}12$ and $\{6,12,18\}$ for $N{=}24$, following Jamba~\citep{jamba}. We also include a fused hybrid (Mamba2), which runs recurrent and attention modules in parallel at layers $\{0,6,11\}$ for $N{=}12$ and $\{0,12,23\}$ for $N{=}24$, following Hymba~\citep{hymba}.
We omit a fused Gated DeltaNet variant due to the lack of an open-weight implementation. \textbf{Post-hoc hybrids} correspond to \amor{} (\amormamba{}, \amorgdn{}), which append $K{=}3$ entropy-gated attention blocks to an otherwise unmodified backbone (Fig.~\ref{fig:architecture}).

\paragraph{Controlled Design.} All models are trained from scratch on FineWeb-Edu~\citep{fineweb_edu} with the Llama-3.1 tokenizer~\citep{llama3} at sequence length $T{=}3{,}072$, using an identical optimization recipe (Appendix~\ref{app:training}). At each scale, all models are trained on the same Chinchilla-optimal token budget~\citep{chinchilla}: $3.69$B / $8.97$B / $30.7$B tokens at 180M / 440M / 1.5B parameters. We follow the high-level designs of Jamba and Hymba while removing orthogonal components to isolate the effect of \emph{fixed-schedule attention}. All hybrids use full causal attention and the same corresponding Mamba2 or Gated DeltaNet backbone, ensuring consistent attention and recurrent primitives across models (see full details in Appendix~\ref{app:arch}).

\paragraph{Tasks.} Following the evaluation protocol of \citet{gdn}, we evaluate on: (a)~common-sense reasoning at 180M, 440M, and 1.5B (Section~\ref{sec:exp:common}); (b)~in-context retrieval at 1.5B (Section~\ref{sec:exp:retrieval}); (c)~long-context behavior at 1.5B (Section~\ref{sec:exp:long}). This follows the largest-scale evaluation design of Gated DeltaNet (1.3B) and Mamba3 (1.5B)~\citep{gdn,mamba3}; Appendix~\ref{app:iclr-router} additionally evaluates the smaller-scale deployment ablation. We use the LM Evaluation Harness~\citep{lm_eval_harness}; task details are in Appendix~\ref{app:eval}, with per-task results in Tables~\ref{tab:common_sense}--\ref{tab:longbench}.

\subsection{Common-Sense Reasoning}
\label{sec:exp:common}

\begin{figure}[htbp]\centering
\includegraphics[width=\linewidth]{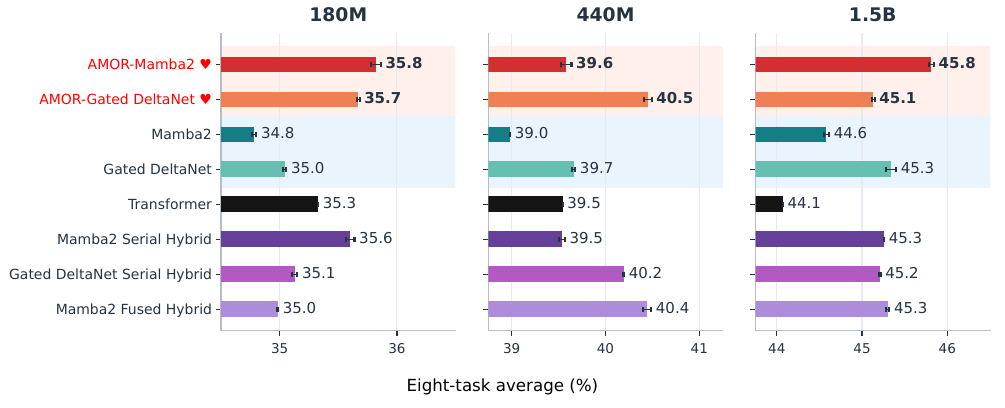}
\caption{{\textbf{An \amor{} variant outperforms common-sense reasoning on every scale.} Eight-task common-sense reasoning accuracy (detailed scores in Table~\ref{tab:common_sense}).}}\label{fig:iclr-common-a}
\end{figure}

We evaluate zero-shot accuracy on LAMBADA~\citep{lambada}, HellaSwag~\citep{hellaswag}, PIQA~\citep{piqa}, ARC-Easy, ARC-Challenge~\citep{arc}, WinoGrande~\citep{winogrande}, OpenBookQA~\citep{openbookqa} and TruthfulQA-mc2~\citep{truthfulqa}. \textbf{An \amor{} variant outperforms common-sense reasoning on every scale} (Fig.~\ref{fig:iclr-common-a}; Table~\ref{tab:common_sense}). Error bars are constructed from five evaluations of the final model checkpoint and show the sample SD. Results across seeds controlling model initialization and training-data order at 180M are reported in Appendix~\ref{app:iclr-seeds}.

\subsection{In-Context Retrieval}
\label{sec:exp:retrieval}

Retrieval is where attention is least substitutable: cloze tasks require recovering spans from earlier in the input. Pure recurrent backbones underperform sharply, while fixed-schedule hybrids lead. \textbf{Both \amor{} variants substantially improve over their backbones and remain competitive with fixed-schedule hybrids}, with gains in 11 of 12 cloze retrieval task-backbone comparisons. On Single Needle-in-a-Haystack (S-NIAH), \amor{}'s advantage is especially evident at 2K context (Fig.~\ref{fig:iclr-retrieval-a}). Task details and full results are in Appendix~\ref{app:eval:retrieval} and Table~\ref{tab:retrieval}.

\amor{} achieves these gains with attention restricted to uncertain positions. Zeroing attention outputs reduces the average over SWDE~\citep{swde}, SQuAD-completion~\citep{squad}, and FDA~\citep{fda} from 30.4\% to 9.0\% for \amormamba{} and 36.4\% to 11.4\% for \amorgdn{} (Table~\ref{tab:iclr-routing-controls}), demonstrating the contribution of its gated attention updates. The Transformer's steep S-NIAH decline at 4K occurs beyond its 3{,}072-token training context, consistent with vanilla RoPE's poor length extrapolation~\citep{yarn}; see Section~\ref{sec:exp:long}.

\begin{figure}[htbp]\centering
\includegraphics[width=\linewidth,trim=0 18 0 0,clip]{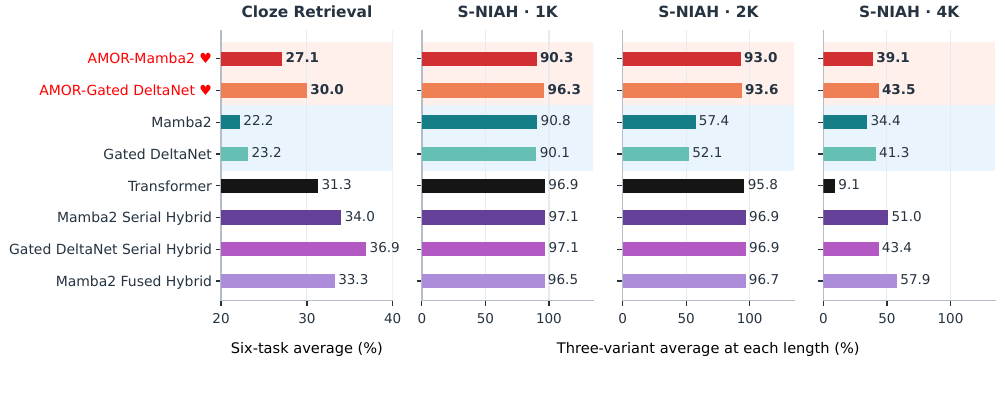}
\caption{\textbf{\amor{} improves in-context retrieval over recurrent backbones and remains competitive with fixed-schedule hybrids.} Left: six-task cloze retrieval average. Right: average across three Single Needle-in-a-Haystack (S-NIAH) variants at varying context lengths. 1.5B; full task scores are in Table~\ref{tab:retrieval}.}\label{fig:iclr-retrieval-a}\label{fig:iclr-niah}
\end{figure}

\subsection{Long-Context Behavior}
\label{sec:exp:long}

LongBench evaluates long-context understanding beyond training. Pure recurrent backbones, which do not explicitly encode position, degrade gracefully, while the pure attention model and fixed-schedule hybrids, especially serial, use full RoPE attention without length extension and fall below recurrent baselines. This pattern is reaffirmed on length extrapolation tasks (Fig.~\ref{fig:length_extrap}).

\begin{figure}[htbp]
\centering
\includegraphics[width=\linewidth]{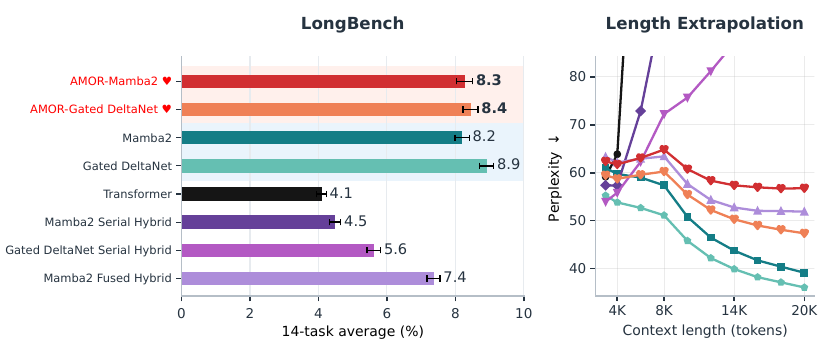}
\caption{{\textbf{\amor{} retains the long-context performance of recurrent backbones while other attention variants degrade.} Left: 14-task LongBench average. Right: NarrativeQA length extrapolation.}}
\label{fig:iclr-long-a}
\label{fig:length_extrap_narrativeqa}
\end{figure}

\textbf{\amor{} stays within 0.5 points of its recurrent backbone on LongBench and outperforms the Transformer and fixed-schedule hybrids} (Fig.~\ref{fig:iclr-long-a}, left; Table~\ref{tab:longbench}). On NarrativeQA, \amor{} has lower perplexity at 20K than at the 3K training context, while the Transformer and serial hybrids deteriorate sharply (Fig.~\ref{fig:length_extrap_narrativeqa}, right).

Our intuition is that \amor{} mitigates RoPE distribution shift via its entropy gate: when quiet, the gate skips attention, predicting via the RoPE-free recurrent backbone. This is consistent with stable long-context firing rates within domains (Table~\ref{tab:ablation_amor_fire}), though causal conclusions require future work.

\section{\texorpdfstring{Deployment Ablation: Inference Efficiency}{Deployment Ablation: Inference Efficiency}}
\label{sec:exp:inference}

To further reduce routing overhead, we optionally distill each block's LM-head entropy estimates into a predictor with a 512-unit hidden layer. The predictors are fitted on the frozen pretrained checkpoint, keeping the pretrained model and gate thresholds fixed. Together, they form a distilled router that adds 3.15M parameters at 1.5B (approximately 0.21\%) and replaces the three extra LM-head evaluations (Appendix~\ref{app:iclr-router}). This provides a deployment ablation of the native entropy gate: the pretrained model weights and frozen entropy thresholds are held fixed, while only the mechanism used to estimate routing entropy is changed. Sections~\ref{sec:exp:common}--\ref{sec:exp:long} evaluate the native entropy gate; here we evaluate this distilled router.

\begin{figure}[htbp]\centering
\includegraphics[width=\linewidth]{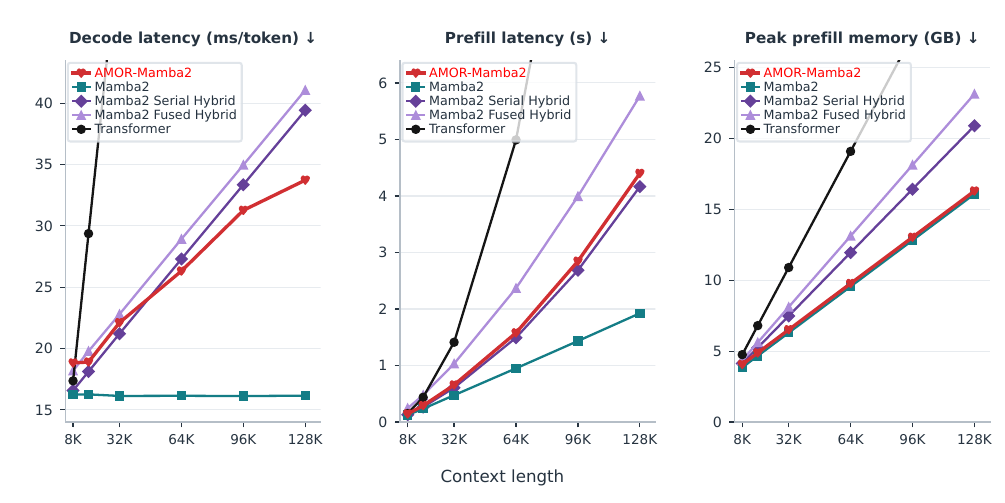}
\caption{\textbf{\amor{} exhibits better inference scaling and lower peak prefill memory than its hybrid counterparts.} Left: decode latency. Middle: prefill latency. Right: peak prefill memory. 1.5B; one H100 PCIe, batch 1, memory-efficient SDPA and FineWeb-Edu prompts from 8K to 128K. \amor{} uses the distilled router with natural firing. Transformer curves exceed the displayed ranges, and it runs out of memory at 128K. All eight models: Fig.~\ref{fig:iclr-eff-all}; values: Table~\ref{tab:iclr-runtime-values}.}\label{fig:iclr-eff-a}
\end{figure}

\textbf{\amormamba{} overtakes the fused hybrid in decode between 8K and 16K and the serial hybrid between 32K and 64K} (Fig.~\ref{fig:iclr-eff-a}, left). \textbf{Its prefill latency is 23.7--40.8\% lower than the fused hybrid and tracks closely with the serial hybrid} (middle). \textbf{Peak prefill memory remains almost identical to its recurrent backbone's, reaching 29.6\% below the fused hybrid and 22.0\% below the serial hybrid at 128K} (right). Compared with the native entropy gate, the distilled router lowers peak prefill memory by 2.3--9.2\,GB and decode latency by 0.4--1.1\,ms/token in the matched 1.5B cohort (Table~\ref{tab:iclr-native-distilled}). The post-hoc design runs the recurrent backbone before building attention activations and KV caches, keeping these allocations out of the backbone's memory peak and elegantly redirecting the traffic. At 1.5B, common-sense, retrieval, S-NIAH and LongBench averages change by at most 0.13 points (Table~\ref{tab:iclr-router-quality}); full task results and length extrapolation comparisons appear in Appendix~\ref{app:iclr-router}. The full benchmarking protocol and memory analysis are in Appendix~\ref{app:iclr-efficiency}; training and 3K decode throughput appear in Tables~\ref{tab:bench_summary} and~\ref{tab:decode_3k}. Appendix~\ref{app:iclr-efficiency} reports measurements at all scales: prefill latency and memory improve consistently, while decode gains vary by backbone and context length.

\section{Conclusion \& Future Work}
\label{sec:conclusion}

{\amor{} is a novel hybrid architecture that allocates attention based on the model's predictive uncertainty.} Our results suggest that \emph{when} attention is applied matters as much as \emph{how much}: selectively engaging attention based on the model's own uncertainty can improve efficiency and robustness relative to fixed schedules. More broadly, \amor{} demonstrates that simple, output-driven mechanisms can effectively coordinate hybrid architectures. Using an entropy gating rule during pretraining and an optional distilled router to reduce deployment overhead, \amor{} points toward a class of \textbf{models that adapt their computation to the demands of the model.} Future work could learn or calibrate the routing policy itself, potentially adapting the allocation of computation across scales, architectures, and tasks (see limitations in Appendix~\ref{app:limitations}).

\clearpage
\paragraph{Ethics Statement.} This work studies language-model architectures by pretraining them on existing datasets; the resulting models may inherit biases from their training data.

\paragraph{Reproducibility Statement.} The training recipe is provided in Table~\ref{tab:training_recipe}, with architecture, evaluation, baselines, ablations and runtime protocols detailed in the corresponding sections and appendices. The code is provided through a link in the introduction.

\paragraph{AI Use Statement.} Generative AI tools assisted with coding, literature retrieval, and manuscript editing and review. The authors reviewed these contributions and checked all claims against references and executed experimental results. The authors are responsible for the research design and final presentation.

\begingroup
\color{black}
\renewcommand{\changed}[1]{#1}
\renewcommand{\changedB}[1]{#1}
\renewcommand{\changedC}[1]{#1}
\renewcommand{\reviewedit}[1]{#1}
\renewcommand{\reviewaddition}[1]{#1}
\renewcommand{\reviewmoved}[1]{#1}
\hypersetup{hidelinks}
\bibliographystyle{iclr2027_conference}
\bibliography{references/references}

\begin{thebibliography}{76}
\providecommand{\natexlab}[1]{#1}
\providecommand{\url}[1]{\texttt{#1}}
\expandafter\ifx\csname urlstyle\endcsname\relax
  \providecommand{\doi}[1]{doi: #1}\else
  \providecommand{\doi}{doi: \begingroup \urlstyle{rm}\Url}\fi

\bibitem[Arora et~al.(2023)Arora, Yang, Eyuboglu, Narayan, Hojel, Trummer, and Ré]{fda}
Simran Arora, Brandon Yang, Sabri Eyuboglu, Avanika Narayan, Andrew Hojel, Immanuel Trummer, and Christopher Ré.
\newblock Language models enable simple systems for generating structured views of heterogeneous data lakes.
\newblock \emph{Proceedings of the VLDB Endowment}, 17\penalty0 (2):\penalty0 92--105, 2023.
\newblock URL \url{https://arxiv.org/abs/2304.09433}.

\bibitem[Arora et~al.(2024{\natexlab{a}})Arora, Eyuboglu, Zhang, Timalsina, Alberti, Zinsley, Zou, Rudra, and Ré]{based}
Simran Arora, Sabri Eyuboglu, Michael Zhang, Aman Timalsina, Silas Alberti, Dylan Zinsley, James Zou, Atri Rudra, and Christopher Ré.
\newblock Simple linear attention language models balance the recall-throughput tradeoff.
\newblock In \emph{ICML 2024 Workshop on Efficient Systems for Foundation Models (ES-FoMo)}, 2024{\natexlab{a}}.
\newblock URL \url{https://arxiv.org/abs/2402.18668}.

\bibitem[Arora et~al.(2024{\natexlab{b}})Arora, Timalsina, Singhal, Spector, Eyuboglu, Zhao, Rao, Rudra, and Ré]{just_read_twice}
Simran Arora, Aman Timalsina, Aaryan Singhal, Benjamin Spector, Sabri Eyuboglu, Xinyi Zhao, Ashish Rao, Atri Rudra, and Christopher Ré.
\newblock Just read twice: closing the recall gap for recurrent language models.
\newblock \emph{arXiv preprint arXiv:2407.05483}, 2024{\natexlab{b}}.
\newblock URL \url{https://arxiv.org/abs/2407.05483}.

\bibitem[Bae et~al.(2025)Bae, Acun, Lin, Habeeb, Kim, Luo, Wang, and Wu]{bae_placement}
Sangmin Bae, Bilge Acun, Chien-Yu Lin, Haroun Habeeb, Seungyeon Kim, Liang Luo, Junjie Wang, and Carole-Jean Wu.
\newblock Hybrid architectures for language models: Systematic analysis and design insights.
\newblock \emph{arXiv preprint arXiv:2510.04800}, 2025.
\newblock URL \url{https://arxiv.org/abs/2510.04800}.

\bibitem[Bai et~al.(2024)Bai, Lv, Zhang, Lyu, Tang, Huang, Du, Liu, Zeng, Hou, Dong, Tang, and Li]{longbench}
Yushi Bai, Xin Lv, Jiajie Zhang, Hongchang Lyu, Jiankai Tang, Zhidian Huang, Zhengxiao Du, Xiao Liu, Aohan Zeng, Lei Hou, Yuxiao Dong, Jie Tang, and Juanzi Li.
\newblock Longbench: A bilingual, multitask benchmark for long context understanding.
\newblock In \emph{Annual Meeting of the Association for Computational Linguistics (ACL) 2024}, 2024.
\newblock URL \url{https://arxiv.org/abs/2308.14508}.

\bibitem[Banino et~al.(2021)Banino, Balaguer, and Blundell]{pondernet}
Andrea Banino, Jan Balaguer, and Charles Blundell.
\newblock Pondernet: Learning to ponder.
\newblock In \emph{8th ICML Workshop on Automated Machine Learning 2021}, 2021.
\newblock URL \url{https://arxiv.org/abs/2107.05407}.

\bibitem[Biderman et~al.(2024)Biderman, Schoelkopf, Sutawika, Gao, Tow, Abbasi, Aji, Ammanamanchi, Black, Clive, DiPofi, Etxaniz, Fattori, Forde, Foster, Hsu, Jaiswal, Lee, Li, Lovering, Muennighoff, Pavlick, Phang, Skowron, Tan, Tang, Wang, Winata, Yvon, and Zou]{lm_eval_harness}
Stella Biderman, Hailey Schoelkopf, Lintang Sutawika, Leo Gao, Jonathan Tow, Baber Abbasi, Alham~Fikri Aji, Pawan~Sasanka Ammanamanchi, Sidney Black, Jordan Clive, Anthony DiPofi, Julen Etxaniz, Benjamin Fattori, Jessica~Zosa Forde, Charles Foster, Jeffrey Hsu, Mimansa Jaiswal, Wilson~Y. Lee, Haonan Li, Charles Lovering, Niklas Muennighoff, Ellie Pavlick, Jason Phang, Aviya Skowron, Samson Tan, Xiangru Tang, Kevin~A. Wang, Genta~Indra Winata, François Yvon, and Andy Zou.
\newblock Lessons from the trenches on reproducible evaluation of language models.
\newblock \emph{arXiv preprint arXiv:2405.14782}, 2024.
\newblock URL \url{https://arxiv.org/abs/2405.14782}.

\bibitem[Bisk et~al.(2020)Bisk, Zellers, Le~Bras, Gao, and Choi]{piqa}
Yonatan Bisk, Rowan Zellers, Ronan Le~Bras, Jianfeng Gao, and Yejin Choi.
\newblock Piqa: Reasoning about physical commonsense in natural language.
\newblock In \emph{AAAI Conference on Artificial Intelligence 2020}, 2020.
\newblock URL \url{https://arxiv.org/abs/1911.11641}.

\bibitem[Brown et~al.(2020)Brown, Mann, Ryder, Subbiah, Kaplan, et~al.]{gpt3}
Tom~B. Brown, Benjamin Mann, Nick Ryder, Melanie Subbiah, Jared Kaplan, et~al.
\newblock Language models are few-shot learners.
\newblock In \emph{Advances in Neural Information Processing Systems (NeurIPS) 2020}, 2020.
\newblock URL \url{https://arxiv.org/abs/2005.14165}.

\bibitem[Chu et~al.(2024)Chu, Kumari, Dao, Gu, Ganti, Agrawal, Srivatsa, Wertheimer, Lim, Viros, Gonzalez, HoangTrong, Arviv, Perlitz, Shmueli, Shen, Zhang, Goodhart, Wang, Hill, Rosenkranz, Liu, Hoque, Yang, Sharma, Uong, Gala, Zawad, and Gordon]{bamba}
Linsong Chu, Divya Kumari, Tri Dao, Albert Gu, Raghu Ganti, Dakshi Agrawal, Mudhakar Srivatsa, Davis Wertheimer, Yu~Chin~Fabian Lim, Antoni Viros, Nelson Gonzalez, Tuan HoangTrong, Ofir Arviv, Yotam Perlitz, Michal Shmueli, Haochen Shen, Minjia Zhang, Gabe Goodhart, Naigang Wang, Nick Hill, Joshua Rosenkranz, Chi-Chun Liu, Adnan Hoque, Chih-Chieh Yang, Sukriti Sharma, Anh Uong, Jay Gala, Syed Zawad, and Ryan Gordon.
\newblock Bamba: Inference-efficient hybrid mamba2 model.
\newblock HuggingFace blog, December 2024, 2024.
\newblock URL \url{https://huggingface.co/blog/bamba}.

\bibitem[Clark et~al.(2018)Clark, Cowhey, Etzioni, Khot, Sabharwal, Schoenick, and Tafjord]{arc}
Peter Clark, Isaac Cowhey, Oren Etzioni, Tushar Khot, Ashish Sabharwal, Carissa Schoenick, and Oyvind Tafjord.
\newblock Think you have solved question answering? try arc, the ai2 reasoning challenge.
\newblock \emph{arXiv preprint arXiv:1803.05457}, 2018.
\newblock URL \url{https://arxiv.org/abs/1803.05457}.

\bibitem[{CodeParrot}(2021)]{codeparrot}
{CodeParrot}.
\newblock \changed{{CodeParrot-Clean-Valid}}.
\newblock Hugging Face Datasets, 2021.
\newblock URL \url{https://huggingface.co/datasets/codeparrot/codeparrot-clean-valid}.

\bibitem[Dao(2024)]{flashattention2}
Tri Dao.
\newblock {\changedB{FlashAttention-2: Faster Attention with Better Parallelism and Work Partitioning}}.
\newblock In \emph{International Conference on Learning Representations (ICLR) 2024}, 2024.
\newblock URL \url{https://arxiv.org/abs/2307.08691}.

\bibitem[Dao \& Gu(2024)Dao and Gu]{mamba2}
Tri Dao and Albert Gu.
\newblock Transformers are ssms: Generalized models and efficient algorithms through structured state space duality.
\newblock In \emph{International Conference on Machine Learning (ICML) 2024}, 2024.
\newblock URL \url{https://arxiv.org/abs/2405.21060}.

\bibitem[Dasigi et~al.(2021)Dasigi, Lo, Beltagy, Cohan, Smith, and Gardner]{qasper}
Pradeep Dasigi, Kyle Lo, Iz~Beltagy, Arman Cohan, Noah~A. Smith, and Matt Gardner.
\newblock A dataset of information-seeking questions and answers anchored in research papers.
\newblock In \emph{Proceedings of the 2021 Conference of the North American Chapter of the Association for Computational Linguistics: Human Language Technologies (NAACL-HLT) 2021}, 2021.
\newblock URL \url{https://arxiv.org/abs/2105.03011}.

\bibitem[De et~al.(2024)De, Smith, Fernando, Botev, Cristian-Muraru, Gu, Haroun, Berrada, Chen, Srinivasan, Desjardins, Doucet, Budden, Teh, Pascanu, De~Freitas, and Gulcehre]{griffin}
Soham De, Samuel~L. Smith, Anushan Fernando, Aleksandar Botev, George Cristian-Muraru, Albert Gu, Ruba Haroun, Leonard Berrada, Yutian Chen, Srivatsan Srinivasan, Guillaume Desjardins, Arnaud Doucet, David Budden, Yee~Whye Teh, Razvan Pascanu, Nando De~Freitas, and Caglar Gulcehre.
\newblock Griffin: Mixing gated linear recurrences with local attention for efficient language models.
\newblock \emph{arXiv preprint arXiv:2402.19427}, 2024.
\newblock URL \url{https://arxiv.org/abs/2402.19427}.

\bibitem[{DeepSeek-AI}(2025)]{deepseek_v32}
{DeepSeek-AI}.
\newblock Deepseek-v3.2: Pushing the frontier of open large language models.
\newblock \emph{arXiv preprint arXiv:2512.02556}, 2025.
\newblock URL \url{https://arxiv.org/abs/2512.02556}.

\bibitem[Dong et~al.(2025)Dong, Fu, Diao, Byeon, Chen, Mahabaleshwarkar, Liu, Van~Keirsbilck, Chen, Suhara, Lin, Kautz, and Molchanov]{hymba}
Xin Dong, Yonggan Fu, Shizhe Diao, Wonmin Byeon, Zijia Chen, Ameya~Sunil Mahabaleshwarkar, Shih-Yang Liu, Matthijs Van~Keirsbilck, Min-Hung Chen, Yoshi Suhara, Yingyan~Celine Lin, Jan Kautz, and Pavlo Molchanov.
\newblock Hymba: A hybrid-head architecture for small language models.
\newblock In \emph{International Conference on Learning Representations (ICLR) 2025}, 2025.
\newblock URL \url{https://arxiv.org/abs/2411.13676}.

\bibitem[Dua et~al.(2019)Dua, Wang, Dasigi, Stanovsky, Singh, and Gardner]{drop}
Dheeru Dua, Yizhong Wang, Pradeep Dasigi, Gabriel Stanovsky, Sameer Singh, and Matt Gardner.
\newblock Drop: A reading comprehension benchmark requiring discrete reasoning over paragraphs.
\newblock In \emph{Proceedings of the 2019 Conference of the North American Chapter of the Association for Computational Linguistics: Human Language Technologies (NAACL-HLT) 2019}, 2019.
\newblock URL \url{https://arxiv.org/abs/1903.00161}.

\bibitem[Fabbri et~al.(2019)Fabbri, Li, She, Li, and Radev]{multinews}
Alexander~R. Fabbri, Irene Li, Tianwei She, Suyi Li, and Dragomir~R. Radev.
\newblock Multi-news: a large-scale multi-document summarization dataset and abstractive hierarchical model.
\newblock In \emph{Proceedings of the 57th Annual Meeting of the Association for Computational Linguistics (ACL) 2019}, 2019.
\newblock URL \url{https://arxiv.org/abs/1906.01749}.

\bibitem[Gliwa et~al.(2019)Gliwa, Mochol, Biesek, and Wawer]{samsum}
Bogdan Gliwa, Iwona Mochol, Maciej Biesek, and Aleksander Wawer.
\newblock Samsum corpus: A human-annotated dialogue dataset for abstractive summarization.
\newblock In \emph{Proceedings of the 2nd Workshop on New Frontiers in Summarization (NewSum), co-located with EMNLP-IJCNLP 2019}, 2019.
\newblock URL \url{https://arxiv.org/abs/1911.12237}.

\bibitem[Glorioso et~al.(2024)Glorioso, Anthony, Tokpanov, Golubeva, Shyam, Whittington, Pilault, and Millidge]{zamba2}
Paolo Glorioso, Quentin Anthony, Yury Tokpanov, Anna Golubeva, Vasudev Shyam, James Whittington, Jonathan Pilault, and Beren Millidge.
\newblock The zamba2 suite: Technical report.
\newblock \emph{arXiv preprint arXiv:2411.15242}, 2024.
\newblock URL \url{https://arxiv.org/abs/2411.15242}.

\bibitem[Grattafiori et~al.(2024)Grattafiori, Dubey, Jauhri, Pandey, Kadian, Al-Dahle, Letman, Mathur, Schelten, Vaughan, et~al.]{llama3}
Aaron Grattafiori, Abhimanyu Dubey, Abhinav Jauhri, Abhinav Pandey, Abhishek Kadian, Ahmad Al-Dahle, Aiesha Letman, Akhil Mathur, Alan Schelten, Alex Vaughan, et~al.
\newblock The llama 3 herd of models.
\newblock \emph{arXiv preprint arXiv:2407.21783}, 2024.
\newblock URL \url{https://arxiv.org/abs/2407.21783}.

\bibitem[Graves(2016)]{act}
Alex Graves.
\newblock Adaptive computation time for recurrent neural networks.
\newblock \emph{arXiv preprint arXiv:1603.08983}, 2016.
\newblock URL \url{https://arxiv.org/abs/1603.08983}.

\bibitem[Gu \& Dao(2024)Gu and Dao]{mamba1}
Albert Gu and Tri Dao.
\newblock {\changedC{Mamba: Linear-Time Sequence Modeling with Selective State Spaces}}.
\newblock In \emph{Conference on Language Modeling (COLM) 2024}, 2024.
\newblock URL \url{https://arxiv.org/abs/2312.00752}.

\bibitem[Gu et~al.(2022)Gu, Goel, and Ré]{s4}
Albert Gu, Karan Goel, and Christopher Ré.
\newblock Efficiently modeling long sequences with structured state spaces.
\newblock In \emph{International Conference on Learning Representations (ICLR) 2022}, 2022.
\newblock URL \url{https://arxiv.org/abs/2111.00396}.

\bibitem[Gu et~al.(2025)Gu, Hu, Yang, Xi, Chen, Han, and Cai]{jet_nemotron}
Yuxian Gu, Qinghao Hu, Shang Yang, Haocheng Xi, Junyu Chen, Song Han, and Han Cai.
\newblock Jet-nemotron: Efficient language model with post neural architecture search.
\newblock \emph{arXiv preprint arXiv:2508.15884}, 2025.
\newblock URL \url{https://arxiv.org/abs/2508.15884}.

\bibitem[Guo et~al.(2023)Guo, Xu, Duan, Yin, and McAuley]{lcc}
Daya Guo, Canwen Xu, Nan Duan, Jian Yin, and Julian McAuley.
\newblock Longcoder: A long-range pre-trained language model for code completion.
\newblock In \emph{International Conference on Machine Learning (ICML) 2023}, 2023.
\newblock URL \url{https://arxiv.org/abs/2306.14893}.

\bibitem[Hao et~al.(2011)Hao, Cai, Pang, and Zhang]{swde}
Qiang Hao, Rui Cai, Yanwei Pang, and Lei Zhang.
\newblock From one tree to a forest: a unified solution for structured web data extraction.
\newblock In \emph{Proceedings of the 34th International ACM SIGIR Conference on Research and Development in Information Retrieval (SIGIR '11)}, pp.\  775--784, 2011.
\newblock URL \url{https://dl.acm.org/doi/10.1145/2009916.2010020}.

\bibitem[Ho et~al.(2020)Ho, Nguyen, Sugawara, and Aizawa]{2wikimqa}
Xanh Ho, Anh-Khoa~Duong Nguyen, Saku Sugawara, and Akiko Aizawa.
\newblock Constructing a multi-hop qa dataset for comprehensive evaluation of reasoning steps.
\newblock In \emph{Proceedings of the 28th International Conference on Computational Linguistics (COLING) 2020}, 2020.
\newblock URL \url{https://arxiv.org/abs/2011.01060}.

\bibitem[Hoffmann et~al.(2022)Hoffmann, Borgeaud, Mensch, Buchatskaya, Cai, Rutherford, de~Las~Casas, Hendricks, Welbl, Clark, Hennigan, Noland, Millican, van~den Driessche, Damoc, Guy, Osindero, Simonyan, Elsen, Vinyals, Rae, and Sifre]{chinchilla}
Jordan Hoffmann, Sebastian Borgeaud, Arthur Mensch, Elena Buchatskaya, Trevor Cai, Eliza Rutherford, Diego de~Las~Casas, Lisa~Anne Hendricks, Johannes Welbl, Aidan Clark, Tom Hennigan, Eric Noland, Katie Millican, George van~den Driessche, Bogdan Damoc, Aurelia Guy, Simon Osindero, Karen Simonyan, Erich Elsen, Oriol Vinyals, Jack~W. Rae, and Laurent Sifre.
\newblock {\changedC{Training Compute-Optimal Large Language Models}}.
\newblock In \emph{Advances in Neural Information Processing Systems (NeurIPS) 2022}, 2022.
\newblock URL \url{https://arxiv.org/abs/2203.15556}.

\bibitem[Hsieh et~al.(2024)Hsieh, Sun, Kriman, Acharya, Rekesh, Jia, Zhang, and Ginsburg]{ruler}
Cheng-Ping Hsieh, Simeng Sun, Samuel Kriman, Shantanu Acharya, Dima Rekesh, Fei Jia, Yang Zhang, and Boris Ginsburg.
\newblock Ruler: What's the real context size of your long-context language models?
\newblock In \emph{Conference on Language Modeling (COLM) 2024}, 2024.
\newblock URL \url{https://arxiv.org/abs/2404.06654}.

\bibitem[Huang et~al.(2021)Huang, Cao, Parulian, Ji, and Wang]{govreport}
Luyang Huang, Shuyang Cao, Nikolaus Parulian, Heng Ji, and Lu~Wang.
\newblock Efficient attentions for long document summarization.
\newblock In \emph{Proceedings of the 2021 Conference of the North American Chapter of the Association for Computational Linguistics: Human Language Technologies (NAACL-HLT) 2021}, 2021.
\newblock URL \url{https://arxiv.org/abs/2104.02112}.

\bibitem[Jelassi et~al.(2024)Jelassi, Brandfonbrener, Kakade, and Malach]{jelassi_repeat_after_me}
Samy Jelassi, David Brandfonbrener, Sham~M. Kakade, and Eran Malach.
\newblock Repeat after me: Transformers are better than state space models at copying.
\newblock In \emph{International Conference on Machine Learning (ICML) 2024}, 2024.
\newblock URL \url{https://arxiv.org/abs/2402.01032}.

\bibitem[Joshi et~al.(2017)Joshi, Choi, Weld, and Zettlemoyer]{triviaqa}
Mandar Joshi, Eunsol Choi, Daniel~S. Weld, and Luke Zettlemoyer.
\newblock Triviaqa: A large scale distantly supervised challenge dataset for reading comprehension.
\newblock In \emph{Proceedings of the 55th Annual Meeting of the Association for Computational Linguistics (ACL) 2017 (Volume 1: Long Papers)}, pp.\  1601--1611, 2017.
\newblock URL \url{https://arxiv.org/abs/1705.03551}.

\bibitem[Kahneman(2011)]{kahneman_thinking_fast_and_slow}
Daniel Kahneman.
\newblock Thinking, fast and slow.
\newblock Farrar, Straus and Giroux, 2011.

\bibitem[{Kimi Team}(2025)]{kimi_linear}
{Kimi Team}.
\newblock Kimi linear: An expressive, efficient attention architecture.
\newblock \emph{arXiv preprint arXiv:2510.26692}, 2025.
\newblock URL \url{https://arxiv.org/abs/2510.26692}.

\bibitem[{Kimi Team}(2026)]{kimi_k3}
{Kimi Team}.
\newblock \changed{Kimi K3: Open Frontier Intelligence}.
\newblock \emph{arXiv preprint arXiv:2607.24653}, 2026.
\newblock URL \url{https://arxiv.org/abs/2607.24653}.

\bibitem[Kočiský et~al.(2018)Kočiský, Schwarz, Blunsom, Dyer, Hermann, Melis, and Grefenstette]{narrativeqa}
Tomáš Kočiský, Jonathan Schwarz, Phil Blunsom, Chris Dyer, Karl~Moritz Hermann, Gábor Melis, and Edward Grefenstette.
\newblock The narrativeqa reading comprehension challenge.
\newblock \emph{Transactions of the Association for Computational Linguistics (TACL)}, 6:\penalty0 317--328, 2018.
\newblock URL \url{https://arxiv.org/abs/1712.07040}.

\bibitem[Kwiatkowski et~al.(2019)Kwiatkowski, Palomaki, Redfield, Collins, Parikh, Alberti, Epstein, Polosukhin, Devlin, Lee, Toutanova, Jones, Kelcey, Chang, Dai, Uszkoreit, Le, and Petrov]{nq}
Tom Kwiatkowski, Jennimaria Palomaki, Olivia Redfield, Michael Collins, Ankur Parikh, Chris Alberti, Danielle Epstein, Illia Polosukhin, Jacob Devlin, Kenton Lee, Kristina Toutanova, Llion Jones, Matthew Kelcey, Ming-Wei Chang, Andrew~M. Dai, Jakob Uszkoreit, Quoc Le, and Slav Petrov.
\newblock Natural questions: A benchmark for question answering research.
\newblock \emph{Transactions of the Association for Computational Linguistics (TACL)}, 7:\penalty0 453--466, 2019.
\newblock URL \url{https://aclanthology.org/Q19-1026/}.

\bibitem[Lahoti et~al.(2026)Lahoti, Li, Chen, Wang, Bick, Kolter, Dao, and Gu]{mamba3}
Aakash Lahoti, Kevin~Y. Li, Berlin Chen, Caitlin Wang, Aviv Bick, J.~Zico Kolter, Tri Dao, and Albert Gu.
\newblock Mamba-3: Improved sequence modeling using state space principles.
\newblock In \emph{International Conference on Learning Representations (ICLR) 2026}, 2026.
\newblock URL \url{https://arxiv.org/abs/2603.15569}.

\bibitem[Li \& Roth(2002)Li and Roth]{trec}
Xin Li and Dan Roth.
\newblock Learning question classifiers.
\newblock In \emph{Proceedings of the 19th International Conference on Computational Linguistics (COLING) 2002}, 2002.
\newblock URL \url{https://aclanthology.org/C02-1150/}.

\bibitem[Lieber et~al.(2024)Lieber, Lenz, Bata, Cohen, Osin, Dalmedigos, Safahi, Meirom, Belinkov, Shalev-Shwartz, Abend, Alon, Asida, Bergman, Glozman, Gokhman, Manevich, Ratner, Rozen, Schwartz, Zusman, and Shoham]{jamba}
Opher Lieber, Barak Lenz, Hofit Bata, Gal Cohen, Jhonathan Osin, Itay Dalmedigos, Erez Safahi, Shaked Meirom, Yonatan Belinkov, Shai Shalev-Shwartz, Omri Abend, Raz Alon, Tomer Asida, Amir Bergman, Roman Glozman, Michael Gokhman, Avshalom Manevich, Nir Ratner, Noam Rozen, Erez Schwartz, Mor Zusman, and Yoav Shoham.
\newblock {\changedC{Jamba: A Hybrid Transformer-Mamba Language Model}}.
\newblock \emph{arXiv preprint arXiv:2403.19887}, 2024.
\newblock URL \url{https://arxiv.org/abs/2403.19887}.

\bibitem[Lin et~al.(2022)Lin, Hilton, and Evans]{truthfulqa}
Stephanie Lin, Jacob Hilton, and Owain Evans.
\newblock Truthfulqa: Measuring how models mimic human falsehoods.
\newblock In \emph{Annual Meeting of the Association for Computational Linguistics (ACL) 2022}, 2022.
\newblock URL \url{https://arxiv.org/abs/2109.07958}.

\bibitem[Liu et~al.(2024)Liu, Xu, and McAuley]{repobench}
Tianyang Liu, Canwen Xu, and Julian McAuley.
\newblock Repobench: Benchmarking repository-level code auto-completion systems.
\newblock In \emph{International Conference on Learning Representations (ICLR) 2024}, 2024.
\newblock URL \url{https://arxiv.org/abs/2306.03091}.

\bibitem[Loshchilov \& Hutter(2019)Loshchilov and Hutter]{adamw}
Ilya Loshchilov and Frank Hutter.
\newblock {\changedB{Decoupled Weight Decay Regularization}}.
\newblock In \emph{International Conference on Learning Representations (ICLR) 2019}, 2019.
\newblock URL \url{https://arxiv.org/abs/1711.05101}.

\bibitem[Lu et~al.(2025)Lu, Jiang, Liu, Du, Jiang, Hong, Liu, He, Yuan, Wang, Huang, Yuan, Xu, Xu, Lai, Chen, Zheng, Yan, Su, Wu, Zhang, Yang, Zhou, Zhang, and Qiu]{moba}
Enzhe Lu, Zhejun Jiang, Jingyuan Liu, Yulun Du, Tao Jiang, Chao Hong, Shaowei Liu, Weiran He, Enming Yuan, Yuzhi Wang, Zhiqi Huang, Huan Yuan, Suting Xu, Xinran Xu, Guokun Lai, Yanru Chen, Huabin Zheng, Junjie Yan, Jianlin Su, Yuxin Wu, Neo~Y. Zhang, Zhilin Yang, Xinyu Zhou, Mingxing Zhang, and Jiezhong Qiu.
\newblock Moba: Mixture of block attention for long-context llms.
\newblock \emph{arXiv preprint arXiv:2502.13189}, 2025.
\newblock URL \url{https://arxiv.org/abs/2502.13189}.

\bibitem[Merrill et~al.(2026)Merrill, Li, Romero, Svete, Costello, Dasigi, Groeneveld, Heineman, Kuehl, Lambert, Li, Lo, Malik, Matusz, Minixhofer, Morrison, Soldaini, Timbers, Walsh, Smith, Hajishirzi, and Sabharwal]{olmo_hybrid}
William Merrill, Yanhong Li, Tyler Romero, Anej Svete, Caia Costello, Pradeep Dasigi, Dirk Groeneveld, David Heineman, Bailey Kuehl, Nathan Lambert, Chuan Li, Kyle Lo, Saumya Malik, DJ~Matusz, Benjamin Minixhofer, Jacob Morrison, Luca Soldaini, Finbarr Timbers, Pete Walsh, Noah~A. Smith, Hannaneh Hajishirzi, and Ashish Sabharwal.
\newblock Olmo hybrid: From theory to practice and back.
\newblock \emph{arXiv preprint arXiv:2604.03444}, 2026.
\newblock URL \url{https://arxiv.org/abs/2604.03444}.

\bibitem[Mihaylov et~al.(2018)Mihaylov, Clark, Khot, and Sabharwal]{openbookqa}
Todor Mihaylov, Peter Clark, Tushar Khot, and Ashish Sabharwal.
\newblock Can a suit of armor conduct electricity? a new dataset for open book question answering.
\newblock In \emph{Conference on Empirical Methods in Natural Language Processing (EMNLP) 2018}, 2018.
\newblock URL \url{https://arxiv.org/abs/1809.02789}.

\bibitem[{NVIDIA}(2026{\natexlab{a}})]{nemotron_3_super}
{NVIDIA}.
\newblock Nemotron 3 super: Open, efficient mixture-of-experts hybrid mamba-transformer model for agentic reasoning.
\newblock \emph{arXiv preprint arXiv:2604.12374}, 2026{\natexlab{a}}.
\newblock URL \url{https://arxiv.org/abs/2604.12374}.

\bibitem[{NVIDIA}(2026{\natexlab{b}})]{nemotron_3_ultra}
{NVIDIA}.
\newblock \changed{Nemotron 3 Ultra: Open, Efficient Mixture-of-Experts Hybrid Mamba-Transformer Model for Agentic Reasoning}.
\newblock \emph{arXiv preprint arXiv:2606.15007}, 2026{\natexlab{b}}.
\newblock URL \url{https://arxiv.org/abs/2606.15007}.

\bibitem[Paperno et~al.(2016)Paperno, Kruszewski, Lazaridou, Pham, Bernardi, Pezzelle, Baroni, Boleda, and Fernández]{lambada}
Denis Paperno, Germán Kruszewski, Angeliki Lazaridou, Quan~Ngoc Pham, Raffaella Bernardi, Sandro Pezzelle, Marco Baroni, Gemma Boleda, and Raquel Fernández.
\newblock The lambada dataset: Word prediction requiring a broad discourse context.
\newblock In \emph{Annual Meeting of the Association for Computational Linguistics (ACL) 2016}, 2016.
\newblock URL \url{https://arxiv.org/abs/1606.06031}.

\bibitem[Penedo et~al.(2024)Penedo, Kydlíček, Ben~allal, Lozhkov, Mitchell, Raffel, Von~Werra, and Wolf]{fineweb_edu}
Guilherme Penedo, Hynek Kydlíček, Loubna Ben~allal, Anton Lozhkov, Margaret Mitchell, Colin Raffel, Leandro Von~Werra, and Thomas Wolf.
\newblock The fineweb datasets: Decanting the web for the finest text data at scale.
\newblock In \emph{38th Conference on Neural Information Processing Systems (NeurIPS 2024) Track on Datasets and Benchmarks}, 2024.
\newblock URL \url{https://arxiv.org/abs/2406.17557}.

\bibitem[Peng et~al.(2023)Peng, Alcaide, Anthony, Albalak, et~al.]{rwkv}
Bo~Peng, Eric Alcaide, Quentin Anthony, Alon Albalak, et~al.
\newblock Rwkv: Reinventing rnns for the transformer era.
\newblock In \emph{Findings of the Association for Computational Linguistics: EMNLP 2023}, 2023.
\newblock URL \url{https://arxiv.org/abs/2305.13048}.

\bibitem[Peng et~al.(2024)Peng, Quesnelle, Fan, and Shippole]{yarn}
Bowen Peng, Jeffrey Quesnelle, Honglu Fan, and Enrico Shippole.
\newblock \changed{YaRN: Efficient Context Window Extension of Large Language Models}.
\newblock In \emph{International Conference on Learning Representations (ICLR) 2024}, 2024.
\newblock URL \url{https://arxiv.org/abs/2309.00071}.

\bibitem[{Qwen Team}(2026)]{qwen3_5}
{Qwen Team}.
\newblock Qwen3.5: Towards native multimodal agents.
\newblock qwen.ai blog / HuggingFace model card, February 2026, 2026.
\newblock URL \url{https://huggingface.co/Qwen/Qwen3.5-397B-A17B}.

\bibitem[Rae et~al.(2020)Rae, Potapenko, Jayakumar, Hillier, and Lillicrap]{pg19}
Jack~W. Rae, Anna Potapenko, Siddhant~M. Jayakumar, Chloe Hillier, and Timothy~P. Lillicrap.
\newblock Compressive transformers for long-range sequence modelling.
\newblock In \emph{International Conference on Learning Representations (ICLR) 2020}, 2020.
\newblock URL \url{https://arxiv.org/abs/1911.05507}.

\bibitem[Rajpurkar et~al.(2016)Rajpurkar, Zhang, Lopyrev, and Liang]{squad}
Pranav Rajpurkar, Jian Zhang, Konstantin Lopyrev, and Percy Liang.
\newblock Squad: 100,000+ questions for machine comprehension of text.
\newblock In \emph{Proceedings of the 2016 Conference on Empirical Methods in Natural Language Processing (EMNLP) 2016}, pp.\  2383--2392, 2016.
\newblock URL \url{https://arxiv.org/abs/1606.05250}.

\bibitem[Raposo et~al.(2024)Raposo, Ritter, Richards, Lillicrap, Humphreys, and Santoro]{mod}
David Raposo, Sam Ritter, Blake Richards, Timothy Lillicrap, Peter~Conway Humphreys, and Adam Santoro.
\newblock Mixture-of-depths: Dynamically allocating compute in transformer-based language models.
\newblock \emph{arXiv preprint arXiv:2404.02258}, 2024.
\newblock URL \url{https://arxiv.org/abs/2404.02258}.

\bibitem[Ren et~al.(2025)Ren, Liu, Lu, Shen, Liang, and Chen]{samba}
Liliang Ren, Yang Liu, Yadong Lu, Yelong Shen, Chen Liang, and Weizhu Chen.
\newblock Samba: Simple hybrid state space models for efficient unlimited context language modeling.
\newblock In \emph{International Conference on Learning Representations (ICLR) 2025}, 2025.
\newblock URL \url{https://arxiv.org/abs/2406.07522}.

\bibitem[Sakaguchi et~al.(2020)Sakaguchi, Le~Bras, Bhagavatula, and Choi]{winogrande}
Keisuke Sakaguchi, Ronan Le~Bras, Chandra Bhagavatula, and Yejin Choi.
\newblock Winogrande: An adversarial winograd schema challenge at scale.
\newblock In \emph{AAAI Conference on Artificial Intelligence 2020}, 2020.
\newblock URL \url{https://arxiv.org/abs/1907.10641}.

\bibitem[Schuster et~al.(2022)Schuster, Fisch, Gupta, Dehghani, Bahri, Tran, Tay, and Metzler]{calm}
Tal Schuster, Adam Fisch, Jai Gupta, Mostafa Dehghani, Dara Bahri, Vinh~Q. Tran, Yi~Tay, and Donald Metzler.
\newblock Confident adaptive language modeling.
\newblock In \emph{Advances in Neural Information Processing Systems (NeurIPS) 2022 (Oral)}, 2022.
\newblock URL \url{https://arxiv.org/abs/2207.07061}.

\bibitem[Shazeer(2020)]{swiglu}
Noam Shazeer.
\newblock {\changedB{GLU Variants Improve Transformer}}.
\newblock \emph{arXiv preprint arXiv:2002.05202}, 2020.
\newblock URL \url{https://arxiv.org/abs/2002.05202}.

\bibitem[Soule \& Bergmann(2025)Soule and Bergmann]{granite4}
Kate Soule and Dave Bergmann.
\newblock {\changedC{IBM Granite 4.0: Hyper-efficient, High Performance Hybrid Models for Enterprise}}.
\newblock IBM announcement, October 2025, 2025.
\newblock URL \url{https://www.ibm.com/new/announcements/ibm-granite-4-0-hyper-efficient-high-performance-hybrid-models}.

\bibitem[Su et~al.(2021)Su, Lu, Pan, Murtadha, Wen, and Liu]{rope}
Jianlin Su, Yu~Lu, Shengfeng Pan, Ahmed Murtadha, Bo~Wen, and Yunfeng Liu.
\newblock {\changedC{RoFormer: Enhanced Transformer with Rotary Position Embedding}}.
\newblock \emph{arXiv preprint arXiv:2104.09864}, 2021.
\newblock URL \url{https://arxiv.org/abs/2104.09864}.

\bibitem[Trivedi et~al.(2022)Trivedi, Balasubramanian, Khot, and Sabharwal]{musique}
Harsh Trivedi, Niranjan Balasubramanian, Tushar Khot, and Ashish Sabharwal.
\newblock Musique: Multihop questions via single-hop question composition.
\newblock \emph{Transactions of the Association for Computational Linguistics (TACL)}, 10:\penalty0 539--554, 2022.
\newblock URL \url{https://arxiv.org/abs/2108.00573}.

\bibitem[Vaswani et~al.(2017)Vaswani, Shazeer, Parmar, Uszkoreit, Jones, Gomez, Kaiser, and Polosukhin]{transformer}
Ashish Vaswani, Noam Shazeer, Niki Parmar, Jakob Uszkoreit, Llion Jones, Aidan~N. Gomez, Łukasz Kaiser, and Illia Polosukhin.
\newblock Attention is all you need.
\newblock In \emph{Advances in Neural Information Processing Systems (NeurIPS) 30}, 2017.
\newblock URL \url{https://arxiv.org/abs/1706.03762}.

\bibitem[Wang et~al.(2025)Wang, Zhu, Abreu, Shan, Kergan, Pan, Chou, Li, Wu, Zhang, Huang, and Eshraghian]{wang_survey}
Dustin Wang, Rui-Jie Zhu, Steven Abreu, Yong Shan, Taylor Kergan, Yuqi Pan, Yuhong Chou, Zheng Li, Jibin Wu, Ge~Zhang, Wenhao Huang, and Jason Eshraghian.
\newblock {\changedC{A Systematic Analysis of Hybrid Linear Attention}}.
\newblock \emph{arXiv preprint arXiv:2507.06457}, 2025.
\newblock URL \url{https://arxiv.org/abs/2507.06457}.

\bibitem[Yang et~al.(2024)Yang, Wang, Zhang, Shen, and Kim]{deltanet}
Songlin Yang, Bailin Wang, Yu~Zhang, Yikang Shen, and Yoon Kim.
\newblock Parallelizing linear transformers with the delta rule over sequence length.
\newblock In \emph{Advances in Neural Information Processing Systems (NeurIPS) 2024}, 2024.
\newblock URL \url{https://arxiv.org/abs/2406.06484}.

\bibitem[Yang et~al.(2025)Yang, Kautz, and Hatamizadeh]{gdn}
Songlin Yang, Jan Kautz, and Ali Hatamizadeh.
\newblock Gated delta networks: Improving mamba2 with delta rule.
\newblock In \emph{International Conference on Learning Representations (ICLR) 2025}, 2025.
\newblock URL \url{https://arxiv.org/abs/2412.06464}.

\bibitem[Yang et~al.(2018)Yang, Qi, Zhang, Bengio, Cohen, Salakhutdinov, and Manning]{hotpotqa}
Zhilin Yang, Peng Qi, Saizheng Zhang, Yoshua Bengio, William~W. Cohen, Ruslan Salakhutdinov, and Christopher~D. Manning.
\newblock Hotpotqa: A dataset for diverse, explainable multi-hop question answering.
\newblock In \emph{Conference on Empirical Methods in Natural Language Processing (EMNLP) 2018}, 2018.
\newblock URL \url{https://arxiv.org/abs/1809.09600}.

\bibitem[Yuan et~al.(2025)Yuan, Gao, Dai, Luo, Zhao, Zhang, Xie, Wei, Wang, Xiao, Wang, Ruan, Zhang, Liang, and Zeng]{nsa}
Jingyang Yuan, Huazuo Gao, Damai Dai, Junyu Luo, Liang Zhao, Zhengyan Zhang, Zhenda Xie, Y.~X. Wei, Lean Wang, Zhiping Xiao, Yuqing Wang, Chong Ruan, Ming Zhang, Wenfeng Liang, and Wangding Zeng.
\newblock Native sparse attention: Hardware-aligned and natively trainable sparse attention.
\newblock \emph{arXiv preprint arXiv:2502.11089}, 2025.
\newblock URL \url{https://arxiv.org/abs/2502.11089}.

\bibitem[Zellers et~al.(2019)Zellers, Holtzman, Bisk, Farhadi, and Choi]{hellaswag}
Rowan Zellers, Ari Holtzman, Yonatan Bisk, Ali Farhadi, and Yejin Choi.
\newblock Hellaswag: Can a machine really finish your sentence?
\newblock In \emph{Annual Meeting of the Association for Computational Linguistics (ACL) 2019}, 2019.
\newblock URL \url{https://arxiv.org/abs/1905.07830}.

\bibitem[Zhang \& Sennrich(2019)Zhang and Sennrich]{rmsnorm}
Biao Zhang and Rico Sennrich.
\newblock {\changedB{Root Mean Square Layer Normalization}}.
\newblock In \emph{Advances in Neural Information Processing Systems (NeurIPS) 32}, 2019.
\newblock URL \url{https://arxiv.org/abs/1910.07467}.

\bibitem[Zhong et~al.(2021)Zhong, Yin, Yu, Zaidi, Mutuma, Jha, Awadallah, Celikyilmaz, Liu, Qiu, and Radev]{qmsum}
Ming Zhong, Da~Yin, Tao Yu, Ahmad Zaidi, Mutethia Mutuma, Rahul Jha, Ahmed~Hassan Awadallah, Asli Celikyilmaz, Yang Liu, Xipeng Qiu, and Dragomir Radev.
\newblock Qmsum: A new benchmark for query-based multi-domain meeting summarization.
\newblock In \emph{Proceedings of the 2021 Conference of the North American Chapter of the Association for Computational Linguistics: Human Language Technologies (NAACL-HLT) 2021}, 2021.
\newblock URL \url{https://arxiv.org/abs/2104.05938}.

\bibitem[Zuo et~al.(2025)Zuo, Velikanov, Chahed, Belkada, Rhayem, Kunsch, Hacid, Yous, Farhat, Khadraoui, Farooq, Campesan, Cojocaru, Djilali, Hu, Chaabane, Khanna, Seddik, Huynh, Le~Khac, AlQadi, Mokeddem, Chami, Abubaker, Lubinets, Piskorski, and Frikha]{falcon_h1}
Jingwei Zuo, Maksim Velikanov, Ilyas Chahed, Younes Belkada, Dhia~Eddine Rhayem, Guillaume Kunsch, Hakim Hacid, Hamza Yous, Brahim Farhat, Ibrahim Khadraoui, Mugariya Farooq, Giulia Campesan, Ruxandra Cojocaru, Yasser Djilali, Shi Hu, Iheb Chaabane, Puneesh Khanna, Mohamed El~Amine Seddik, Ngoc~Dung Huynh, Phuc Le~Khac, Leen AlQadi, Billel Mokeddem, Mohamed Chami, Abdalgader Abubaker, Mikhail Lubinets, Kacper Piskorski, and Slim Frikha.
\newblock Falcon-h1: A family of hybrid-head language models redefining efficiency and performance.
\newblock \emph{arXiv preprint arXiv:2507.22448}, 2025.
\newblock URL \url{https://arxiv.org/abs/2507.22448}.

\end{thebibliography}
\endgroup

\newpage
\appendix
\makeatletter
\setlength{\@fptop}{0pt}
\makeatother
\FloatBarrier
\Needspace{5\baselineskip}
\section{Full Benchmark Tables}
\label{app:full-results}
Tables~\ref{tab:common_sense}--\ref{tab:longbench} give the full results underlying the main-text figures. Common-sense results include repeated-evaluation SDs, except training-time FineWeb-Edu perplexity (FW-Edu PPL); LongBench includes benchmark standard errors. LM Evaluation Harness returned N/A for retrieval and S-NIAH standard errors in our evaluations, so we report point estimates.
\begingroup
\setlength{\intextsep}{6pt}
\begin{table}[H]
\centering
\footnotesize
\setlength{\tabcolsep}{3pt}
\caption{\textbf{Common-sense results across scales.} Scores are percentages unless marked PPL; $\pm$ denotes the sample SD of five reevaluations per checkpoint.}
\label{tab:common_sense}
\resizebox{\textwidth}{!}{%
\begin{tabular}{lccccccccccc}
\toprule
\textbf{Model} & \textbf{FW-Edu} & \textbf{LAMB} & \textbf{LAMB} & \textbf{HSwag} & \textbf{PIQA} & \textbf{ARC-E} & \textbf{ARC-C} & \textbf{WinoG} & \textbf{OBQA} & \textbf{TQA mc2} & \textbf{Avg} \\
 & PPL $\downarrow$ & PPL $\downarrow$ & acc $\uparrow$ & acc $\uparrow$ & acc $\uparrow$ & acc $\uparrow$ & acc $\uparrow$ & acc $\uparrow$ & acc $\uparrow$ & acc $\uparrow$ & acc $\uparrow$ \\
\midrule
\multicolumn{12}{@{}l}{\textbf{\small 180M}} \\
Transformer & 31.96 & 402.56\scriptsize$\pm$0.00 & 18.3\scriptsize$\pm$0.00 & 27.2\scriptsize$\pm$0.00 & 60.2\scriptsize$\pm$0.00 & \underline{47.6}\scriptsize$\pm$0.00 & 18.4\scriptsize$\pm$0.00 & 50.3\scriptsize$\pm$0.00 & 15.6\scriptsize$\pm$0.00 & 45.0\scriptsize$\pm$0.00 & 35.3\scriptsize$\pm$0.00 \\
Mamba2 & 32.47 & 352.28\scriptsize$\pm$0.02 & 16.8\scriptsize$\pm$0.00 & 27.4\scriptsize$\pm$0.01 & 60.0\scriptsize$\pm$0.12 & 45.7\scriptsize$\pm$0.09 & 18.5\scriptsize$\pm$0.00 & 48.9\scriptsize$\pm$0.13 & \textbf{17.2}\scriptsize$\pm$0.00 & 43.8\scriptsize$\pm$0.02 & 34.8\scriptsize$\pm$0.02 \\
Gated DeltaNet & 31.17 & 392.58\scriptsize$\pm$0.08 & 15.6\scriptsize$\pm$0.04 & \underline{27.5}\scriptsize$\pm$0.01 & 60.9\scriptsize$\pm$0.10 & 47.4\scriptsize$\pm$0.04 & \underline{19.0}\scriptsize$\pm$0.04 & 52.1\scriptsize$\pm$0.11 & 14.2\scriptsize$\pm$0.00 & 43.6\scriptsize$\pm$0.02 & 35.0\scriptsize$\pm$0.01 \\
\addlinespace[3pt]
Mamba2 Serial Hybrid & 30.69 & 398.90\scriptsize$\pm$0.01 & 18.8\scriptsize$\pm$0.01 & \underline{27.5}\scriptsize$\pm$0.01 & 60.6\scriptsize$\pm$0.07 & 46.8\scriptsize$\pm$0.04 & \textbf{19.7}\scriptsize$\pm$0.04 & \textbf{53.0}\scriptsize$\pm$0.11 & 15.4\scriptsize$\pm$0.09 & 43.0\scriptsize$\pm$0.01 & 35.6\scriptsize$\pm$0.03 \\
Gated DeltaNet Serial Hybrid & \textbf{30.24} & 366.69\scriptsize$\pm$0.02 & 18.1\scriptsize$\pm$0.03 & 27.3\scriptsize$\pm$0.01 & \textbf{61.6}\scriptsize$\pm$0.00 & 45.7\scriptsize$\pm$0.02 & 18.6\scriptsize$\pm$0.00 & \underline{52.6}\scriptsize$\pm$0.14 & 15.8\scriptsize$\pm$0.00 & 41.3\scriptsize$\pm$0.01 & 35.1\scriptsize$\pm$0.02 \\
Mamba2 Fused Hybrid & 30.73 & 304.32\scriptsize$\pm$0.09 & 18.7\scriptsize$\pm$0.01 & \textbf{27.6}\scriptsize$\pm$0.00 & 60.1\scriptsize$\pm$0.00 & 45.5\scriptsize$\pm$0.07 & 18.6\scriptsize$\pm$0.05 & 49.4\scriptsize$\pm$0.09 & 16.0\scriptsize$\pm$0.00 & 44.0\scriptsize$\pm$0.04 & 35.0\scriptsize$\pm$0.00 \\
\addlinespace[3pt]
\textcolor{red}{\amormamba{}\amorheart{}} & 31.46 & \underline{240.27}\scriptsize$\pm$0.04 & \textbf{20.3}\scriptsize$\pm$0.06 & 27.4\scriptsize$\pm$0.00 & 60.5\scriptsize$\pm$0.12 & 46.5\scriptsize$\pm$0.05 & 18.0\scriptsize$\pm$0.09 & 51.3\scriptsize$\pm$0.22 & \underline{16.6}\scriptsize$\pm$0.00 & \textbf{45.9}\scriptsize$\pm$0.03 & \textbf{35.8}\scriptsize$\pm$0.04 \\
\textcolor{red}{\amorgdn{}\amorheart{}} & \underline{30.60} & \textbf{234.20}\scriptsize$\pm$0.02 & \underline{19.5}\scriptsize$\pm$0.05 & 27.3\scriptsize$\pm$0.01 & \underline{61.0}\scriptsize$\pm$0.04 & \textbf{47.7}\scriptsize$\pm$0.05 & 17.3\scriptsize$\pm$0.00 & 51.1\scriptsize$\pm$0.11 & 16.2\scriptsize$\pm$0.00 & \underline{45.2}\scriptsize$\pm$0.01 & \underline{35.7}\scriptsize$\pm$0.01 \\
\midrule
\multicolumn{12}{@{}l}{\textbf{\small 440M}} \\
Transformer & 19.92 & 74.09\scriptsize$\pm$0.00 & 27.8\scriptsize$\pm$0.00 & 31.2\scriptsize$\pm$0.00 & 64.9\scriptsize$\pm$0.00 & 55.9\scriptsize$\pm$0.00 & 23.9\scriptsize$\pm$0.00 & 49.9\scriptsize$\pm$0.00 & 20.4\scriptsize$\pm$0.00 & \textbf{42.4}\scriptsize$\pm$0.00 & 39.5\scriptsize$\pm$0.00 \\
Mamba2 & 19.78 & 74.19\scriptsize$\pm$0.00 & 25.3\scriptsize$\pm$0.00 & \underline{31.9}\scriptsize$\pm$0.00 & 65.2\scriptsize$\pm$0.00 & \textbf{56.9}\scriptsize$\pm$0.00 & 23.2\scriptsize$\pm$0.00 & 51.1\scriptsize$\pm$0.00 & 21.6\scriptsize$\pm$0.00 & 36.7\scriptsize$\pm$0.03 & 39.0\scriptsize$\pm$0.00 \\
Gated DeltaNet & 19.39 & 62.37\scriptsize$\pm$0.01 & 26.6\scriptsize$\pm$0.02 & \underline{31.9}\scriptsize$\pm$0.02 & \textbf{66.3}\scriptsize$\pm$0.05 & \underline{56.7}\scriptsize$\pm$0.06 & 23.4\scriptsize$\pm$0.08 & 50.5\scriptsize$\pm$0.21 & 21.2\scriptsize$\pm$0.09 & 40.7\scriptsize$\pm$0.03 & 39.7\scriptsize$\pm$0.02 \\
\addlinespace[3pt]
Mamba2 Serial Hybrid & \underline{19.11} & 69.31\scriptsize$\pm$0.01 & 28.6\scriptsize$\pm$0.03 & \textbf{32.0}\scriptsize$\pm$0.01 & \underline{66.1}\scriptsize$\pm$0.02 & 55.9\scriptsize$\pm$0.02 & 23.6\scriptsize$\pm$0.11 & 50.8\scriptsize$\pm$0.18 & 20.4\scriptsize$\pm$0.00 & 38.8\scriptsize$\pm$0.01 & 39.5\scriptsize$\pm$0.03 \\
Gated DeltaNet Serial Hybrid & \textbf{18.94} & \underline{56.21}\scriptsize$\pm$0.00 & \underline{29.6}\scriptsize$\pm$0.01 & 31.8\scriptsize$\pm$0.02 & 65.8\scriptsize$\pm$0.00 & \textbf{56.9}\scriptsize$\pm$0.07 & 24.3\scriptsize$\pm$0.14 & \underline{52.0}\scriptsize$\pm$0.09 & 21.6\scriptsize$\pm$0.11 & 39.5\scriptsize$\pm$0.03 & 40.2\scriptsize$\pm$0.01 \\
Mamba2 Fused Hybrid & 19.24 & 58.17\scriptsize$\pm$0.00 & 28.9\scriptsize$\pm$0.04 & \underline{31.9}\scriptsize$\pm$0.00 & 65.8\scriptsize$\pm$0.00 & 56.1\scriptsize$\pm$0.02 & 24.3\scriptsize$\pm$0.19 & 51.9\scriptsize$\pm$0.18 & \underline{22.2}\scriptsize$\pm$0.00 & \textbf{42.4}\scriptsize$\pm$0.02 & \underline{40.4}\scriptsize$\pm$0.04 \\
\addlinespace[3pt]
\textcolor{red}{\amormamba{}\amorheart{}} & 19.60 & 57.37\scriptsize$\pm$0.02 & 29.3\scriptsize$\pm$0.00 & 31.4\scriptsize$\pm$0.03 & 65.9\scriptsize$\pm$0.12 & 55.4\scriptsize$\pm$0.16 & \textbf{24.7}\scriptsize$\pm$0.05 & \textbf{52.4}\scriptsize$\pm$0.26 & 21.0\scriptsize$\pm$0.00 & 36.4\scriptsize$\pm$0.01 & 39.6\scriptsize$\pm$0.06 \\
\textcolor{red}{\amorgdn{}\amorheart{}} & 19.27 & \textbf{50.15}\scriptsize$\pm$0.00 & \textbf{30.2}\scriptsize$\pm$0.04 & 31.8\scriptsize$\pm$0.02 & 65.0\scriptsize$\pm$0.05 & 56.0\scriptsize$\pm$0.06 & \underline{24.6}\scriptsize$\pm$0.15 & 51.3\scriptsize$\pm$0.18 & \textbf{23.8}\scriptsize$\pm$0.00 & \underline{40.9}\scriptsize$\pm$0.01 & \textbf{40.5}\scriptsize$\pm$0.04 \\
\midrule
\multicolumn{12}{@{}l}{\textbf{\small 1.5B}} \\
Transformer & 13.41 & 25.93\scriptsize$\pm$0.00 & 38.2\scriptsize$\pm$0.00 & 38.3\scriptsize$\pm$0.00 & 69.5\scriptsize$\pm$0.00 & 65.2\scriptsize$\pm$0.00 & 29.9\scriptsize$\pm$0.00 & 52.2\scriptsize$\pm$0.00 & 23.8\scriptsize$\pm$0.00 & 35.4\scriptsize$\pm$0.00 & 44.1\scriptsize$\pm$0.00 \\
Mamba2 & 13.32 & 26.75\scriptsize$\pm$0.00 & 36.1\scriptsize$\pm$0.01 & 38.9\scriptsize$\pm$0.02 & 70.3\scriptsize$\pm$0.12 & 66.1\scriptsize$\pm$0.07 & 30.4\scriptsize$\pm$0.05 & \textbf{55.1}\scriptsize$\pm$0.52 & 25.6\scriptsize$\pm$0.00 & 34.3\scriptsize$\pm$0.00 & 44.6\scriptsize$\pm$0.03 \\
Gated DeltaNet & 13.11 & \underline{22.03}\scriptsize$\pm$0.00 & 38.5\scriptsize$\pm$0.08 & \underline{39.2}\scriptsize$\pm$0.02 & 70.7\scriptsize$\pm$0.05 & \underline{67.7}\scriptsize$\pm$0.02 & \textbf{33.4}\scriptsize$\pm$0.09 & 53.1\scriptsize$\pm$0.25 & 25.0\scriptsize$\pm$0.00 & 35.0\scriptsize$\pm$0.00 & \underline{45.3}\scriptsize$\pm$0.06 \\
\addlinespace[3pt]
Mamba2 Serial Hybrid & \underline{12.97} & 25.28\scriptsize$\pm$0.00 & 38.1\scriptsize$\pm$0.00 & \textbf{39.4}\scriptsize$\pm$0.00 & \textbf{70.9}\scriptsize$\pm$0.00 & \textbf{67.9}\scriptsize$\pm$0.00 & 31.3\scriptsize$\pm$0.00 & 53.0\scriptsize$\pm$0.00 & 26.0\scriptsize$\pm$0.00 & 35.4\scriptsize$\pm$0.02 & \underline{45.3}\scriptsize$\pm$0.00 \\
Gated DeltaNet Serial Hybrid & \textbf{12.94} & 22.28\scriptsize$\pm$0.00 & \textbf{39.7}\scriptsize$\pm$0.05 & \underline{39.2}\scriptsize$\pm$0.03 & \underline{70.8}\scriptsize$\pm$0.03 & 64.9\scriptsize$\pm$0.00 & 30.5\scriptsize$\pm$0.00 & 54.2\scriptsize$\pm$0.00 & 24.6\scriptsize$\pm$0.00 & \underline{37.7}\scriptsize$\pm$0.02 & 45.2\scriptsize$\pm$0.01 \\
Mamba2 Fused Hybrid & 13.04 & 23.71\scriptsize$\pm$0.00 & 37.5\scriptsize$\pm$0.05 & \underline{39.2}\scriptsize$\pm$0.00 & 70.4\scriptsize$\pm$0.12 & 67.4\scriptsize$\pm$0.05 & \underline{32.3}\scriptsize$\pm$0.09 & 53.7\scriptsize$\pm$0.13 & \underline{26.6}\scriptsize$\pm$0.00 & 35.2\scriptsize$\pm$0.00 & \underline{45.3}\scriptsize$\pm$0.02 \\
\addlinespace[3pt]
\textcolor{red}{\amormamba{}\amorheart{}} & 13.30 & 23.01\scriptsize$\pm$0.01 & \underline{39.2}\scriptsize$\pm$0.02 & 38.6\scriptsize$\pm$0.01 & \textbf{70.9}\scriptsize$\pm$0.05 & 67.0\scriptsize$\pm$0.05 & 30.2\scriptsize$\pm$0.07 & \underline{54.7}\scriptsize$\pm$0.16 & \textbf{26.8}\scriptsize$\pm$0.00 & \textbf{39.0}\scriptsize$\pm$0.02 & \textbf{45.8}\scriptsize$\pm$0.03 \\
\textcolor{red}{\amorgdn{}\amorheart{}} & 13.10 & \textbf{21.49}\scriptsize$\pm$0.01 & 39.1\scriptsize$\pm$0.06 & \underline{39.2}\scriptsize$\pm$0.02 & 70.6\scriptsize$\pm$0.07 & 66.4\scriptsize$\pm$0.02 & 30.5\scriptsize$\pm$0.00 & 53.7\scriptsize$\pm$0.07 & 24.8\scriptsize$\pm$0.09 & 36.9\scriptsize$\pm$0.00 & 45.1\scriptsize$\pm$0.02 \\
\bottomrule
\end{tabular}
}
\end{table}

\begin{table}[H]
  \centering
  \footnotesize
  \setlength{\tabcolsep}{3pt}
  \caption{\textbf{Retrieval and S-NIAH.} 1.5B; retrieval measures answer-containment accuracy at 2K; S-NIAH reports accuracy by variant and context length. Scores are percentages.}
  \label{tab:retrieval}
  \resizebox{\textwidth}{!}{%
  \begin{tabular}{l ccccccccccccccc}
    \toprule
    \textbf{Model} & \textbf{SWDE}$\uparrow$ & \textbf{SQuAD}$\uparrow$ & \textbf{FDA}$\uparrow$ & \textbf{TQA}$\uparrow$ & \textbf{NQ}$\uparrow$ & \textbf{DROP}$\uparrow$ & \multicolumn{3}{c}{\textbf{S-NIAH-1}$\uparrow$} & \multicolumn{3}{c}{\textbf{S-NIAH-2}$\uparrow$} & \multicolumn{3}{c}{\textbf{S-NIAH-3}$\uparrow$} \\
    \cmidrule(lr){2-7}\cmidrule(lr){8-10}\cmidrule(lr){11-13}\cmidrule(lr){14-16}
    \textbf{Context length} & \multicolumn{6}{c}{2048} & 1024 & 2048 & 4096 & 1024 & 2048 & 4096 & 1024 & 2048 & 4096 \\
    \midrule
    Transformer & \underline{54.5} & 22.7 & 40.1 & \underline{41.7} & 10.5 & 18.3 & \textbf{100.0} & \textbf{100.0} & 12.6 & \textbf{100.0} & \textbf{100.0} & 14.6 & 90.6 & 87.4 & \phantom{0}0.2 \\
    Mamba2 & 19.9 & 33.5 & 14.5 & 39.6 & \phantom{0}8.7 & 17.0 & \textbf{100.0} & 93.6 & 60.4 & \textbf{100.0} & 38.4 & 28.4 & 72.4 & 40.2 & 14.4 \\
    Gated DeltaNet & 25.3 & 35.4 & 11.8 & 38.7 & \phantom{0}9.6 & 18.1 & \underline{99.4} & 91.4 & 60.8 & \underline{99.4} & 17.8 & 38.8 & 71.6 & 47.2 & 24.2 \\
    \addlinespace[3pt]
    Mamba2 Serial Hybrid & 49.1 & \textbf{44.6} & 37.9 & \textbf{43.1} & \textbf{10.8} & 18.4 & \textbf{100.0} & \underline{99.6} & \underline{68.6} & \textbf{100.0} & \textbf{100.0} & \underline{64.0} & \underline{91.2} & \textbf{91.0} & 20.4 \\
    Gated DeltaNet Serial Hybrid & \textbf{57.1} & \underline{41.5} & \textbf{49.4} & \textbf{43.1} & \underline{10.7} & \underline{19.8} & \textbf{100.0} & \textbf{100.0} & 67.8 & \textbf{100.0} & \textbf{100.0} & 34.0 & \textbf{91.4} & \underline{90.8} & \underline{28.4} \\
    Mamba2 Fused Hybrid & 45.1 & 39.9 & \underline{48.5} & 38.4 & \phantom{0}9.6 & 18.1 & \textbf{100.0} & \textbf{100.0} & \textbf{81.4} & \textbf{100.0} & \textbf{100.0} & \textbf{69.6} & 89.6 & 90.2 & 22.8 \\
    \addlinespace[3pt]
    \textcolor{red}{\amormamba{}\amorheart{}} & 33.8 & 36.0 & 21.2 & 40.4 & 10.2 & \textbf{21.1} & \textbf{100.0} & 95.8 & 44.4 & \textbf{100.0} & \underline{99.6} & 47.2 & 70.8 & 83.6 & 25.6 \\
    \textcolor{red}{\amorgdn{}\amorheart{}} & 47.4 & 36.8 & 24.9 & \textbf{43.1} & 10.5 & 17.4 & \textbf{100.0} & 96.2 & 48.0 & \textbf{100.0} & 98.6 & 53.0 & 88.8 & 86.0 & \textbf{29.4} \\
    \bottomrule
  \end{tabular}}
\end{table}

\begin{table}[H]
  \centering
  \footnotesize
  \setlength{\tabcolsep}{3pt}
  \caption{\textbf{LongBench.} 1.5B; scores are percentages; $\pm$ denotes standard error.}
  \label{tab:longbench}
  \resizebox{\textwidth}{!}{%
  \begin{tabular}{l ccccccccccccccc}
    \toprule
    \textbf{Model} & \multicolumn{3}{c}{\textbf{Single-Doc QA}$\uparrow$} & \multicolumn{3}{c}{\textbf{Multi-Doc QA}$\uparrow$} & \multicolumn{3}{c}{\textbf{Summarization}$\uparrow$} & \multicolumn{3}{c}{\textbf{Few-shot}$\uparrow$} & \multicolumn{2}{c}{\textbf{Code}$\uparrow$} & \textbf{Avg}$\uparrow$ \\
    \cmidrule(lr){2-4}\cmidrule(lr){5-7}\cmidrule(lr){8-10}\cmidrule(lr){11-13}\cmidrule(lr){14-15}
     & \textbf{NQA} & \textbf{QQA} & \textbf{MFQ} & \textbf{HQA} & \textbf{2WM} & \textbf{Mus} & \textbf{GvR} & \textbf{QMS} & \textbf{MNs} & \textbf{TRC} & \textbf{TQA} & \textbf{SSM} & \textbf{LCC} & \textbf{RBP} &  \\
    \midrule
    Transformer & \phantom{0}0.3\scriptsize$\pm$0.11 & \phantom{0}1.6\scriptsize$\pm$0.18 & \phantom{0}4.9\scriptsize$\pm$0.65 & \phantom{0}0.6\scriptsize$\pm$0.31 & \phantom{0}1.4\scriptsize$\pm$0.38 & \phantom{0}0.2\scriptsize$\pm$0.13 & \phantom{0}5.3\scriptsize$\pm$0.24 & \phantom{0}5.8\scriptsize$\pm$0.37 & 10.7\scriptsize$\pm$0.49 & \phantom{0}3.0\scriptsize$\pm$1.21 & \phantom{0}3.3\scriptsize$\pm$0.93 & \phantom{0}3.6\scriptsize$\pm$0.49 & \phantom{0}9.0\scriptsize$\pm$0.27 & \phantom{0}7.7\scriptsize$\pm$0.31 & \phantom{0}4.1\scriptsize$\pm$0.14 \\
    Mamba2 & \underline{\phantom{0}1.6}\scriptsize$\pm$0.16 & \underline{\phantom{0}3.9}\scriptsize$\pm$0.40 & \underline{11.1}\scriptsize$\pm$0.91 & \textbf{\phantom{0}4.6}\scriptsize$\pm$0.60 & \phantom{0}6.8\scriptsize$\pm$0.74 & \textbf{\phantom{0}2.8}\scriptsize$\pm$0.44 & \phantom{0}7.5\scriptsize$\pm$0.28 & \textbf{15.6}\scriptsize$\pm$0.58 & \underline{11.3}\scriptsize$\pm$0.43 & \phantom{0}7.0\scriptsize$\pm$1.81 & \underline{14.3}\scriptsize$\pm$1.33 & \phantom{0}7.2\scriptsize$\pm$0.57 & \underline{10.1}\scriptsize$\pm$0.33 & 11.0\scriptsize$\pm$0.44 & \phantom{0}8.2\scriptsize$\pm$0.21 \\
    Gated DeltaNet & \phantom{0}1.4\scriptsize$\pm$0.15 & \phantom{0}3.6\scriptsize$\pm$0.33 & \phantom{0}9.4\scriptsize$\pm$0.96 & \phantom{0}3.3\scriptsize$\pm$0.44 & \textbf{\phantom{0}8.3}\scriptsize$\pm$0.80 & \underline{\phantom{0}2.4}\scriptsize$\pm$0.37 & \underline{\phantom{0}8.5}\scriptsize$\pm$0.27 & 14.9\scriptsize$\pm$0.56 & 10.8\scriptsize$\pm$0.40 & \phantom{0}5.5\scriptsize$\pm$1.62 & \textbf{17.0}\scriptsize$\pm$1.48 & \textbf{19.3}\scriptsize$\pm$0.93 & \phantom{0}9.5\scriptsize$\pm$0.29 & 10.8\scriptsize$\pm$0.40 & \textbf{\phantom{0}8.9}\scriptsize$\pm$0.21 \\
    \addlinespace[3pt]
    Mamba2 Serial Hybrid & \phantom{0}0.2\scriptsize$\pm$0.07 & \phantom{0}2.2\scriptsize$\pm$0.31 & \phantom{0}4.4\scriptsize$\pm$0.79 & \phantom{0}0.2\scriptsize$\pm$0.10 & \phantom{0}2.5\scriptsize$\pm$0.49 & \phantom{0}0.4\scriptsize$\pm$0.17 & \phantom{0}2.8\scriptsize$\pm$0.25 & \phantom{0}4.5\scriptsize$\pm$0.30 & 10.0\scriptsize$\pm$0.53 & \phantom{0}4.5\scriptsize$\pm$1.47 & \phantom{0}4.2\scriptsize$\pm$1.05 & \phantom{0}6.6\scriptsize$\pm$0.57 & 10.0\scriptsize$\pm$0.29 & 10.3\scriptsize$\pm$0.32 & \phantom{0}4.5\scriptsize$\pm$0.16 \\
    Gated DeltaNet Serial Hybrid & \phantom{0}0.5\scriptsize$\pm$0.10 & \phantom{0}2.4\scriptsize$\pm$0.26 & \phantom{0}5.6\scriptsize$\pm$0.70 & \phantom{0}1.3\scriptsize$\pm$0.27 & \phantom{0}4.8\scriptsize$\pm$0.67 & \phantom{0}1.1\scriptsize$\pm$0.26 & \phantom{0}4.1\scriptsize$\pm$0.28 & \phantom{0}5.6\scriptsize$\pm$0.39 & 10.5\scriptsize$\pm$0.47 & \phantom{0}7.5\scriptsize$\pm$1.87 & \phantom{0}9.2\scriptsize$\pm$1.34 & \phantom{0}5.4\scriptsize$\pm$0.68 & \phantom{0}9.6\scriptsize$\pm$0.33 & \underline{11.2}\scriptsize$\pm$0.36 & \phantom{0}5.6\scriptsize$\pm$0.20 \\
    Mamba2 Fused Hybrid & \phantom{0}1.2\scriptsize$\pm$0.14 & \phantom{0}3.4\scriptsize$\pm$0.34 & \phantom{0}9.5\scriptsize$\pm$0.82 & \phantom{0}3.1\scriptsize$\pm$0.47 & \phantom{0}6.4\scriptsize$\pm$0.80 & \phantom{0}2.2\scriptsize$\pm$0.40 & \textbf{\phantom{0}9.3}\scriptsize$\pm$0.29 & 14.8\scriptsize$\pm$0.50 & \phantom{0}8.1\scriptsize$\pm$0.40 & \phantom{0}3.0\scriptsize$\pm$1.21 & 14.2\scriptsize$\pm$1.57 & \phantom{0}6.0\scriptsize$\pm$0.52 & \textbf{11.2}\scriptsize$\pm$0.53 & 10.8\scriptsize$\pm$0.47 & \phantom{0}7.4\scriptsize$\pm$0.19 \\
    \addlinespace[3pt]
    \textcolor{red}{\amormamba{}\amorheart{}} & \underline{\phantom{0}1.6}\scriptsize$\pm$0.15 & \phantom{0}3.3\scriptsize$\pm$0.32 & 10.1\scriptsize$\pm$0.83 & \phantom{0}2.9\scriptsize$\pm$0.50 & \phantom{0}6.1\scriptsize$\pm$0.71 & \phantom{0}1.8\scriptsize$\pm$0.37 & \phantom{0}7.6\scriptsize$\pm$0.23 & 14.7\scriptsize$\pm$0.46 & \textbf{13.2}\scriptsize$\pm$0.47 & \textbf{13.0}\scriptsize$\pm$2.38 & 13.2\scriptsize$\pm$1.49 & \phantom{0}7.5\scriptsize$\pm$0.51 & \phantom{0}9.3\scriptsize$\pm$0.26 & \textbf{11.5}\scriptsize$\pm$0.29 & \phantom{0}8.3\scriptsize$\pm$0.23 \\
    \textcolor{red}{\amorgdn{}\amorheart{}} & \textbf{\phantom{0}1.7}\scriptsize$\pm$0.17 & \textbf{\phantom{0}4.0}\scriptsize$\pm$0.35 & \textbf{11.4}\scriptsize$\pm$0.87 & \underline{\phantom{0}3.6}\scriptsize$\pm$0.49 & \underline{\phantom{0}7.0}\scriptsize$\pm$0.74 & \phantom{0}1.9\scriptsize$\pm$0.32 & \phantom{0}8.1\scriptsize$\pm$0.30 & \underline{15.1}\scriptsize$\pm$0.50 & 10.8\scriptsize$\pm$0.43 & \underline{11.0}\scriptsize$\pm$2.22 & 11.8\scriptsize$\pm$1.22 & \underline{11.4}\scriptsize$\pm$0.68 & \underline{10.1}\scriptsize$\pm$0.38 & 10.3\scriptsize$\pm$0.41 & \underline{\phantom{0}8.4}\scriptsize$\pm$0.22 \\
    \bottomrule
  \end{tabular}}
\end{table}

\endgroup

LongBench standard errors come from the stored harness outputs. The average's standard error is $\sqrt{\sum_i\mathrm{SE}_i^2}/14$, assuming independent tasks; it measures fixed-checkpoint sampling error.

\FloatBarrier
\Needspace{5\baselineskip}
\subsection{Length extrapolation results}
\begingroup

RoPE-induced distribution shift beyond the $3$K training context length leads to sharp perplexity increases for the Transformer and the two serial hybrids (Fig.~\ref{fig:length_extrap}).  \amor{} does not suffer as much, behaving more similarly to recurrent models. For visualization clarity, the y-axes in each panel are linearly capped at $1.3\times$ the maximum value observed among the well-behaved subset $\{\text{Mamba2}, \text{Gated DeltaNet}, \text{Mamba2 Fused Hybrid}, \text{\amormamba{}}, \text{\amorgdn{}}\}$, ensuring that extreme outliers do not dominate the scale while preserving relative trends among stable models.

\begin{figure}[h]
  \centering
  \includegraphics[width=\textwidth]{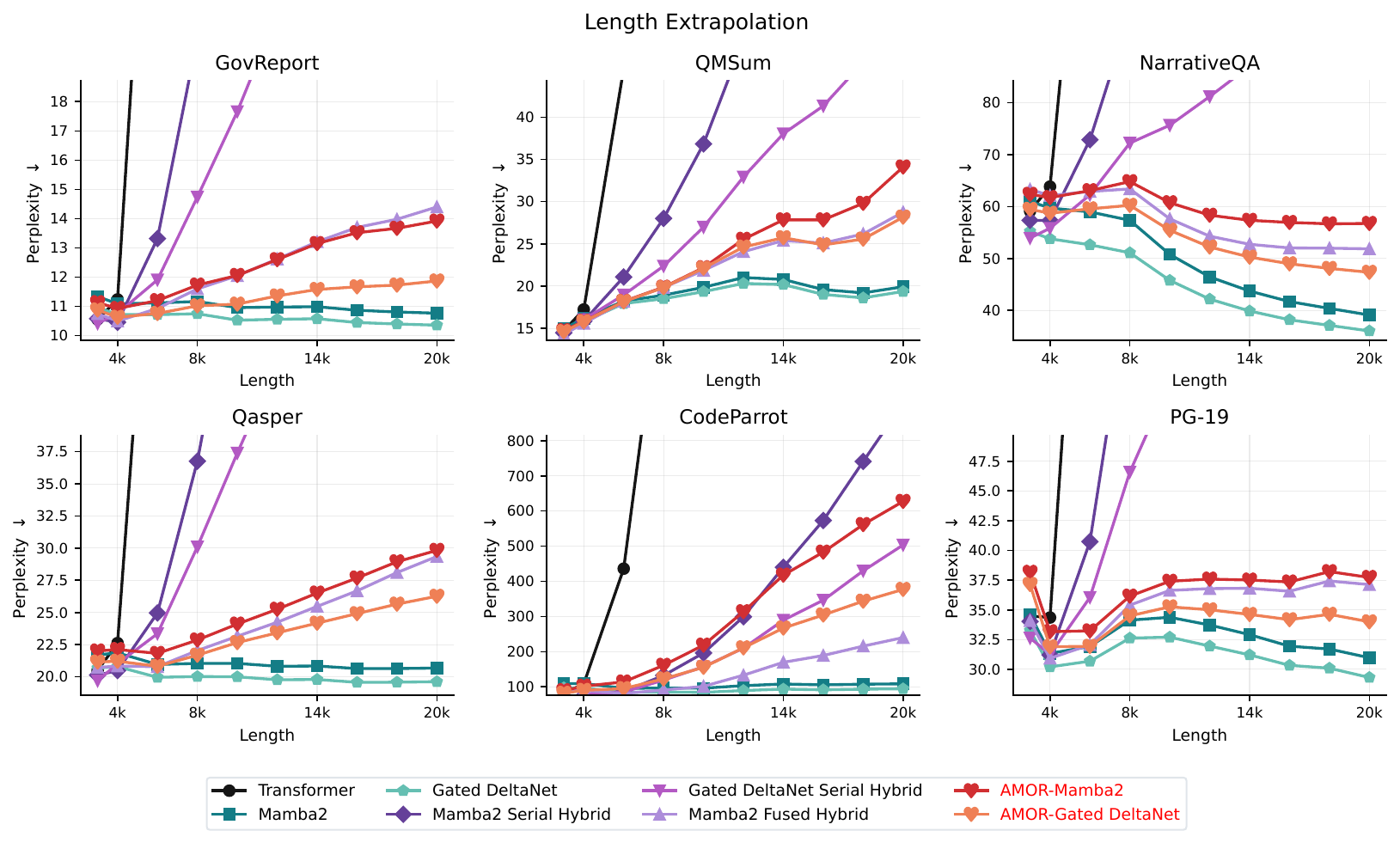}
  \caption{\textbf{Length extrapolation across six domains.} 1.5B; per-token perplexity from the 3K training context to 20K; lower is better.}
  \label{fig:length_extrap}
\end{figure}
\endgroup

\FloatBarrier
\Needspace{5\baselineskip}
\subsection{Repeatability and Training-Seed Results}
\label{app:iclr-repeats}
\textbf{Evaluation repeatability.} We reevaluate each fixed checkpoint five times. Table~\ref{tab:common_sense} reports canonical scores with sample SDs from these repeats; the average's SD is computed across the five eight-task means. The largest SD is 0.06 points for an eight-task mean and 0.52 points for a single task (WinoGrande).

\textbf{Independent training runs.} At 180M, every architecture is additionally trained with seeds 37, 67, 69 and 78, each trained to a Chinchilla-optimal token budget. Sample SDs of the eight-task mean range from 0.20 to 0.48 points (median 0.24; Table~\ref{tab:iclr-seeds}).
Seeds 37, 67, 69 and 78 control initialization and training-data order. The canonical seed-42 run is excluded from these statistics. Consistent with the canonical seed-42 results in the main paper, an \amor{} variant leads at each training seed and in the average across all four seeds.
\label{app:iclr-seeds}
\begin{table}[h]
  \centering
  \small
  \setlength{\tabcolsep}{5pt}
  \caption{\textbf{Eight-task common-sense averages at 180M across training seeds.} Evaluation settings follow Appendix~\ref{app:eval:common}.}
  \label{tab:iclr-seeds}
  \begin{tabular}{lrrrrrr}
    \toprule
    \textbf{Model} & \multicolumn{6}{c}{\textbf{Common-sense avg} (acc, \%) $\uparrow$} \\
    \cmidrule(lr){2-7}
     & Seed 37 & Seed 67 & Seed 69 & Seed 78 & Avg & SD \\
    \midrule
    Transformer & 34.6 & 35.1 & 35.1 & 34.9 & 34.9 & 0.24 \\
    Mamba2 & 35.1 & 35.0 & 34.7 & 34.7 & 34.9 & 0.20 \\
    Gated DeltaNet & 35.4 & \underline{35.6} & 34.8 & 35.0 & 35.2 & 0.40 \\
    \addlinespace[3pt]
    Mamba2 Serial Hybrid & 35.5 & 35.2 & 35.5 & 35.1 & 35.3 & 0.23 \\
    Gated DeltaNet Serial Hybrid & \underline{36.0} & 35.5 & 34.9 & 35.1 & \underline{35.4} & 0.48 \\
    Mamba2 Fused Hybrid & 35.4 & 35.3 & 35.6 & 35.1 & 35.3 & 0.21 \\
    \addlinespace[3pt]
    \textcolor{red}{\amormamba{}\amorheart{}} & \textbf{36.2} & 35.5 & \underline{35.7} & \textbf{35.8} & \textbf{35.8} & 0.28 \\
    \textcolor{red}{\amorgdn{}\amorheart{}} & 35.6 & \textbf{35.9} & \textbf{36.0} & \underline{35.6} & \textbf{35.8} & 0.20 \\
    \bottomrule
  \end{tabular}
\end{table}

\FloatBarrier
\Needspace{5\baselineskip}
\section{Architectural Details}
\label{app:arch}

All eight architectures share model width $d_{\mathrm{model}}$ and backbone depth $N$. The reference dimensions are $d_{\mathrm{model}}/N/d_{\mathrm{ff}} = 768/12/1216$ (180M), $1024/24/1984$ (440M), and $2048/24/4096$ (1.5B), where $d_{\mathrm{ff}}$ is the SwiGLU hidden width~\citep{swiglu}. Table~\ref{tab:per_model_params} reports parameter counts and attention placement. Canonical \amor{} appends $K=3$ blocks to the $N$ mixer--MLP layers; the $K=1$ ablation covers both backbones and all scales (Table~\ref{tab:vs_classic}).

\paragraph{Controlled design.}
To isolate the effect of \emph{fixed-schedule attention}, we remove orthogonal components from the canonical Jamba and Hymba designs. All hybrids use full causal attention rather than sliding-window attention, and use the same corresponding Mamba2 or Gated DeltaNet backbone. We exclude Hymba's meta tokens and KV reuse, as well as Jamba's mixture-of-experts layers.

Fixing backbone depth does not equalize parameter counts because the sequence mixers have different parameterizations. The additional \amor{} blocks introduce a modest parameter overhead relative to their recurrent backbones (approximately 2--4\%; Table~\ref{tab:per_model_params}). The Transformer has the fewest parameters at each scale because its attention mixers are lighter than the recurrent mixers in this comparison.

\begin{table}[h]
  \centering
  \small
  \setlength{\tabcolsep}{4pt}
  \caption{\textbf{Architecture configurations across scales.} 180M, 440M and 1.5B; parameters, attention placement, head dimension and backbone depth. N/A denotes no attention layers.}
  \label{tab:per_model_params}
  \resizebox{\textwidth}{!}{%
  \begin{tabular}{lcccccc}
    \toprule
    \multirow{2}{*}{\textbf{Model}}
      & \multicolumn{3}{c}{\textbf{Params}}
      & \multirow{2}{*}{\textbf{Attn layers}}
      & \multirow{2}{*}{\textbf{Attn dim}}
      & \multirow{2}{*}{\textbf{$N$}} \\
    \cmidrule{2-4}
    & 180M & 440M & 1.5B & & & \\
    \midrule
    Transformer                        & 160.5M & 378.3M & 1{,}269.4M & all $N$       & 64 head & 12 / 24 / 24 \\
    Mamba2                             & 177.3M & 436.1M & 1{,}487.1M & 0             & N/A     & 12 / 24 / 24 \\
    Gated DeltaNet                     & 174.9M & 429.5M & 1{,}473.7M & 0             & N/A     & 12 / 24 / 24 \\
    Mamba2 Serial Hybrid               & 174.5M & 428.8M & 1{,}459.9M & $\{4,8\}$ / $\{6,12,18\}$ & 64 head & 12 / 24 / 24 \\
    Gated DeltaNet Serial Hybrid       & 172.5M & 423.1M & 1{,}448.1M & $\{4,8\}$ / $\{6,12,18\}$ & 64 head & 12 / 24 / 24 \\
    Mamba2 Fused Hybrid                & 179.1M & 439.2M & 1{,}499.7M & $\{0,6,11\}$ / $\{0,12,23\}$ & 64 head & 12 / 24 / 24 \\
    \textcolor{red}{\amormamba{}\amorheart{}}     & 184.4M & 448.7M & 1{,}537.4M & $K{=}3$ post-hoc & 64 head & 12 / 24 / 24 \\
    \textcolor{red}{\amorgdn{}\amorheart{}}       & 182.0M & 442.1M & 1{,}524.0M & $K{=}3$ post-hoc & 64 head & 12 / 24 / 24 \\
    \bottomrule
  \end{tabular}}
\end{table}

Block normalization and the \amor{} block design are described in Section~\ref{sec:method:arch}; symmetric-layout architecture ablation results are in Appendix~\ref{app:ablations}.

\FloatBarrier
\Needspace{5\baselineskip}
\section{Training Recipe}
\label{app:training}

Canonical runs share the training corpus, Llama-3.1 tokenizer~\citep{llama3} and AdamW~\citep{adamw} recipe within each scale (Table~\ref{tab:training_recipe}); seed 42 controls initialization and training-data order. The common token budget is 20 times the parameter count of \amormamba{}, following Chinchilla scaling~\citep{chinchilla} and giving every architecture the same data volume at that scale. Results across training seeds are in Appendix~\ref{app:iclr-seeds}.

\begin{table}[h]
  \centering
  \small
  \setlength{\tabcolsep}{6pt}
  \caption{\textbf{All architectures share one training recipe per scale.} 180M, 440M and 1.5B; \amor{}'s $\boldsymbol{\alpha}$ uses a separate AdamW group without weight decay.}
  \label{tab:training_recipe}
  \begin{tabular}{lccc}
    \toprule
    & 180M & 440M & 1.5B \\
    \midrule
    Tokens                       & 3.69B    & 8.97B    & 30.7B \\
    Steps                        & 7{,}503  & 18{,}255 & 62{,}558 \\
    Sequence length              & 3{,}072     & 3{,}072     & 3{,}072 \\
    Effective batch (tokens)     & 491,520  & 491,520  & 491,520 \\
    Optimizer                    & \multicolumn{3}{c}{AdamW $(\beta_1,\beta_2){=}(0.9, 0.95)$, weight decay 0.1, grad clip 1.0} \\
    Backbone peak LR             & $3{\times}10^{-4}$ & $3{\times}10^{-4}$ & $1.5{\times}10^{-4}$ \\
    LR floor                     & $10^{-5}$  & $10^{-5}$  & $10^{-5}$ \\
    LR warmup steps              & 2{,}000  & 2{,}000  & 2{,}000 \\
    LR schedule                  & \multicolumn{3}{c}{Cosine} \\
    Precision                    & \multicolumn{3}{c}{bfloat16 autocast} \\
    Tokenizer                    & \multicolumn{3}{c}{Llama-3.1, $V{=}128{,}256$} \\
    Seed                         & 42 & 42 & 42 \\
    \midrule
    \multicolumn{4}{l}{\emph{\amor{}-specific (separate AdamW group)}} \\
    \midrule
    $\boldsymbol{\alpha}$ peak LR & $3{\times}10^{-3}$ & $3{\times}10^{-3}$ & $3{\times}10^{-3}$ \\
    $\boldsymbol{\alpha}$ weight decay & 0 & 0 & 0 \\
    Init $\tilde{\boldsymbol{\alpha}}$ & 0 ($\boldsymbol{\alpha}{=}0.5$) & 0 & 0 \\
    Init $\mathbf{W}_{\!O}$ & $\mathbf{0}$ & $\mathbf{0}$ & $\mathbf{0}$ \\
    EMA momentum & 0.99 & 0.99 & 0.99 \\
    $\kappa$ (offset multiplier) & 0.2 & 0.2 & 0.2 \\
    \bottomrule
  \end{tabular}
\end{table}

After construction, we initialize embedding and linear projection weights from $\mathcal N(0,0.02)$, except the \amor{} attention output projections, which start at zero. Normalization weights and recurrent-specific parameters retain their respective initializations. The embedding and LM head are tied, and the canonical models' linear projections have no bias.

\FloatBarrier
\Needspace{5\baselineskip}
\subsection{Training Mode}
\label{app:method:train}

We train in $\mathrm{full\_with\_mask}$ mode, in which $\mathbf{Q}, \mathbf{K}, \mathbf{V}$ and the dense causal SDPA are computed for every position, and the gate masks the attention output per position. This is mathematically equivalent to $\mathrm{true\_sparse}$ formulation that computes attention only at firing positions; both modes produce the same residual update in exact arithmetic.

We use the dense formulation for training purely for efficiency: it maps to PyTorch's \texttt{scaled\_dot\_product\_attention} (FlashAttention-2 kernels), whereas $\mathrm{true\_sparse}$ requires per-row dynamic gather/scatter without a comparable fused implementation, resulting in lower GPU throughput. The choice therefore affects only implementation efficiency, not model semantics.

\FloatBarrier
\Needspace{5\baselineskip}
\section{Entropy Gate Behavior}
\label{app:gate-stability}
\FloatBarrier
\Needspace{5\baselineskip}
\subsection{Training statistics}

\begin{figure}[h]
  \centering
  \includegraphics[width=\textwidth]{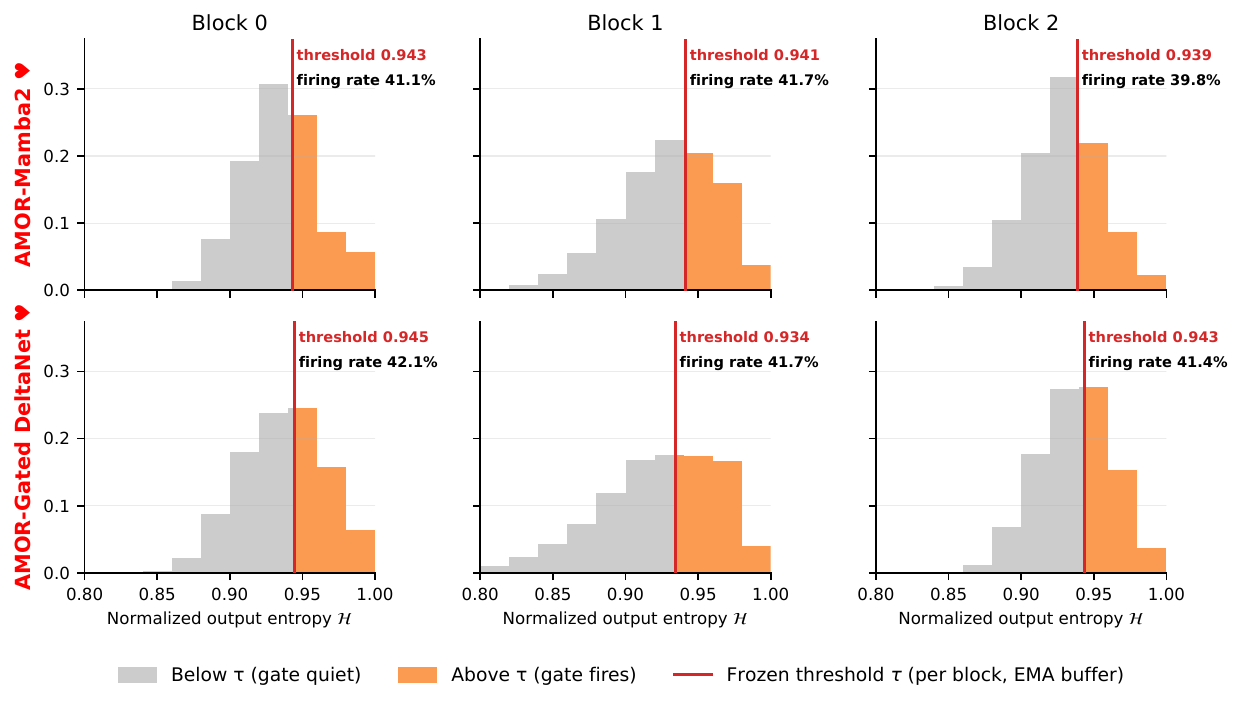}
  \caption{\textbf{Training entropy concentrates near the frozen thresholds.} 1.5B; final-window histograms. Red lines mark frozen entropy thresholds, and orange marks mass above them.}
  \label{fig:entropy_hist_training}
\end{figure}

Table~\ref{tab:gate_stats} summarizes Block~0 training statistics over the final logged entropy window: mean batch median $\bar\mu$, mean batch SD $\bar\sigma$, derived threshold $\bar\tau=\bar\mu+0.2\bar\sigma$, and the mean per-channel scaling vector $\boldsymbol{\alpha}$. The derived threshold is not a direct readout of the frozen EMA buffers at the final checkpoint. With $\kappa=0.2$ fixed, entropy medians rise and their standard deviations fall with scale, while the 10th--90th percentile ranges of logged batch firing rates remain near 33--51\%. Figure~\ref{fig:entropy_hist_training} shows per-block entropy histograms and frozen entropy thresholds at 1.5B. Each histogram aggregates the final 20 entropy-log entries for its block, sampled every 500 steps from step 53,000 to 62,500 (approximately 15\% of training). Firing-rate annotations average the logged rates over that window.

\begin{table}[h]
  \centering
  \footnotesize
  \setlength{\tabcolsep}{4.5pt}
  \caption{\textbf{Training firing-rate ranges remain similar across scales.} 180M, 440M and 1.5B; Block~0 statistics: $\bar\mu$ and $\bar\sigma$ are mean batch median and SD, $\bar\tau=\bar\mu+0.2\bar\sigma$.}
  \label{tab:gate_stats}
  \begin{tabular}{llcccccc}
    \toprule
    \textbf{Model} & \textbf{Scale} & \textbf{FW-Edu} & \textbf{$\bar\mu$} & \textbf{$\bar\sigma$} & \textbf{$\bar\tau$} & \textbf{$\boldsymbol{\alpha}$} & \textbf{Training firing rate} \\
     &  & PPL $\downarrow$ & \multicolumn{3}{c}{normalized entropy} & mean & 10th--90th percentile \\
    \midrule
    \textcolor{red}{\amormamba{}\amorheart{}} & 180M & 31.46 & 0.706 & 0.106 & 0.728 & 0.801 & 33.1--45.8\% \\
    \textcolor{red}{\amormamba{}\amorheart{}} & 440M & 19.60 & 0.847 & 0.057 & 0.858 & 0.810 & 34.1--47.6\% \\
    \textcolor{red}{\amormamba{}\amorheart{}} & 1.5B & 13.30 & 0.934 & 0.028 & 0.940 & 0.662 & 32.9--49.4\% \\
    \addlinespace[3pt]
    \textcolor{red}{\amorgdn{}\amorheart{}} & 180M & 30.60 & 0.707 & 0.108 & 0.728 & 0.779 & 33.3--45.5\% \\
    \textcolor{red}{\amorgdn{}\amorheart{}} & 440M & 19.27 & 0.853 & 0.053 & 0.864 & 0.613 & 33.9--46.9\% \\
    \textcolor{red}{\amorgdn{}\amorheart{}} & 1.5B & 13.10 & 0.937 & 0.030 & 0.943 & 0.544 & 32.6--50.6\% \\
    \bottomrule
  \end{tabular}
\end{table}

\FloatBarrier
\Needspace{5\baselineskip}
\subsection{Held-out gating behavior}
For \amor{} variants pretrained with a target firing rate of ${\sim}40\%$ at 1.5B, observed firing rates during inference prefill on held-out text vary by domain but remain similar from 3K to 20K within each dataset (training context: 3K). Rates are 37--43\% on FineWeb-Edu~\citep{fineweb_edu}, 64--69\% on PG-19~\citep{pg19}, and 97--99\% on CodeParrot~\citep{codeparrot} (Table~\ref{tab:ablation_amor_fire}). During generative decoding from a 3K held-out FineWeb-Edu prompt, firing rates range from 18--31\% across greedy and sampled decoding with the native entropy gate or distilled router (Table~\ref{tab:decode_3k}).
\begin{table}[t]
  \centering
  \footnotesize
  \setlength{\tabcolsep}{5pt}
  \caption{\textbf{Prefill firing varies more by domain than context length.} 1.5B; firing rates average all positions and three blocks over six text segments per context length.}
  \label{tab:ablation_amor_fire}
  \begin{tabular}{llcccc}
    \toprule
    \multicolumn{6}{l}{\textbf{Prefill firing rate on held-out text}} \\
    \midrule
    \textbf{Model} & \textbf{Dataset} & 3K & 4K & 8K & 20K \\
    \midrule
    \textcolor{red}{\amormamba{}\amorheart{}} & FineWeb-Edu & 38.0\% & 37.6\% & 38.0\% & 37.2\% \\
     & PG-19 & 64.1\% & 66.4\% & 67.4\% & 64.9\% \\
     & CodeParrot & 98.9\% & 98.9\% & 98.3\% & 97.1\% \\
    \addlinespace[3pt]
    \textcolor{red}{\amorgdn{}\amorheart{}} & FineWeb-Edu & 42.7\% & 41.8\% & 40.5\% & 40.8\% \\
     & PG-19 & 66.9\% & 69.1\% & 68.1\% & 64.4\% \\
     & CodeParrot & 99.4\% & 99.4\% & 99.0\% & 98.2\% \\
    \bottomrule
  \end{tabular}
\end{table}

\begingroup

\FloatBarrier
\Needspace{5\baselineskip}
\subsection{Additional gate-fire examples}
\label{app:gate-traces}

Figure~\ref{fig:gate_more} shows four selected prompts. Individual blocks fire on 27--74\% of the displayed tokens, including rare subword pieces, numbers and names, while remaining closed on locally predictable continuations.

\begin{figure}[h]
  \centering
  \begin{minipage}{0.5\textwidth}
    \includegraphics[width=\linewidth]{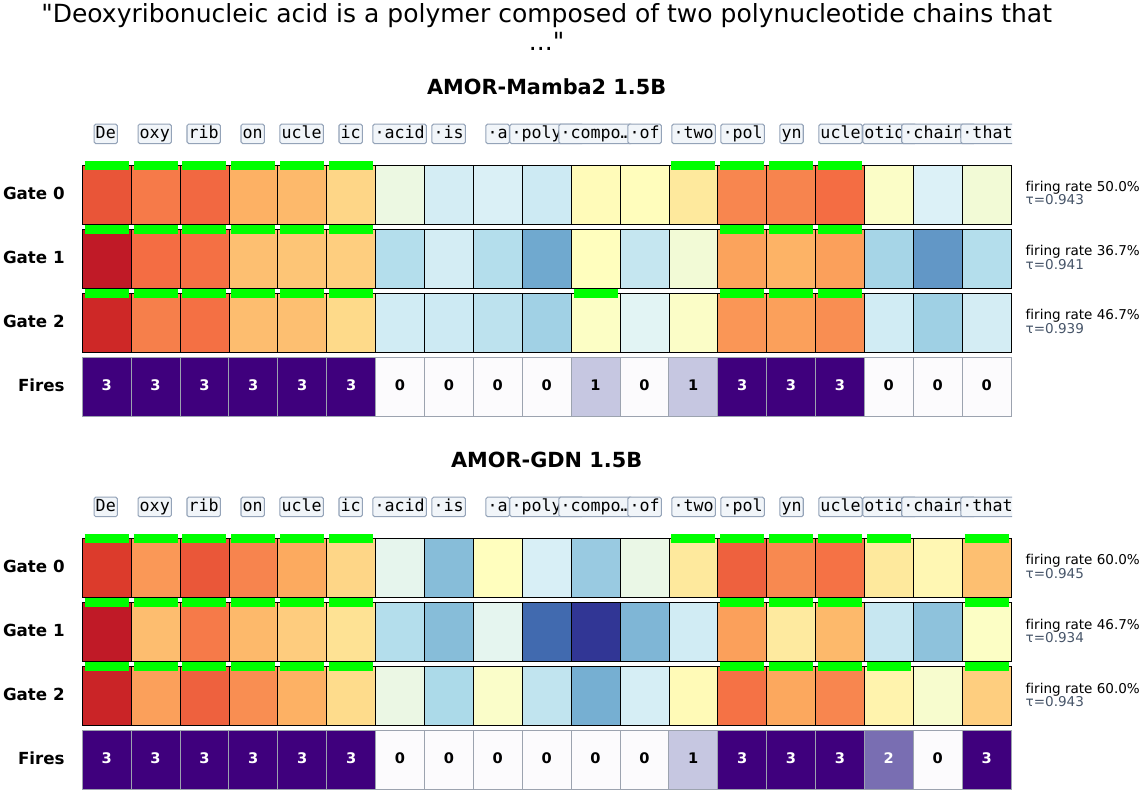}
  \end{minipage}\hfill
  \begin{minipage}{0.5\textwidth}
    \includegraphics[width=\linewidth]{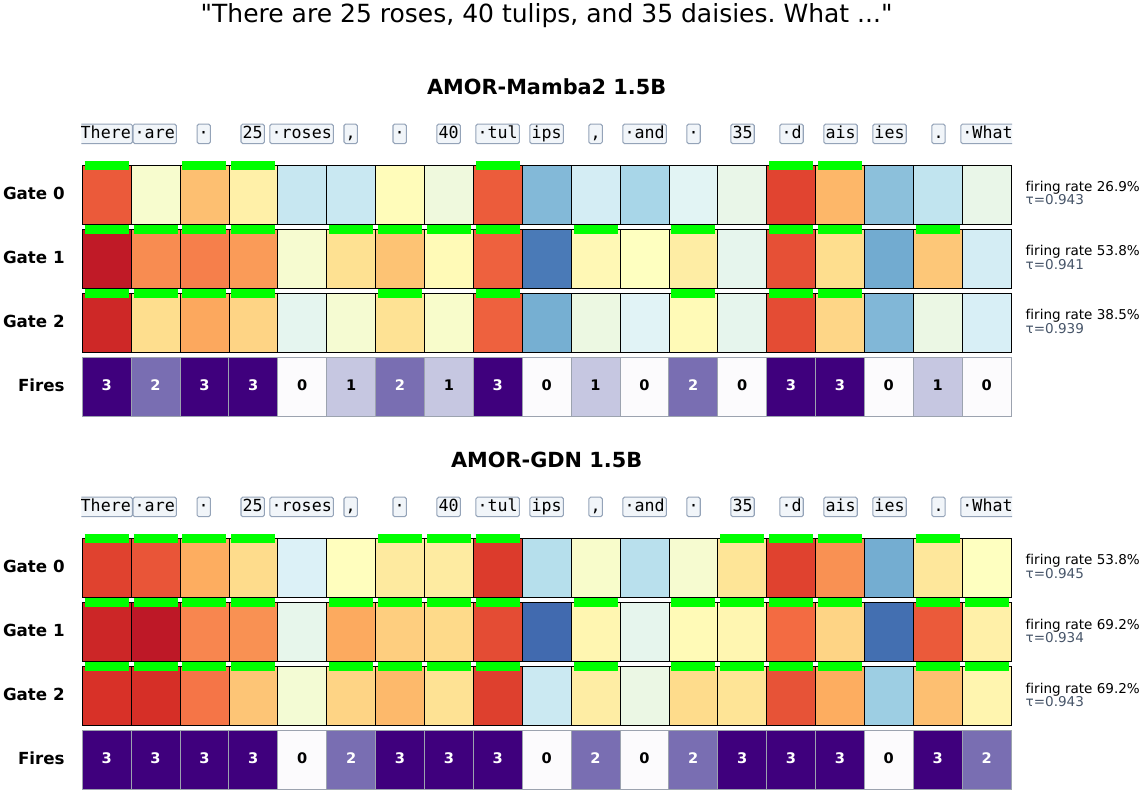}
  \end{minipage}\\[6pt]
  \begin{minipage}{0.5\textwidth}
    \includegraphics[width=\linewidth]{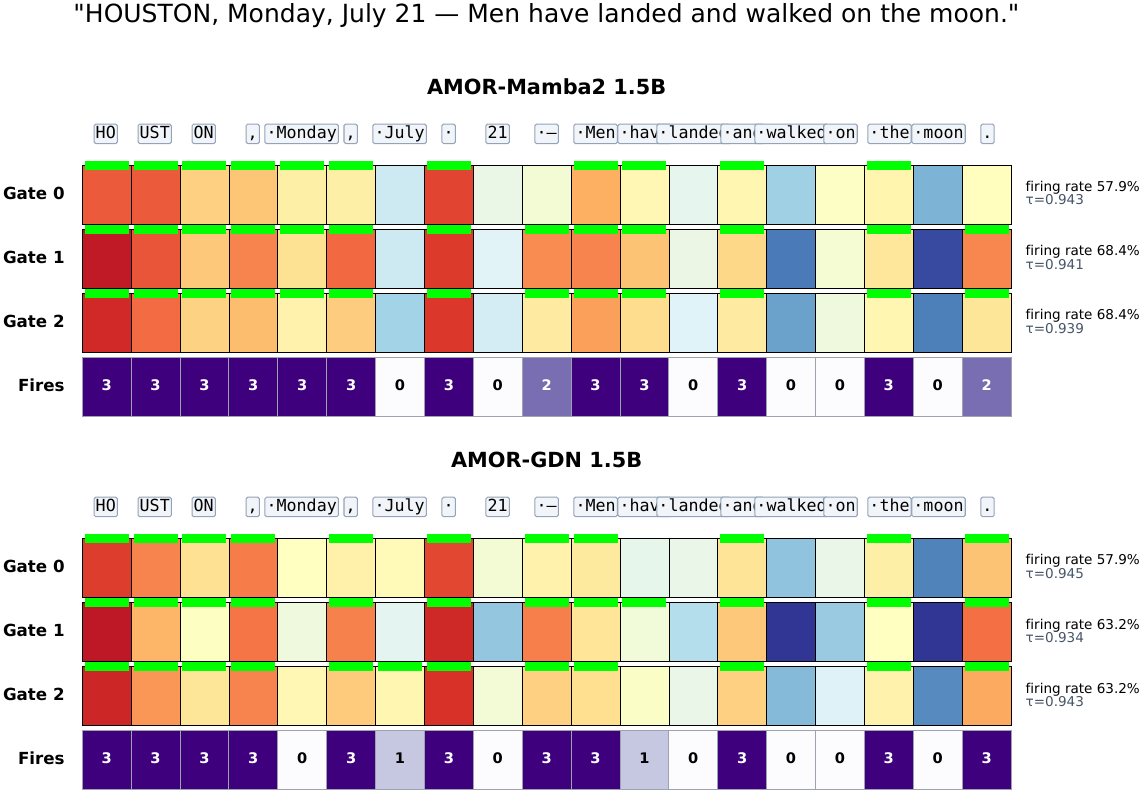}
  \end{minipage}\hfill
  \begin{minipage}{0.5\textwidth}
    \includegraphics[width=\linewidth]{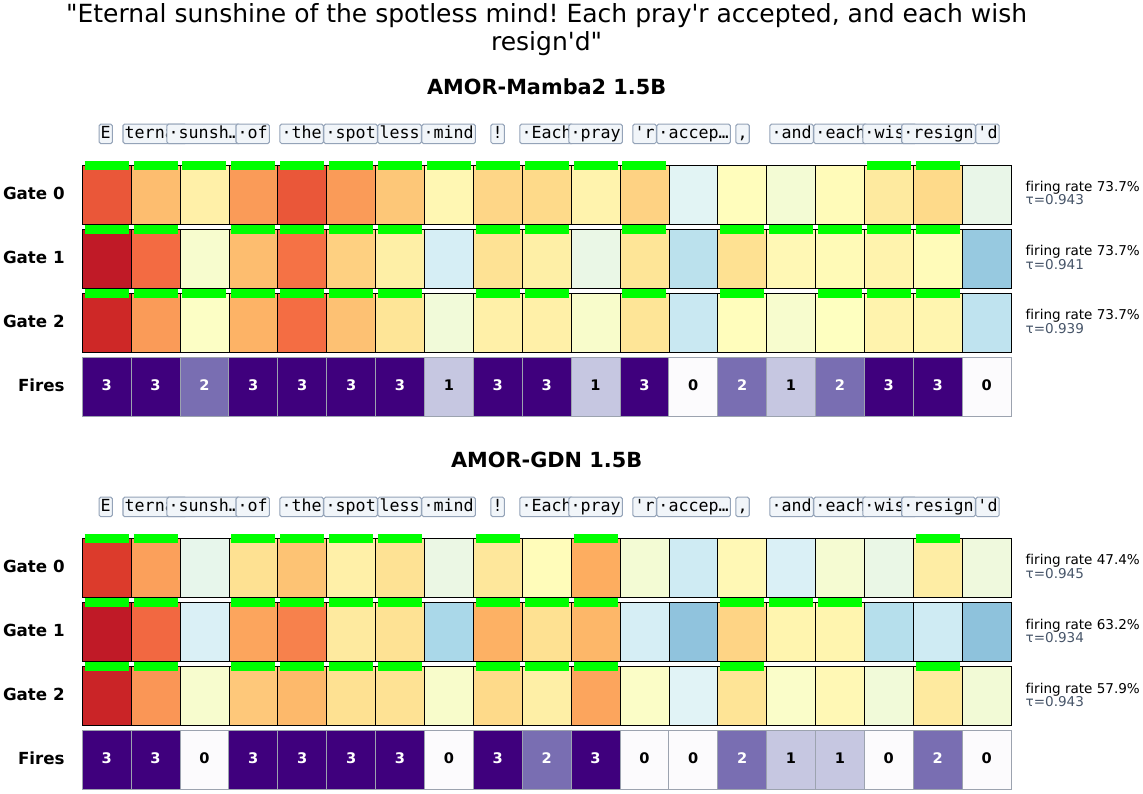}
  \end{minipage}
  \caption{\textbf{Gate patterns vary across domains.} 1.5B; native entropy gates for \amormamba{} and \amorgdn{} on DNA, mathematical reasoning, news and poetry prompts. Colors follow Fig.~\ref{fig:gate_s0_teaser}.}
  \label{fig:gate_more}
\end{figure}
\endgroup

\FloatBarrier
\Needspace{5\baselineskip}
\section{Evaluation Setup}
\label{app:eval}

Downstream benchmark evaluations run via the LM Evaluation Harness~\citep{lm_eval_harness}; loglikelihood tasks use the harness defaults, and generation tasks use greedy decoding.

\FloatBarrier
\Needspace{5\baselineskip}
\subsection{Common-Sense Reasoning}
\label{app:eval:common}

We evaluate all models on eight common-sense zero-shot loglikelihood tasks across 180M, 440M, and 1.5B scales. The tasks are: LAMBADA~\citep{lambada} (last-word prediction in narrative passages), HellaSwag~\citep{hellaswag} (multiple-choice sentence completion stressing physical and temporal commonsense), PIQA~\citep{piqa} (physical-interaction QA), ARC-Easy and ARC-Challenge~\citep{arc} (grade-school science multiple-choice), WinoGrande~\citep{winogrande} (pronoun-disambiguation Winograd schemas), OpenBookQA~\citep{openbookqa} (open-book elementary-science QA), and TruthfulQA-mc2~\citep{truthfulqa} (multiple-choice questions whose distractors are popular misconceptions).

Every common-sense accuracy score in this paper, in Tables~\ref{tab:common_sense}, \ref{tab:iclr-seeds}, \ref{tab:ablation_440m}, \ref{tab:common_sense_amor_h_1p5b}, \ref{tab:vs_classic} and~\ref{tab:iclr-placement} and in Fig.~\ref{fig:iclr-common-a}, is the harness's raw accuracy (\texttt{acc}).

\FloatBarrier
\Needspace{5\baselineskip}
\subsection{In-Context Retrieval}
\label{app:eval:retrieval}

Following Gated DeltaNet and Mamba3~\citep{gdn,mamba3}, we evaluate all eight models at 1.5B for retrieval, varying context length.

Retrieval evaluation comprises two task suites. The first uses the six-task cloze-completion protocol of \citet{just_read_twice}, truncated to a 2K context window. It includes SWDE~\citep{swde} for structured HTML relation extraction, FDA~\citep{fda} for PDF key–value retrieval, SQuAD-completion~\citep{squad} for short-passage extractive question answering, and cloze-formatted variants of TriviaQA~\citep{triviaqa}, Natural Questions~\citep{nq}, and DROP~\citep{drop}. The second suite is Single Needle-in-a-Haystack (S-NIAH), drawn from RULER~\citep{ruler} and evaluated at context lengths of 1K, 2K, and 4K. It consists of S-NIAH-1 (passkey retrieval in a repeated synthetic haystack), S-NIAH-2 (numeric needle embedded in a real-essay haystack), and S-NIAH-3 (UUID needle embedded in a real-essay haystack).

\FloatBarrier
\Needspace{5\baselineskip}
\subsection{Long-Context Behavior}
\label{app:eval:longctx}
\label{app:length-extrap}

Long-context comparisons follow the same convention as Gated DeltaNet and Mamba3~\citep{gdn,mamba3}, using 1.5B models across varying context lengths.

\subsubsection{LongBench}
We evaluate 14 tasks from \textbf{LongBench}~\citep{longbench}: single-document QA on NarrativeQA~\citep{narrativeqa}, Qasper~\citep{qasper}, and MultiFieldQA-en; multi-document QA on HotpotQA~\citep{hotpotqa}, 2WikiMultihopQA~\citep{2wikimqa}, and MuSiQue~\citep{musique}; summarization on GovReport~\citep{govreport}, QMSum~\citep{qmsum}, and MultiNews~\citep{multinews}; few-shot in-context learning on TREC~\citep{trec}, TriviaQA~\citep{triviaqa}, and SAMSum~\citep{samsum}; and code completion on LCC~\citep{lcc} and RepoBench-P~\citep{repobench}.

\FloatBarrier
\Needspace{5\baselineskip}
\subsubsection{Length Extrapolation}
\label{app:eval:extrapolation}
We set out to explore per-token perplexity as a function of context length, to assess the model's length extrapolation capability beyond its training distribution (see Fig.~\ref{fig:length_extrap}). We computed the per-token perplexity at ten lengths ranging from $3$K (the training context) to $20$K tokens, using $50$ segments per length per dataset. We follow the six benchmarks used in \citet{gdn} for length extrapolation: GovReport~\citep{govreport} (long-form government reports), QMSum~\citep{qmsum} (query-based meeting summarization), NarrativeQA~\citep{narrativeqa} (literary-passage QA), Qasper~\citep{qasper} (research-paper QA), CodeParrot~\citep{codeparrot} (GitHub-Python code), and PG-19~\citep{pg19} (long-form public-domain books).

\FloatBarrier
\Needspace{5\baselineskip}
\section{\texorpdfstring{Efficiency Measurements}{Efficiency Measurements}}
\label{app:bench}

\FloatBarrier
\Needspace{5\baselineskip}
\subsection{Training Throughput}
\begingroup

\amor{} reaches 85--87\% of its recurrent backbone's throughput; \amormamba{} is $1.4\times$ faster than the Mamba2 fused hybrid. Forward FLOPs include each \amor{} block's LM-head evaluation.

\begin{table}[h]
\centering
\footnotesize
\setlength{\tabcolsep}{5pt}
\caption{\textbf{\amor{} retains most recurrent-model training throughput.} 1.5B; one H100 NVL, 100 training steps with microbatch 2 and 80 accumulation steps.}
\label{tab:bench_summary}
\begin{tabular}{lrrr}
    \toprule
    \textbf{Model} & \textbf{Fwd FLOPs} & \textbf{Throughput} & \textbf{Peak training memory} \\
     & GFLOPs/tok $\downarrow$ & thousand tok/s $\uparrow$ & GiB $\downarrow$ \\
    \midrule
    Transformer & 3.15 & 23.59 & 44.35 \\
    Mamba2 & 2.60 & 25.09 & 48.07 \\
    Gated DeltaNet & 2.95 & 22.82 & 53.60 \\
    \addlinespace[3pt]
    Mamba2 Serial Hybrid & 2.67 & 24.61 & 47.61 \\
    Gated DeltaNet Serial Hybrid & 2.97 & 23.36 & 52.45 \\
    Mamba2 Fused Hybrid & 2.70 & 15.31 & 49.52 \\
    \addlinespace[3pt]
    \textcolor{red}{\amormamba{}\amorheart{}} & 4.35 & 21.46 & 49.63 \\
    \textcolor{red}{\amorgdn{}\amorheart{}} & 4.70 & 19.83 & 55.16 \\
    \bottomrule
\end{tabular}
\end{table}

\FloatBarrier
\Needspace{5\baselineskip}
\subsection{Short-Context Decoding Throughput}

With batch size 1, each trial prefills the same 3,072-token prompt from held-out FineWeb-Edu text, then generates 256 tokens. We report throughput from the mean decoding time over three trials, following a separate 32-token warm-up.

\begin{table}[h]
\centering
\footnotesize
\setlength{\tabcolsep}{5pt}
\caption{\textbf{At 3K, \amor{} tracks recurrent-model decode throughput.} 1.5B; one H100 NVL. Firing rates average generated positions and three blocks. $T$ denotes sampling temperature; N/A denotes no gate.}
\label{tab:decode_3k}
\begin{tabular}{lrrrr}
    \toprule
    \textbf{Model} & \multicolumn{2}{c}{\textbf{Greedy}, $T{=}0$} & \multicolumn{2}{c}{\textbf{Sampling}, $T{=}1$} \\
    \cmidrule(lr){2-3}\cmidrule(lr){4-5}
     & tok/s $\uparrow$ & firing rate & tok/s $\uparrow$ & firing rate \\
    \midrule
    Transformer & 102.0 & N/A & 100.5 & N/A \\
    Mamba2 & 63.8 & N/A & 64.6 & N/A \\
    Gated DeltaNet & 44.6 & N/A & 44.4 & N/A \\
    \addlinespace[3pt]
    Mamba2 Serial Hybrid & 64.5 & N/A & 64.7 & N/A \\
    Gated DeltaNet Serial Hybrid & 48.6 & N/A & 48.1 & N/A \\
    Mamba2 Fused Hybrid & 60.1 & N/A & 59.8 & N/A \\
    \addlinespace[3pt]
    \textcolor{red}{\amormamba{}\amorheart{}} & 57.8 & 20.8\% & 57.9 & 24.2\% \\
    \quad + distilled router & 58.3 & 20.6\% & 58.0 & 17.6\% \\
    \textcolor{red}{\amorgdn{}\amorheart{}} & 41.7 & 30.1\% & 41.9 & 19.4\% \\
    \quad + distilled router & 41.9 & 30.7\% & 41.4 & 20.6\% \\
    \bottomrule
\end{tabular}
\end{table}

\endgroup

\clearpage
\begingroup
\setlength{\intextsep}{4pt}
\subsection{Long-context inference efficiency}
\label{app:bench:empirical}
\label{app:iclr-efficiency}
\label{app:iclr-eff-scales}

\begin{table}[H]

\centering
\scriptsize
\captionsetup{font=footnotesize,skip=4pt}
\renewcommand{\arraystretch}{0.88}
\setlength{\aboverulesep}{1pt}
\setlength{\belowrulesep}{1pt}
\setlength{\tabcolsep}{4pt}
\caption{\textbf{Inference efficiency.} 1.5B; one H100 PCIe, batch 1, memory-efficient SDPA. Prefill latency and peak prefill memory use one warmed measurement with final-position logits. Decode latency averages three 32-token runs after eight warm-up tokens. OOM denotes out of memory.}
\label{tab:iclr-runtime-values}
\label{tab:iclr-native-distilled}
\begin{tabular}{lrrrrrr}
\toprule
\textbf{Model} & \multicolumn{6}{c}{\textbf{Context length}} \\
\cmidrule(lr){2-7}
 & 8K & 16K & 32K & 64K & 96K & 128K \\
\midrule
\multicolumn{7}{@{}l}{\textbf{Prefill latency (s) $\downarrow$}} \\
Transformer & 0.15 & 0.44 & 1.41 & 4.99 & 10.83 & OOM \\
Mamba2 & 0.13 & 0.24 & 0.48 & 0.95 & 1.43 & 1.92 \\
Gated DeltaNet & 0.11 & 0.20 & 0.40 & 0.80 & 1.21 & 1.61 \\
\addlinespace[1pt]
Mamba2 Serial Hybrid & 0.13 & 0.27 & 0.61 & 1.49 & 2.69 & 4.17 \\
Gated DeltaNet Serial Hybrid & 0.11 & 0.24 & 0.54 & 1.37 & 2.49 & 3.89 \\
Mamba2 Fused Hybrid & 0.24 & 0.48 & 1.03 & 2.37 & 3.99 & 5.77 \\
\addlinespace[1pt]
\textcolor{red}{\amormamba{}\amorheart{}} & 0.26 & 0.53 & 1.13 & 2.53 & 4.27 & 6.30 \\
\quad + distilled router & 0.14 & 0.29 & 0.66 & 1.58 & 2.85 & 4.40 \\
\textcolor{red}{\amorgdn{}\amorheart{}} & 0.24 & 0.49 & 1.05 & 2.40 & 4.03 & 5.96 \\
\quad + distilled router & 0.12 & 0.26 & 0.58 & 1.43 & 2.58 & 4.01 \\
\midrule
\multicolumn{7}{@{}l}{\textbf{Peak prefill memory (GB) $\downarrow$}} \\
Transformer & 4.75 & 6.80 & 10.89 & 19.08 & 27.27 & OOM \\
Mamba2 & 3.85 & 4.67 & 6.30 & 9.55 & 12.81 & 16.07 \\
Gated DeltaNet & 3.54 & 4.08 & 5.16 & 7.32 & 9.49 & 11.65 \\
\addlinespace[1pt]
Mamba2 Serial Hybrid & 4.13 & 5.25 & 7.48 & 11.95 & 16.42 & 20.88 \\
Gated DeltaNet Serial Hybrid & 3.79 & 4.60 & 6.22 & 9.45 & 12.69 & 15.92 \\
Mamba2 Fused Hybrid & 4.36 & 5.61 & 8.11 & 13.12 & 18.12 & 23.12 \\
\addlinespace[1pt]
\textcolor{red}{\amormamba{}\amorheart{}} & 13.06 & 13.43 & 14.17 & 15.65 & 17.12 & 18.60 \\
\quad + distilled router & 4.06 & 4.88 & 6.51 & 9.77 & 13.03 & 16.29 \\
\textcolor{red}{\amorgdn{}\amorheart{}} & 13.06 & 13.46 & 14.27 & 15.88 & 17.49 & 19.10 \\
\quad + distilled router & 3.82 & 4.42 & 5.63 & 8.05 & 10.46 & 12.88 \\
\midrule
\multicolumn{7}{@{}l}{\textbf{Decode latency (ms/token) $\downarrow$}} \\
Transformer & 17.36 & 29.37 & 53.28 & 100.75 & 148.61 & OOM \\
Mamba2 & 16.24 & 16.26 & 16.13 & 16.14 & 16.12 & 16.15 \\
Gated DeltaNet & 22.57 & 22.76 & 22.63 & 22.56 & 22.59 & 22.45 \\
\addlinespace[1pt]
Mamba2 Serial Hybrid & 16.58 & 18.12 & 21.21 & 27.30 & 33.35 & 39.42 \\
Gated DeltaNet Serial Hybrid & 21.13 & 21.03 & 21.01 & 20.95 & 23.99 & 29.98 \\
Mamba2 Fused Hybrid & 18.19 & 19.78 & 22.80 & 28.92 & 34.97 & 41.07 \\
\addlinespace[1pt]
\textcolor{red}{\amormamba{}\amorheart{}} & 19.22 & 19.48 & 22.67 & 26.81 & 32.00 & 34.21 \\
\quad + distilled router & 18.82 & 18.90 & 22.11 & 26.32 & 31.27 & 33.73 \\
\textcolor{red}{\amorgdn{}\amorheart{}} & 25.94 & 26.10 & 27.53 & 31.90 & 37.57 & 42.08 \\
\quad + distilled router & 25.07 & 25.34 & 27.08 & 31.26 & 36.46 & 41.57 \\
\bottomrule
\end{tabular}
\end{table}

\begin{figure}[H]
  \centering
  \captionsetup{font=footnotesize,skip=4pt}
  \includegraphics[width=\textwidth]{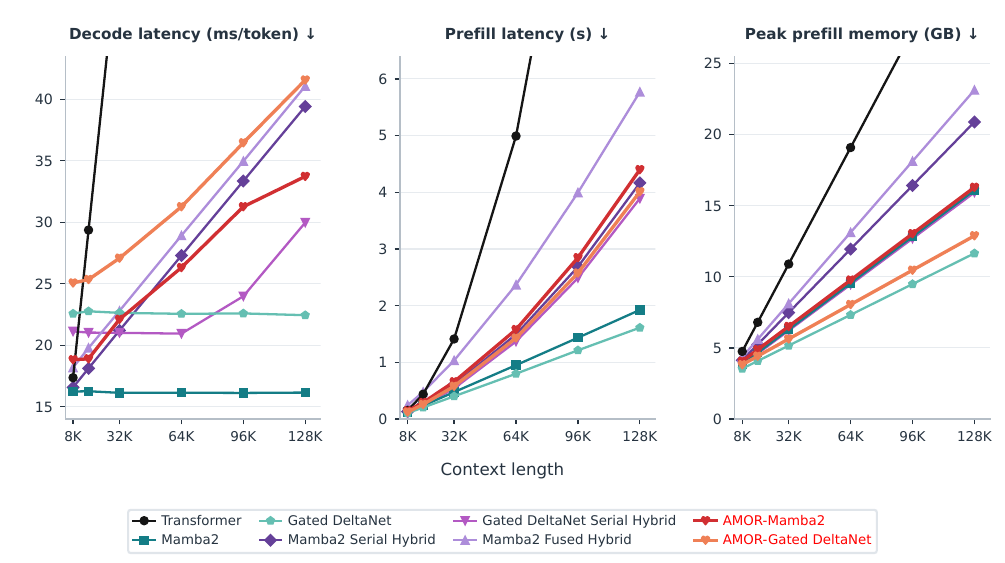}
  \caption{\textbf{Long-context inference efficiency.} 1.5B; all eight models on one H100 PCIe. \amor{} uses the distilled router. Transformer curves exceed the axis ranges and reach OOM at 128K.}
  \label{fig:iclr-eff-all}
\end{figure}

\clearpage
Tables~\ref{tab:iclr-runtime-values}--\ref{tab:iclr-runtime-440m} report inference efficiency for \amor{} with the native entropy gate and the distilled router alongside all baselines at three scales. The distilled router lowers prefill latency and peak prefill memory across scales, while decode gains vary by backbone and context length. In decode latency, \amormamba{} with the distilled router overtakes the fused hybrid between 8K and 16K at every scale. Its crossover with the serial hybrid occurs at 64K--96K (180M), 16K--32K (440M) and 32K--64K (1.5B).

\begin{table}[H]

\centering
\scriptsize
\captionsetup{font=footnotesize,skip=4pt}
\renewcommand{\arraystretch}{0.88}
\setlength{\aboverulesep}{1pt}
\setlength{\belowrulesep}{1pt}
\setlength{\tabcolsep}{4pt}
\caption{\textbf{Inference efficiency.} 180M; all eight architectures on one H100 PCIe, using the protocol of Table~\ref{tab:iclr-runtime-values}.}
\label{tab:iclr-runtime-180m}
\begin{tabular}{lrrrrrr}
\toprule
\textbf{Model} & \multicolumn{6}{c}{\textbf{Context length}} \\
\cmidrule(lr){2-7}
 & 8K & 16K & 32K & 64K & 96K & 128K \\
\midrule
\multicolumn{7}{@{}l}{\textbf{Prefill latency (s) $\downarrow$}} \\
Transformer & 0.03 & 0.08 & 0.25 & 0.90 & 1.94 & 3.48 \\
Mamba2 & 0.03 & 0.04 & 0.07 & 0.14 & 0.21 & 0.28 \\
Gated DeltaNet & 0.03 & 0.04 & 0.07 & 0.13 & 0.19 & 0.25 \\
\addlinespace[1pt]
Mamba2 Serial Hybrid & 0.03 & 0.05 & 0.11 & 0.27 & 0.51 & 0.81 \\
Gated DeltaNet Serial Hybrid & 0.02 & 0.04 & 0.10 & 0.26 & 0.49 & 0.80 \\
Mamba2 Fused Hybrid & 0.03 & 0.06 & 0.14 & 0.37 & 0.70 & 1.16 \\
\addlinespace[1pt]
\textcolor{red}{\amormamba{}\amorheart{}} & 0.14 & 0.27 & 0.55 & 1.20 & 1.96 & 2.84 \\
\quad + distilled router & 0.03 & 0.06 & 0.14 & 0.38 & 0.72 & 1.17 \\
\textcolor{red}{\amorgdn{}\amorheart{}} & 0.14 & 0.26 & 0.54 & 1.19 & 1.94 & 2.80 \\
\quad + distilled router & 0.03 & 0.06 & 0.13 & 0.36 & 0.70 & 1.14 \\
\midrule
\multicolumn{7}{@{}l}{\textbf{Peak prefill memory (GB) $\downarrow$}} \\
Transformer & 0.84 & 1.29 & 2.19 & 3.99 & 5.79 & 7.59 \\
Mamba2 & 0.71 & 1.02 & 1.64 & 2.89 & 4.13 & 5.37 \\
Gated DeltaNet & 0.59 & 0.79 & 1.20 & 2.01 & 2.82 & 3.63 \\
\addlinespace[1pt]
Mamba2 Serial Hybrid & 0.83 & 1.23 & 2.03 & 3.62 & 5.22 & 6.81 \\
Gated DeltaNet Serial Hybrid & 0.70 & 0.98 & 1.53 & 2.65 & 3.76 & 4.87 \\
Mamba2 Fused Hybrid & 0.92 & 1.39 & 2.34 & 4.24 & 6.14 & 8.03 \\
\addlinespace[1pt]
\textcolor{red}{\amormamba{}\amorheart{}} & 10.11 & 10.25 & 10.53 & 11.08 & 11.64 & 12.19 \\
\quad + distilled router & 0.83 & 1.14 & 1.77 & 3.01 & 4.25 & 5.50 \\
\textcolor{red}{\amorgdn{}\amorheart{}} & 10.12 & 10.27 & 10.57 & 11.18 & 11.78 & 12.39 \\
\quad + distilled router & 0.74 & 0.97 & 1.42 & 2.33 & 3.23 & 4.14 \\
\midrule
\multicolumn{7}{@{}l}{\textbf{Decode latency (ms/token) $\downarrow$}} \\
Transformer & 7.13 & 12.40 & 22.50 & 42.83 & 62.50 & 80.44 \\
Mamba2 & 8.08 & 8.17 & 8.09 & 8.11 & 8.14 & 8.10 \\
Gated DeltaNet & 11.63 & 11.59 & 11.55 & 11.62 & 11.57 & 11.59 \\
\addlinespace[1pt]
Mamba2 Serial Hybrid & 8.05 & 8.97 & 10.75 & 14.25 & 17.70 & 21.19 \\
Gated DeltaNet Serial Hybrid & 10.53 & 10.52 & 10.51 & 10.53 & 12.54 & 15.97 \\
Mamba2 Fused Hybrid & 9.68 & 11.10 & 13.70 & 19.00 & 24.20 & 29.02 \\
\addlinespace[1pt]
\textcolor{red}{\amormamba{}\amorheart{}} & 10.50 & 10.55 & 12.65 & 14.43 & 20.38 & 21.85 \\
\quad + distilled router & 10.19 & 10.14 & 11.89 & 15.22 & 16.80 & 18.16 \\
\textcolor{red}{\amorgdn{}\amorheart{}} & 13.97 & 13.94 & 15.15 & 16.52 & 18.82 & 19.98 \\
\quad + distilled router & 13.73 & 13.88 & 15.88 & 16.41 & 17.87 & 19.99 \\
\bottomrule
\end{tabular}
\end{table}

\begin{table}[H]

\centering
\scriptsize
\captionsetup{font=footnotesize,skip=4pt}
\renewcommand{\arraystretch}{0.88}
\setlength{\aboverulesep}{1pt}
\setlength{\belowrulesep}{1pt}
\setlength{\tabcolsep}{4pt}
\caption{\textbf{Inference efficiency.} 440M; all eight architectures on one H100 PCIe, using the protocol of Table~\ref{tab:iclr-runtime-values}.}
\label{tab:iclr-runtime-440m}
\begin{tabular}{lrrrrrr}
\toprule
\textbf{Model} & \multicolumn{6}{c}{\textbf{Context length}} \\
\cmidrule(lr){2-7}
 & 8K & 16K & 32K & 64K & 96K & 128K \\
\midrule
\multicolumn{7}{@{}l}{\textbf{Prefill latency (s) $\downarrow$}} \\
Transformer & 0.07 & 0.21 & 0.70 & 2.52 & 5.49 & 9.56 \\
Mamba2 & 0.07 & 0.11 & 0.20 & 0.45 & 0.60 & 0.80 \\
Gated DeltaNet & 0.05 & 0.09 & 0.17 & 0.34 & 0.51 & 0.68 \\
\addlinespace[1pt]
Mamba2 Serial Hybrid & 0.06 & 0.12 & 0.26 & 0.66 & 1.21 & 1.88 \\
Gated DeltaNet Serial Hybrid & 0.05 & 0.11 & 0.24 & 0.61 & 1.13 & 1.78 \\
Mamba2 Fused Hybrid & 0.10 & 0.18 & 0.38 & 0.90 & 1.55 & 2.35 \\
\addlinespace[1pt]
\textcolor{red}{\amormamba{}\amorheart{}} & 0.18 & 0.35 & 0.71 & 1.56 & 2.56 & 3.68 \\
\quad + distilled router & 0.07 & 0.13 & 0.29 & 0.71 & 1.29 & 1.99 \\
\textcolor{red}{\amorgdn{}\amorheart{}} & 0.16 & 0.33 & 0.68 & 1.51 & 2.48 & 3.58 \\
\quad + distilled router & 0.06 & 0.12 & 0.26 & 0.65 & 1.20 & 1.88 \\
\midrule
\multicolumn{7}{@{}l}{\textbf{Peak prefill memory (GB) $\downarrow$}} \\
Transformer & 1.85 & 2.87 & 4.91 & 9.00 & 13.08 & 17.16 \\
Mamba2 & 1.36 & 1.77 & 2.59 & 4.24 & 5.89 & 7.53 \\
Gated DeltaNet & 1.19 & 1.46 & 2.00 & 3.08 & 4.16 & 5.24 \\
\addlinespace[1pt]
Mamba2 Serial Hybrid & 1.52 & 2.09 & 3.21 & 5.46 & 7.71 & 9.96 \\
Gated DeltaNet Serial Hybrid & 1.34 & 1.74 & 2.55 & 4.17 & 5.79 & 7.41 \\
Mamba2 Fused Hybrid & 1.60 & 2.23 & 3.49 & 6.01 & 8.53 & 11.05 \\
\addlinespace[1pt]
\textcolor{red}{\amormamba{}\amorheart{}} & 10.71 & 10.89 & 11.26 & 12.00 & 12.74 & 13.48 \\
\quad + distilled router & 1.49 & 1.90 & 2.72 & 4.37 & 6.02 & 7.66 \\
\textcolor{red}{\amorgdn{}\amorheart{}} & 10.71 & 10.91 & 11.31 & 12.12 & 12.93 & 13.73 \\
\quad + distilled router & 1.36 & 1.66 & 2.26 & 3.47 & 4.68 & 5.89 \\
\midrule
\multicolumn{7}{@{}l}{\textbf{Decode latency (ms/token) $\downarrow$}} \\
Transformer & 14.87 & 25.63 & 46.99 & 89.82 & 132.80 & 175.66 \\
Mamba2 & 16.07 & 16.08 & 16.16 & 16.17 & 16.17 & 16.16 \\
Gated DeltaNet & 22.53 & 22.61 & 22.51 & 22.51 & 22.46 & 22.44 \\
\addlinespace[1pt]
Mamba2 Serial Hybrid & 15.88 & 17.35 & 20.16 & 25.38 & 31.03 & 36.49 \\
Gated DeltaNet Serial Hybrid & 21.34 & 21.47 & 21.46 & 21.51 & 21.58 & 25.80 \\
Mamba2 Fused Hybrid & 17.42 & 18.87 & 21.64 & 27.11 & 32.50 & 37.99 \\
\addlinespace[1pt]
\textcolor{red}{\amormamba{}\amorheart{}} & 18.33 & 18.03 & 19.12 & 25.98 & 22.85 & 27.10 \\
\quad + distilled router & 17.98 & 17.69 & 18.61 & 25.15 & 21.37 & 26.62 \\
\textcolor{red}{\amorgdn{}\amorheart{}} & 24.65 & 24.87 & 27.33 & 28.43 & 30.62 & 37.13 \\
\quad + distilled router & 24.68 & 24.96 & 26.57 & 28.76 & 30.46 & 36.97 \\
\bottomrule
\end{tabular}
\end{table}

\endgroup

\FloatBarrier
\Needspace{5\baselineskip}
\subsection{Memory usage}
At 1.5B, three-attention-layer hybrids use one eighth of the Transformer's KV storage: 3.22 versus 25.77\,GB at 128K (Table~\ref{tab:iclr-decode-cache}). Every position contributes keys and values regardless of firing. Recurrent-state storage is fixed with context length: 26.00\,MB for 24 Mamba2 layers and 20.05\,MB for 24 Gated DeltaNet layers; serial hybrids use 21 recurrent layers.

\begin{table}[h]
\centering
\footnotesize
\setlength{\tabcolsep}{4pt}
\caption{\textbf{KV cache and recurrent-state storage.} 1.5B; bfloat16 caches and three attention layers per hybrid. N/A denotes an unused cache or state.}
\label{tab:iclr-decode-cache}
  \begingroup
\begin{tabular}{lrrrrrr}
    \toprule
    \textbf{Model} & \textbf{Recurrent state size} & \multicolumn{5}{c}{\textbf{Key-value cache} (GB) $\downarrow$} \\
    \cmidrule(lr){3-7}
     & MB, fixed & 3K & 8K & 16K & 32K & 128K \\
    \midrule
    Transformer & N/A & 0.60 & 1.61 & 3.22 & 6.44 & 25.77 \\
    Mamba2 & 26.00 & N/A & N/A & N/A & N/A & N/A \\
    Gated DeltaNet & 20.05 & N/A & N/A & N/A & N/A & N/A \\
    \addlinespace[3pt]
    Mamba2 Serial Hybrid & 22.75 & 0.08 & 0.20 & 0.40 & 0.81 & 3.22 \\
    Gated DeltaNet Serial Hybrid & 17.54 & 0.08 & 0.20 & 0.40 & 0.81 & 3.22 \\
    Mamba2 Fused Hybrid & 26.00 & 0.08 & 0.20 & 0.40 & 0.81 & 3.22 \\
    \addlinespace[3pt]
    \textcolor{red}{\amormamba{}\amorheart{}} & 26.00 & 0.08 & 0.20 & 0.40 & 0.81 & 3.22 \\
    \textcolor{red}{\amorgdn{}\amorheart{}} & 20.05 & 0.08 & 0.20 & 0.40 & 0.81 & 3.22 \\
    \bottomrule
\end{tabular}
  \endgroup
\end{table}

\begingroup

\FloatBarrier
\Needspace{5\baselineskip}
\subsection{Decode Cost and Break-Even}
\label{app:bench:breakeven}

All costs below are matrix-multiply FLOPs per decoded token and per block, counting scalar multiplication and addition as one floating-point operation each. Let $D$ be model width, $V$ vocabulary size, $L$ the number of attended cache positions (including the current token), and $0\le f<1$ a fixed firing rate. Here $\mathbf A$ denotes the softmax attention weights.

\paragraph{Native entropy gate decode vs.\ always-on attention.} For otherwise identical attention blocks, \amor{} skips the following when $\gate_t=0$:
{\begin{itemize}\itemsep1pt
  \item \textbf{Attention scores $\mathbf Q\mathbf K^\top$:} $2DL$ FLOPs, summed across heads; the subsequent softmax is also skipped.
  \item \textbf{Weighted values $\mathbf A\mathbf V$:} $2DL$ FLOPs, summed across heads.
  \item \textbf{Query projection:} $2D^2$ FLOPs; RoPE on $\mathbf Q$ is also skipped.
  \item \textbf{Output projection $\mathbf W_O$:} $2D^2$ FLOPs.
\end{itemize}
}
Key/value projections and cache updates always run. The native entropy gate adds $C_{\mathrm{lm\_head}}=2VD$ matrix-multiply FLOPs.

\textit{Attention matmul savings only.} Counting the skipped $\mathbf Q\mathbf K^\top$ and $\mathbf A\mathbf V$ operations gives:
{\begin{equation}
\begin{aligned}
  C_{\mathrm{attn}}(L) &= 4DL, \\
  (1-f)\,C_{\mathrm{attn}}(L) &> C_{\mathrm{lm\_head}}, \\
  L &> \frac{V}{2(1-f)}.
\end{aligned}
\label{eq:decode_breakeven}
\end{equation}
}
\textit{Including query and output projection savings.} Adding their $4D^2$ FLOPs to the skipped cost gives:
{\begin{equation}
\begin{aligned}
  (1-f)(4DL+4D^2) &> C_{\mathrm{lm\_head}}, \\
  L &> \frac{V}{2(1-f)}-D.
\end{aligned}
\label{eq:decode_breakeven_projections}
\end{equation}
}
At $V=128{,}256$, $D=2048$ and an illustrative $f=0.4$, the equality thresholds are 106,880 tokens in Eq.~\ref{eq:decode_breakeven} and 104,832 in Eq.~\ref{eq:decode_breakeven_projections}. These are arithmetic thresholds, not measured latency crossovers.

\paragraph{Distilled router vs.\ native entropy gate.} The width-$r$ predictor replaces the vocabulary projection with three linear maps:
{\begin{itemize}\itemsep1pt
  \item Hidden projection, $D\to r$: $2Dr$ FLOPs.
  \item Scalar output, $r\to1$: $2r$ FLOPs.
  \item Linear skip, $D\to1$: $2D$ FLOPs.
\end{itemize}
}
For the evaluated 1.5B models, $D=2048$, $V=128{,}256$ and $r=512$:
{\begin{equation}
\begin{aligned}
  C_{\mathrm{route}} &= 2Dr+2r+2D = 2{,}102{,}272, \\
  C_{\mathrm{lm\_head}} &= 2VD = 525{,}336{,}576, \\
  \frac{C_{\mathrm{lm\_head}}}{C_{\mathrm{route}}} &\approx 250.
\end{aligned}
\label{eq:router_cost_ratio}
\end{equation}
}
This ratio measures gating matmul savings, not total model speedup. Both comparisons omit softmax, entropy reduction and other non-matmul work; measured latency also depends on memory traffic, kernels, synchronization and architecture (Appendix~\ref{app:iclr-efficiency}; Fig.~\ref{fig:iclr-eff-all}).
\endgroup

\FloatBarrier
\Needspace{5\baselineskip}
\section{Distilled Entropy Router}
\label{app:iclr-router}
\label{sec:method:router}
\FloatBarrier
\Needspace{5\baselineskip}
\subsection{Definition and fitting}

The deployment ablation replaces each block's LM-head evaluation with a predictor fitted on the frozen pretrained model:
{
\begin{equation}
\widehat{\entropy}_\ell(\tilde h_{t})
=a_\ell^\top\mathrm{SiLU}(W_\ell\tilde h_{t}+b_\ell)+s_\ell^\top\tilde h_{t}+c_\ell,
\qquad W_\ell\in\mathbb R^{512\times D}.
\label{eq:distilled_router}
\end{equation}
}
Here $W_\ell$ and $b_\ell$ parameterize the hidden projection, $a_\ell$ and $c_\ell$ the scalar output, and $s_\ell$ the linear skip. The gate remains $g_t=\mathbf1[\widehat{\entropy}_\ell(\tilde h_t)>\tau_\ell]$ with the frozen entropy threshold. Each predictor has a 512-unit hidden layer at all scales. The three predictors add 1.19M / 1.58M / 3.15M parameters at 180M / 440M / 1.5B.

At each scale, we fit the predictor for Block~0 first, then fit Blocks~1 and~2 with earlier predictors enabled. FineWeb-Edu, PG-19 and CodeParrot provide disjoint fitting, calibration and evaluation regions with about 1.57M, 0.295M and 0.590M positions per block, respectively. We initialize the linear skip with ridge regression, fit with boundary-weighted SmoothL1 loss for 30 epochs (learning rate $10^{-3}$, batch size 16,384), and calibrate the output bias on the held-out calibration split. Predictors run in float32 within the model's bfloat16 execution.

\FloatBarrier
\Needspace{5\baselineskip}
\subsection{Downstream quality}
Table~\ref{tab:iclr-router-quality} summarizes paired results with the native entropy gate and distilled router on the same frozen pretrained checkpoints. Across scales, common-sense averages change by at most 0.12 points and LongBench averages by less than 0.08 points. At 1.5B, the largest change across all four evaluation suites is 0.13 points.
Forward FLOPs are calculated for 3K inputs, replacing the three LM-head evaluation costs with predictor costs for the distilled router (Appendix~\ref{app:bench:breakeven}).

\begin{table}[H]
\centering
\footnotesize
\setlength{\tabcolsep}{4pt}
\caption{\textbf{Native entropy gate vs.\ distilled router: downstream performance and forward FLOPs.} 180M, 440M and 1.5B. Evaluation settings follow Appendix~\ref{app:eval}. $\Delta = \text{distilled router} - \text{native entropy gate}$.}
\label{tab:iclr-router-quality}
\resizebox{\textwidth}{!}{%
\begin{tabular}{llrrrrr}
\toprule
\textbf{Model} & \textbf{Gate} & \textbf{Common-sense} & \textbf{Cloze retrieval} & \textbf{S-NIAH} & \textbf{LongBench} & \textbf{Fwd FLOPs} \\
 & & avg acc (\%) $\uparrow$ & avg (\%) $\uparrow$ & avg acc (\%) $\uparrow$ & avg score (\%) $\uparrow$ & GFLOPs/tok $\downarrow$ \\
\midrule
\multicolumn{7}{@{}l}{\textbf{180M}} \\
\textcolor{red}{\amormamba{}\amorheart{}} & Native entropy gate & 35.82 & N/A & N/A & 5.89 & 0.96 \\
 & Distilled router & 35.84 & N/A & N/A & 5.91 & 0.38 \\
 & $\Delta$ & +0.02 & N/A & N/A & +0.02 & -0.59 \\
\addlinespace[3pt]
\textcolor{red}{\amorgdn{}\amorheart{}} & Native entropy gate & 35.67 & N/A & N/A & 6.12 & 0.98 \\
 & Distilled router & 35.62 & N/A & N/A & 6.05 & 0.39 \\
 & $\Delta$ & -0.05 & N/A & N/A & -0.07 & -0.59 \\
\addlinespace[3pt]
\midrule
\multicolumn{7}{@{}l}{\textbf{440M}} \\
\textcolor{red}{\amormamba{}\amorheart{}} & Native entropy gate & 39.58 & N/A & N/A & 6.76 & 1.63 \\
 & Distilled router & 39.55 & N/A & N/A & 6.83 & 0.85 \\
 & $\Delta$ & -0.03 & N/A & N/A & +0.06 & -0.78 \\
\addlinespace[3pt]
\textcolor{red}{\amorgdn{}\amorheart{}} & Native entropy gate & 40.45 & N/A & N/A & 6.95 & 1.71 \\
 & Distilled router & 40.56 & N/A & N/A & 7.00 & 0.93 \\
 & $\Delta$ & +0.10 & N/A & N/A & +0.05 & -0.78 \\
\addlinespace[3pt]
\midrule
\multicolumn{7}{@{}l}{\textbf{1.5B}} \\
\textcolor{red}{\amormamba{}\amorheart{}} & Native entropy gate & 45.81 & 27.13 & 74.11 & 8.27 & 4.35 \\
 & Distilled router & 45.76 & 27.04 & 74.16 & 8.25 & 2.78 \\
 & $\Delta$ & -0.06 & -0.09 & +0.04 & -0.02 & -1.57 \\
\addlinespace[3pt]
\textcolor{red}{\amorgdn{}\amorheart{}} & Native entropy gate & 45.13 & 30.01 & 77.78 & 8.44 & 4.70 \\
 & Distilled router & 45.01 & 30.14 & 77.67 & 8.43 & 3.13 \\
 & $\Delta$ & -0.12 & +0.13 & -0.11 & -0.01 & -1.57 \\
\addlinespace[3pt]
\bottomrule
\end{tabular}}
\end{table}

Tables~\ref{tab:iclr-router-common}--\ref{tab:iclr-router-longbench} give per-task results in the format of Tables~\ref{tab:common_sense}--\ref{tab:longbench}. Retrieval and S-NIAH follow the 1.5B evaluation setting in Appendix~\ref{app:eval:retrieval}.

\begin{table*}[htbp]

\centering
\footnotesize
\setlength{\tabcolsep}{3pt}
\caption{\textbf{Common-sense routing comparison.} 180M, 440M and 1.5B.}
\label{tab:iclr-router-common}
\resizebox{\textwidth}{!}{%
\begin{tabular}{lccccccccccc}
\toprule
\textbf{Model} & \textbf{FW-Edu} & \textbf{LAMB} & \textbf{LAMB} & \textbf{HSwag} & \textbf{PIQA} & \textbf{ARC-E} & \textbf{ARC-C} & \textbf{WinoG} & \textbf{OBQA} & \textbf{TQA mc2} & \textbf{Avg} \\
 & PPL $\downarrow$ & PPL $\downarrow$ & acc $\uparrow$ & acc $\uparrow$ & acc $\uparrow$ & acc $\uparrow$ & acc $\uparrow$ & acc $\uparrow$ & acc $\uparrow$ & acc $\uparrow$ & acc $\uparrow$ \\
\midrule
\multicolumn{12}{@{}l}{\textbf{\small 180M}} \\
\textcolor{red}{\amormamba{}\amorheart{}} & 31.46 & 240.27 & 20.3 & 27.4 & 60.5 & 46.5 & 18.0 & 51.3 & 16.6 & 45.9 & 35.8 \\
\quad + distilled router & N/A & 250.77 & 20.3 & 27.3 & 60.7 & 46.6 & 17.7 & 51.6 & 16.8 & 45.6 & 35.8 \\
\addlinespace[3pt]
\textcolor{red}{\amorgdn{}\amorheart{}} & 30.60 & 234.20 & 19.5 & 27.3 & 61.0 & 47.7 & 17.3 & 51.1 & 16.2 & 45.2 & 35.7 \\
\quad + distilled router & N/A & 237.83 & 19.4 & 27.3 & 60.9 & 47.7 & 17.5 & 51.0 & 15.8 & 45.3 & 35.6 \\
\addlinespace[3pt]
\midrule
\multicolumn{12}{@{}l}{\textbf{\small 440M}} \\
\textcolor{red}{\amormamba{}\amorheart{}} & 19.60 & 57.37 & 29.3 & 31.4 & 65.9 & 55.4 & 24.7 & 52.4 & 21.0 & 36.4 & 39.6 \\
\quad + distilled router & N/A & 57.51 & 29.3 & 31.5 & 65.8 & 55.6 & 24.8 & 52.2 & 21.2 & 36.0 & 39.6 \\
\addlinespace[3pt]
\textcolor{red}{\amorgdn{}\amorheart{}} & 19.27 & 50.15 & 30.2 & 31.8 & 65.0 & 56.0 & 24.6 & 51.3 & 23.8 & 40.9 & 40.5 \\
\quad + distilled router & N/A & 50.30 & 30.2 & 31.8 & 65.2 & 56.0 & 25.2 & 51.6 & 23.8 & 40.7 & 40.6 \\
\addlinespace[3pt]
\midrule
\multicolumn{12}{@{}l}{\textbf{\small 1.5B}} \\
\textcolor{red}{\amormamba{}\amorheart{}} & 13.30 & 23.01 & 39.2 & 38.6 & 70.9 & 67.0 & 30.2 & 54.7 & 26.8 & 39.0 & 45.8 \\
\quad + distilled router & N/A & 23.02 & 39.2 & 38.6 & 70.5 & 66.8 & 30.1 & 55.0 & 26.8 & 39.0 & 45.8 \\
\addlinespace[3pt]
\textcolor{red}{\amorgdn{}\amorheart{}} & 13.10 & 21.49 & 39.1 & 39.2 & 70.6 & 66.4 & 30.5 & 53.7 & 24.8 & 36.9 & 45.1 \\
\quad + distilled router & N/A & 21.54 & 39.0 & 39.2 & 70.2 & 66.3 & 30.5 & 53.2 & 24.8 & 36.8 & 45.0 \\
\addlinespace[3pt]
\bottomrule
\end{tabular}}
\end{table*}

\begin{table*}[htbp]

\centering
\footnotesize
\setlength{\tabcolsep}{3pt}
\caption{\textbf{Retrieval and S-NIAH routing comparison.} 1.5B.}
\label{tab:iclr-router-retrieval}
\resizebox{\textwidth}{!}{%
\begin{tabular}{l ccccccccccccccc}
    \toprule
    \textbf{Model} & \textbf{SWDE}$\uparrow$ & \textbf{SQuAD}$\uparrow$ & \textbf{FDA}$\uparrow$ & \textbf{TQA}$\uparrow$ & \textbf{NQ}$\uparrow$ & \textbf{DROP}$\uparrow$ & \multicolumn{3}{c}{\textbf{S-NIAH-1}$\uparrow$} & \multicolumn{3}{c}{\textbf{S-NIAH-2}$\uparrow$} & \multicolumn{3}{c}{\textbf{S-NIAH-3}$\uparrow$} \\
    \cmidrule(lr){2-7}\cmidrule(lr){8-10}\cmidrule(lr){11-13}\cmidrule(lr){14-16}
    \textbf{Context length} & \multicolumn{6}{c}{2048} & 1024 & 2048 & 4096 & 1024 & 2048 & 4096 & 1024 & 2048 & 4096 \\
    \midrule
\textcolor{red}{\amormamba{}\amorheart{}} & 33.8 & 36.0 & 21.2 & 40.4 & 10.2 & 21.1 & 100.0 & 95.8 & 44.4 & 100.0 & 99.6 & 47.2 & 70.8 & 83.6 & 25.6 \\
\quad + distilled router & 33.7 & 36.2 & 21.6 & 39.8 & 10.1 & 20.9 & 100.0 & 95.8 & 44.8 & 100.0 & 99.6 & 47.2 & 71.2 & 83.4 & 25.4 \\
\addlinespace[3pt]
\textcolor{red}{\amorgdn{}\amorheart{}} & 47.4 & 36.8 & 24.9 & 43.1 & 10.5 & 17.4 & 100.0 & 96.2 & 48.0 & 100.0 & 98.6 & 53.0 & 88.8 & 86.0 & 29.4 \\
\quad + distilled router & 47.6 & 37.0 & 25.4 & 42.5 & 10.7 & 17.7 & 100.0 & 96.2 & 48.0 & 100.0 & 98.4 & 53.0 & 89.0 & 86.0 & 28.4 \\
\addlinespace[3pt]
\bottomrule
\end{tabular}}
\end{table*}

\begin{table*}[htbp]

\centering
\footnotesize
\setlength{\tabcolsep}{3pt}
\caption{\textbf{LongBench routing comparison.} 180M, 440M and 1.5B.}
\label{tab:iclr-router-longbench}
\resizebox{\textwidth}{!}{%
\begin{tabular}{l ccccccccccccccc}
    \toprule
    \textbf{Model} & \multicolumn{3}{c}{\textbf{Single-Doc QA}$\uparrow$} & \multicolumn{3}{c}{\textbf{Multi-Doc QA}$\uparrow$} & \multicolumn{3}{c}{\textbf{Summarization}$\uparrow$} & \multicolumn{3}{c}{\textbf{Few-shot}$\uparrow$} & \multicolumn{2}{c}{\textbf{Code}$\uparrow$} & \textbf{Avg}$\uparrow$ \\
    \cmidrule(lr){2-4}\cmidrule(lr){5-7}\cmidrule(lr){8-10}\cmidrule(lr){11-13}\cmidrule(lr){14-15}
     & \textbf{NQA} & \textbf{QQA} & \textbf{MFQ} & \textbf{HQA} & \textbf{2WM} & \textbf{Mus} & \textbf{GvR} & \textbf{QMS} & \textbf{MNs} & \textbf{TRC} & \textbf{TQA} & \textbf{SSM} & \textbf{LCC} & \textbf{RBP} &  \\
    \midrule
\multicolumn{16}{@{}l}{\textbf{\small 180M}} \\
\textcolor{red}{\amormamba{}\amorheart{}} & 1.0\scriptsize$\pm$0.15 & 2.0\scriptsize$\pm$0.24 & 5.9\scriptsize$\pm$0.58 & 1.8\scriptsize$\pm$0.36 & 2.5\scriptsize$\pm$0.46 & 1.7\scriptsize$\pm$0.31 & 5.7\scriptsize$\pm$0.25 & 11.5\scriptsize$\pm$0.41 & 8.5\scriptsize$\pm$0.30 & 1.5\scriptsize$\pm$0.86 & 6.8\scriptsize$\pm$0.85 & 8.3\scriptsize$\pm$0.44 & 14.0\scriptsize$\pm$0.45 & 11.2\scriptsize$\pm$0.32 & 5.9\scriptsize$\pm$0.13 \\
\quad + distilled router & 1.1\scriptsize$\pm$0.17 & 2.2\scriptsize$\pm$0.26 & 5.7\scriptsize$\pm$0.57 & 1.8\scriptsize$\pm$0.36 & 2.6\scriptsize$\pm$0.47 & 1.7\scriptsize$\pm$0.31 & 5.7\scriptsize$\pm$0.25 & 12.1\scriptsize$\pm$0.42 & 8.3\scriptsize$\pm$0.29 & 1.5\scriptsize$\pm$0.86 & 6.7\scriptsize$\pm$0.84 & 8.0\scriptsize$\pm$0.46 & 14.0\scriptsize$\pm$0.45 & 11.3\scriptsize$\pm$0.34 & 5.9\scriptsize$\pm$0.13 \\
\addlinespace[3pt]
\textcolor{red}{\amorgdn{}\amorheart{}} & 0.8\scriptsize$\pm$0.11 & 2.0\scriptsize$\pm$0.25 & 4.6\scriptsize$\pm$0.51 & 1.6\scriptsize$\pm$0.31 & 1.8\scriptsize$\pm$0.36 & 1.6\scriptsize$\pm$0.31 & 4.4\scriptsize$\pm$0.23 & 11.5\scriptsize$\pm$0.44 & 7.4\scriptsize$\pm$0.31 & 8.8\scriptsize$\pm$1.99 & 6.8\scriptsize$\pm$0.63 & 8.3\scriptsize$\pm$0.51 & 12.0\scriptsize$\pm$0.31 & 14.0\scriptsize$\pm$0.32 & 6.1\scriptsize$\pm$0.17 \\
\quad + distilled router & 0.8\scriptsize$\pm$0.11 & 2.1\scriptsize$\pm$0.27 & 4.5\scriptsize$\pm$0.51 & 1.5\scriptsize$\pm$0.32 & 2.1\scriptsize$\pm$0.41 & 1.6\scriptsize$\pm$0.29 & 4.4\scriptsize$\pm$0.23 & 11.4\scriptsize$\pm$0.46 & 7.2\scriptsize$\pm$0.31 & 8.2\scriptsize$\pm$1.93 & 6.6\scriptsize$\pm$0.62 & 8.4\scriptsize$\pm$0.51 & 11.8\scriptsize$\pm$0.31 & 14.0\scriptsize$\pm$0.32 & 6.1\scriptsize$\pm$0.17 \\
\addlinespace[3pt]
\midrule
\multicolumn{16}{@{}l}{\textbf{\small 440M}} \\
\textcolor{red}{\amormamba{}\amorheart{}} & 1.3\scriptsize$\pm$0.14 & 2.3\scriptsize$\pm$0.22 & 7.5\scriptsize$\pm$0.77 & 2.6\scriptsize$\pm$0.44 & 4.5\scriptsize$\pm$0.59 & 2.1\scriptsize$\pm$0.39 & 7.3\scriptsize$\pm$0.24 & 12.6\scriptsize$\pm$0.47 & 11.5\scriptsize$\pm$0.36 & 5.0\scriptsize$\pm$1.54 & 9.0\scriptsize$\pm$1.15 & 9.2\scriptsize$\pm$0.54 & 9.8\scriptsize$\pm$0.33 & 10.1\scriptsize$\pm$0.32 & 6.8\scriptsize$\pm$0.17 \\
\quad + distilled router & 1.2\scriptsize$\pm$0.14 & 2.4\scriptsize$\pm$0.23 & 7.5\scriptsize$\pm$0.76 & 2.9\scriptsize$\pm$0.49 & 4.6\scriptsize$\pm$0.59 & 2.1\scriptsize$\pm$0.39 & 7.5\scriptsize$\pm$0.25 & 13.0\scriptsize$\pm$0.46 & 11.6\scriptsize$\pm$0.37 & 4.5\scriptsize$\pm$1.47 & 8.9\scriptsize$\pm$1.15 & 9.1\scriptsize$\pm$0.55 & 10.0\scriptsize$\pm$0.33 & 10.3\scriptsize$\pm$0.33 & 6.8\scriptsize$\pm$0.17 \\
\addlinespace[3pt]
\textcolor{red}{\amorgdn{}\amorheart{}} & 2.0\scriptsize$\pm$0.33 & 3.2\scriptsize$\pm$0.31 & 10.6\scriptsize$\pm$0.90 & 3.3\scriptsize$\pm$0.48 & 6.6\scriptsize$\pm$0.72 & 2.0\scriptsize$\pm$0.32 & 6.9\scriptsize$\pm$0.22 & 13.0\scriptsize$\pm$0.51 & 7.5\scriptsize$\pm$0.46 & 6.0\scriptsize$\pm$1.68 & 8.3\scriptsize$\pm$0.84 & 8.9\scriptsize$\pm$0.56 & 9.2\scriptsize$\pm$0.26 & 9.7\scriptsize$\pm$0.24 & 7.0\scriptsize$\pm$0.18 \\
\quad + distilled router & 2.0\scriptsize$\pm$0.33 & 3.2\scriptsize$\pm$0.29 & 10.6\scriptsize$\pm$0.90 & 3.3\scriptsize$\pm$0.48 & 6.7\scriptsize$\pm$0.73 & 2.1\scriptsize$\pm$0.35 & 7.1\scriptsize$\pm$0.23 & 13.1\scriptsize$\pm$0.49 & 7.6\scriptsize$\pm$0.47 & 6.5\scriptsize$\pm$1.75 & 8.2\scriptsize$\pm$0.80 & 8.7\scriptsize$\pm$0.55 & 9.3\scriptsize$\pm$0.26 & 9.8\scriptsize$\pm$0.24 & 7.0\scriptsize$\pm$0.18 \\
\addlinespace[3pt]
\midrule
\multicolumn{16}{@{}l}{\textbf{\small 1.5B}} \\
\textcolor{red}{\amormamba{}\amorheart{}} & 1.6\scriptsize$\pm$0.15 & 3.3\scriptsize$\pm$0.32 & 10.1\scriptsize$\pm$0.83 & 2.9\scriptsize$\pm$0.50 & 6.1\scriptsize$\pm$0.71 & 1.8\scriptsize$\pm$0.37 & 7.6\scriptsize$\pm$0.23 & 14.7\scriptsize$\pm$0.46 & 13.2\scriptsize$\pm$0.47 & 13.0\scriptsize$\pm$2.38 & 13.2\scriptsize$\pm$1.49 & 7.5\scriptsize$\pm$0.51 & 9.3\scriptsize$\pm$0.26 & 11.5\scriptsize$\pm$0.29 & 8.3\scriptsize$\pm$0.23 \\
\quad + distilled router & 1.6\scriptsize$\pm$0.15 & 3.2\scriptsize$\pm$0.31 & 10.1\scriptsize$\pm$0.83 & 3.0\scriptsize$\pm$0.53 & 5.8\scriptsize$\pm$0.70 & 1.8\scriptsize$\pm$0.34 & 7.6\scriptsize$\pm$0.23 & 14.4\scriptsize$\pm$0.46 & 13.3\scriptsize$\pm$0.46 & 13.0\scriptsize$\pm$2.38 & 13.3\scriptsize$\pm$1.45 & 7.4\scriptsize$\pm$0.49 & 9.5\scriptsize$\pm$0.27 & 11.5\scriptsize$\pm$0.29 & 8.3\scriptsize$\pm$0.23 \\
\addlinespace[3pt]
\textcolor{red}{\amorgdn{}\amorheart{}} & 1.7\scriptsize$\pm$0.17 & 4.0\scriptsize$\pm$0.35 & 11.4\scriptsize$\pm$0.87 & 3.6\scriptsize$\pm$0.49 & 7.0\scriptsize$\pm$0.74 & 1.9\scriptsize$\pm$0.32 & 8.1\scriptsize$\pm$0.30 & 15.1\scriptsize$\pm$0.50 & 10.8\scriptsize$\pm$0.43 & 11.0\scriptsize$\pm$2.22 & 11.8\scriptsize$\pm$1.22 & 11.4\scriptsize$\pm$0.68 & 10.1\scriptsize$\pm$0.38 & 10.3\scriptsize$\pm$0.41 & 8.4\scriptsize$\pm$0.22 \\
\quad + distilled router & 1.7\scriptsize$\pm$0.17 & 4.0\scriptsize$\pm$0.36 & 11.1\scriptsize$\pm$0.88 & 3.8\scriptsize$\pm$0.51 & 6.8\scriptsize$\pm$0.74 & 1.9\scriptsize$\pm$0.32 & 8.1\scriptsize$\pm$0.30 & 15.4\scriptsize$\pm$0.50 & 10.9\scriptsize$\pm$0.41 & 11.0\scriptsize$\pm$2.22 & 11.4\scriptsize$\pm$1.22 & 11.4\scriptsize$\pm$0.67 & 10.1\scriptsize$\pm$0.37 & 10.4\scriptsize$\pm$0.41 & 8.4\scriptsize$\pm$0.22 \\
\addlinespace[3pt]
\bottomrule
\end{tabular}}
\end{table*}

\FloatBarrier
\Needspace{5\baselineskip}
Tables~\ref{tab:iclr-router-length-180m-six}--\ref{tab:iclr-router-length-1p5b-fineweb} report length extrapolation results across all three scales (Appendix~\ref{app:eval:extrapolation}). FineWeb-Edu uses the same protocol with additional 1K and 2K contexts. At 20K, relative differences remain within 1.62\% on six of seven domains. CodeParrot is the exception: the distilled router lowers perplexity for both AMOR variants at every scale, including 979.92 to 475.75 for \amormamba{} and 289.14 to 203.17 for \amorgdn{} at 440M. Larger differences also occur at shorter contexts: 180M \amormamba{}'s PG-19 perplexity at 3K changes from 106.05 to 87.37.

\begingroup
\makeatletter
\setlength{\@fpsep}{12pt}
\setlength{\@dblfpsep}{12pt}
\makeatother
\begin{table*}[htbp]

\centering
\footnotesize
\setlength{\tabcolsep}{3pt}
\caption{\textbf{Length extrapolation routing comparison.} 180M.}
\label{tab:iclr-router-length-180m-six}
\resizebox{\textwidth}{!}{%
\begin{tabular}{lrrrrrrrrrr}
\toprule
\textbf{Model} & \multicolumn{10}{c}{\textbf{Context length; PPL $\downarrow$}} \\
\cmidrule(lr){2-11}
 & 3K & 4K & 6K & 8K & 10K & 12K & 14K & 16K & 18K & 20K \\
\midrule
\multicolumn{11}{@{}l}{\textbf{GovReport}} \\
\textcolor{red}{\amormamba{}\amorheart{}} & 24.72 & 24.27 & 24.95 & 26.13 & 26.72 & 27.63 & 28.43 & 28.85 & 29.05 & 29.34 \\
\quad + distilled router & 25.14 & 24.58 & 25.20 & 26.38 & 26.93 & 27.87 & 28.72 & 29.17 & 29.40 & 29.68 \\
\textcolor{red}{\amorgdn{}\amorheart{}} & 23.14 & 22.53 & 22.93 & 23.57 & 23.56 & 23.97 & 24.33 & 24.31 & 24.29 & 24.28 \\
\quad + distilled router & 23.42 & 22.77 & 23.14 & 23.76 & 23.73 & 24.14 & 24.48 & 24.46 & 24.43 & 24.42 \\
\addlinespace[3pt]
\multicolumn{11}{@{}l}{\textbf{QMSum}} \\
\textcolor{red}{\amormamba{}\amorheart{}} & 29.20 & 31.71 & 36.29 & 38.51 & 41.38 & 44.67 & 46.26 & 45.61 & 46.51 & 50.10 \\
\quad + distilled router & 29.39 & 32.05 & 36.59 & 38.62 & 41.06 & 44.08 & 45.57 & 44.92 & 45.93 & 49.63 \\
\textcolor{red}{\amorgdn{}\amorheart{}} & 28.22 & 30.14 & 33.64 & 35.39 & 37.92 & 40.22 & 40.50 & 39.14 & 39.12 & 41.63 \\
\quad + distilled router & 28.34 & 30.30 & 33.77 & 35.41 & 37.80 & 40.02 & 40.27 & 38.93 & 38.93 & 41.46 \\
\addlinespace[3pt]
\multicolumn{11}{@{}l}{\textbf{NarrativeQA}} \\
\textcolor{red}{\amormamba{}\amorheart{}} & 159.22 & 160.34 & 164.00 & 165.90 & 152.32 & 143.05 & 137.83 & 133.96 & 131.37 & 129.06 \\
\quad + distilled router & 154.98 & 153.30 & 155.65 & 156.90 & 144.39 & 137.06 & 133.64 & 131.36 & 129.90 & 128.43 \\
\textcolor{red}{\amorgdn{}\amorheart{}} & 146.35 & 145.54 & 147.09 & 146.89 & 132.77 & 122.98 & 117.35 & 112.55 & 109.26 & 106.30 \\
\quad + distilled router & 145.61 & 143.67 & 144.28 & 143.42 & 129.08 & 119.81 & 114.71 & 110.46 & 107.68 & 105.05 \\
\addlinespace[3pt]
\multicolumn{11}{@{}l}{\textbf{Qasper}} \\
\textcolor{red}{\amormamba{}\amorheart{}} & 54.66 & 54.89 & 54.16 & 56.99 & 58.99 & 60.79 & 62.88 & 64.50 & 66.34 & 68.26 \\
\quad + distilled router & 50.30 & 50.78 & 50.73 & 54.10 & 56.58 & 58.64 & 61.22 & 63.18 & 65.52 & 67.64 \\
\textcolor{red}{\amorgdn{}\amorheart{}} & 51.17 & 51.33 & 49.31 & 50.71 & 51.37 & 51.65 & 52.57 & 52.71 & 53.45 & 54.12 \\
\quad + distilled router & 49.19 & 49.58 & 47.83 & 49.40 & 50.24 & 50.66 & 51.70 & 51.94 & 52.80 & 53.55 \\
\addlinespace[3pt]
\multicolumn{11}{@{}l}{\textbf{CodeParrot}} \\
\textcolor{red}{\amormamba{}\amorheart{}} & 314.18 & 369.89 & 427.68 & 688.59 & 1045.73 & 1547.90 & 2152.41 & 2530.52 & 2855.07 & 3080.56 \\
\quad + distilled router & 303.74 & 324.28 & 311.97 & 372.18 & 426.20 & 503.95 & 610.43 & 674.70 & 760.62 & 845.78 \\
\textcolor{red}{\amorgdn{}\amorheart{}} & 393.21 & 431.41 & 416.22 & 508.64 & 595.98 & 728.60 & 876.08 & 945.24 & 1017.08 & 1068.88 \\
\quad + distilled router & 383.10 & 396.98 & 348.64 & 376.57 & 395.67 & 445.66 & 492.60 & 505.26 & 532.69 & 559.38 \\
\addlinespace[3pt]
\multicolumn{11}{@{}l}{\textbf{PG-19}} \\
\textcolor{red}{\amormamba{}\amorheart{}} & 106.05 & 89.15 & 83.93 & 93.12 & 96.37 & 96.45 & 96.04 & 94.77 & 95.69 & 94.77 \\
\quad + distilled router & 87.37 & 79.33 & 79.48 & 90.26 & 94.36 & 95.05 & 95.18 & 94.24 & 95.44 & 94.85 \\
\textcolor{red}{\amorgdn{}\amorheart{}} & 91.00 & 80.41 & 77.95 & 85.40 & 87.65 & 86.94 & 86.31 & 84.32 & 84.14 & 82.57 \\
\quad + distilled router & 82.15 & 74.79 & 74.83 & 83.14 & 85.91 & 85.65 & 85.35 & 83.57 & 83.50 & 82.00 \\
\bottomrule
\end{tabular}}
\end{table*}

\begin{table*}[htbp]

\centering
\footnotesize
\setlength{\tabcolsep}{3pt}
\caption{\textbf{FineWeb-Edu length extrapolation routing comparison.} 180M.}
\label{tab:iclr-router-length-180m-fineweb}
\resizebox{\textwidth}{!}{%
\begin{tabular}{lrrrrrrrrrrrr}
\toprule
\textbf{Model} & \multicolumn{12}{c}{\textbf{Context length; PPL $\downarrow$}} \\
\cmidrule(lr){2-13}
 & 1K & 2K & 3K & 4K & 6K & 8K & 10K & 12K & 14K & 16K & 18K & 20K \\
\midrule
\multicolumn{13}{@{}l}{\textbf{FineWeb-Edu}} \\
\textcolor{red}{\amormamba{}\amorheart{}} & 32.91 & 28.07 & 28.83 & 29.35 & 29.31 & 29.54 & 32.67 & 32.18 & 32.35 & 32.18 & 32.17 & 31.90 \\
\quad + distilled router & 33.02 & 28.24 & 29.00 & 29.55 & 29.51 & 29.74 & 32.89 & 32.38 & 32.55 & 32.39 & 32.42 & 32.15 \\
\textcolor{red}{\amorgdn{}\amorheart{}} & 32.21 & 27.42 & 28.12 & 28.55 & 28.34 & 28.41 & 31.24 & 30.56 & 30.59 & 30.25 & 30.14 & 29.83 \\
\quad + distilled router & 32.33 & 27.54 & 28.24 & 28.68 & 28.48 & 28.55 & 31.36 & 30.67 & 30.71 & 30.35 & 30.25 & 29.94 \\
\bottomrule
\end{tabular}}
\end{table*}

\begin{table*}[htbp]

\centering
\footnotesize
\setlength{\tabcolsep}{3pt}
\caption{\textbf{Length extrapolation routing comparison.} 440M.}
\label{tab:iclr-router-length-440m-six}
\resizebox{\textwidth}{!}{%
\begin{tabular}{lrrrrrrrrrr}
\toprule
\textbf{Model} & \multicolumn{10}{c}{\textbf{Context length; PPL $\downarrow$}} \\
\cmidrule(lr){2-11}
 & 3K & 4K & 6K & 8K & 10K & 12K & 14K & 16K & 18K & 20K \\
\midrule
\multicolumn{11}{@{}l}{\textbf{GovReport}} \\
\textcolor{red}{\amormamba{}\amorheart{}} & 15.10 & 14.78 & 15.11 & 15.80 & 16.17 & 16.80 & 17.36 & 17.69 & 17.83 & 18.06 \\
\quad + distilled router & 15.17 & 14.81 & 15.13 & 15.83 & 16.20 & 16.85 & 17.43 & 17.77 & 17.92 & 18.17 \\
\textcolor{red}{\amorgdn{}\amorheart{}} & 14.70 & 14.35 & 14.51 & 14.72 & 14.64 & 14.80 & 14.96 & 14.92 & 14.91 & 14.94 \\
\quad + distilled router & 14.74 & 14.40 & 14.53 & 14.73 & 14.64 & 14.80 & 14.97 & 14.91 & 14.90 & 14.94 \\
\addlinespace[3pt]
\multicolumn{11}{@{}l}{\textbf{QMSum}} \\
\textcolor{red}{\amormamba{}\amorheart{}} & 19.71 & 21.13 & 24.03 & 26.25 & 29.17 & 32.06 & 33.70 & 33.41 & 34.71 & 38.26 \\
\quad + distilled router & 19.70 & 21.19 & 24.07 & 26.22 & 29.10 & 31.98 & 33.64 & 33.40 & 34.72 & 38.33 \\
\textcolor{red}{\amorgdn{}\amorheart{}} & 19.33 & 20.64 & 22.85 & 23.78 & 25.05 & 26.57 & 26.68 & 25.67 & 25.58 & 26.92 \\
\quad + distilled router & 19.36 & 20.68 & 22.86 & 23.76 & 25.02 & 26.52 & 26.64 & 25.65 & 25.58 & 26.95 \\
\addlinespace[3pt]
\multicolumn{11}{@{}l}{\textbf{NarrativeQA}} \\
\textcolor{red}{\amormamba{}\amorheart{}} & 92.51 & 92.96 & 95.95 & 97.95 & 89.49 & 83.61 & 80.57 & 78.38 & 77.14 & 76.18 \\
\quad + distilled router & 90.35 & 89.51 & 91.90 & 93.75 & 86.02 & 81.28 & 79.10 & 77.51 & 76.77 & 76.15 \\
\textcolor{red}{\amorgdn{}\amorheart{}} & 86.58 & 85.33 & 84.97 & 83.62 & 74.31 & 68.22 & 64.54 & 61.97 & 60.25 & 58.79 \\
\quad + distilled router & 86.04 & 84.28 & 83.35 & 81.58 & 72.47 & 66.84 & 63.55 & 61.30 & 59.80 & 58.49 \\
\addlinespace[3pt]
\multicolumn{11}{@{}l}{\textbf{Qasper}} \\
\textcolor{red}{\amormamba{}\amorheart{}} & 32.21 & 32.31 & 31.79 & 33.52 & 34.92 & 36.30 & 37.71 & 39.05 & 40.51 & 41.63 \\
\quad + distilled router & 30.26 & 30.50 & 30.26 & 32.18 & 33.74 & 35.31 & 36.89 & 38.35 & 39.98 & 41.17 \\
\textcolor{red}{\amorgdn{}\amorheart{}} & 30.93 & 30.90 & 29.58 & 30.04 & 30.28 & 30.25 & 30.67 & 30.73 & 31.23 & 31.65 \\
\quad + distilled router & 29.83 & 29.96 & 28.84 & 29.36 & 29.69 & 29.72 & 30.20 & 30.31 & 30.88 & 31.35 \\
\addlinespace[3pt]
\multicolumn{11}{@{}l}{\textbf{CodeParrot}} \\
\textcolor{red}{\amormamba{}\amorheart{}} & 142.47 & 171.03 & 188.98 & 275.66 & 378.98 & 531.07 & 702.68 & 796.29 & 899.08 & 979.92 \\
\quad + distilled router & 137.46 & 147.19 & 142.72 & 173.52 & 209.09 & 265.56 & 332.95 & 371.48 & 422.99 & 475.75 \\
\textcolor{red}{\amorgdn{}\amorheart{}} & 158.09 & 173.46 & 166.35 & 188.63 & 199.46 & 224.47 & 251.23 & 260.66 & 276.53 & 289.14 \\
\quad + distilled router & 154.91 & 158.52 & 140.95 & 148.50 & 151.28 & 164.88 & 178.38 & 181.91 & 192.72 & 203.17 \\
\addlinespace[3pt]
\multicolumn{11}{@{}l}{\textbf{PG-19}} \\
\textcolor{red}{\amormamba{}\amorheart{}} & 57.90 & 49.16 & 48.45 & 53.19 & 54.84 & 54.89 & 54.61 & 54.04 & 54.72 & 53.87 \\
\quad + distilled router & 49.65 & 45.28 & 46.41 & 51.90 & 54.06 & 54.50 & 54.48 & 54.02 & 54.77 & 54.07 \\
\textcolor{red}{\amorgdn{}\amorheart{}} & 50.80 & 45.05 & 45.11 & 48.63 & 49.47 & 48.89 & 48.23 & 47.22 & 47.23 & 46.24 \\
\quad + distilled router & 47.57 & 43.27 & 44.09 & 47.89 & 48.91 & 48.51 & 47.97 & 47.01 & 47.09 & 46.17 \\
\bottomrule
\end{tabular}}
\end{table*}

\begin{table*}[htbp]

\centering
\footnotesize
\setlength{\tabcolsep}{3pt}
\caption{\textbf{FineWeb-Edu length extrapolation routing comparison.} 440M.}
\label{tab:iclr-router-length-440m-fineweb}
\resizebox{\textwidth}{!}{%
\begin{tabular}{lrrrrrrrrrrrr}
\toprule
\textbf{Model} & \multicolumn{12}{c}{\textbf{Context length; PPL $\downarrow$}} \\
\cmidrule(lr){2-13}
 & 1K & 2K & 3K & 4K & 6K & 8K & 10K & 12K & 14K & 16K & 18K & 20K \\
\midrule
\multicolumn{13}{@{}l}{\textbf{FineWeb-Edu}} \\
\textcolor{red}{\amormamba{}\amorheart{}} & 20.43 & 17.80 & 18.28 & 18.65 & 18.45 & 18.57 & 20.46 & 20.17 & 20.29 & 20.22 & 20.30 & 20.13 \\
\quad + distilled router & 20.45 & 17.83 & 18.32 & 18.68 & 18.48 & 18.60 & 20.49 & 20.20 & 20.31 & 20.24 & 20.31 & 20.14 \\
\textcolor{red}{\amorgdn{}\amorheart{}} & 20.18 & 17.54 & 17.99 & 18.32 & 18.02 & 17.98 & 19.59 & 19.13 & 19.14 & 18.96 & 18.90 & 18.72 \\
\quad + distilled router & 20.19 & 17.56 & 18.00 & 18.33 & 18.03 & 17.98 & 19.60 & 19.14 & 19.14 & 18.96 & 18.91 & 18.73 \\
\bottomrule
\end{tabular}}
\end{table*}

\begin{table*}[htbp]

\centering
\footnotesize
\setlength{\tabcolsep}{3pt}
\caption{\textbf{Length extrapolation routing comparison.} 1.5B.}
\label{tab:iclr-router-length-1p5b-six}
\resizebox{\textwidth}{!}{%
\begin{tabular}{lrrrrrrrrrr}
\toprule
\textbf{Model} & \multicolumn{10}{c}{\textbf{Context length; PPL $\downarrow$}} \\
\cmidrule(lr){2-11}
 & 3K & 4K & 6K & 8K & 10K & 12K & 14K & 16K & 18K & 20K \\
\midrule
\multicolumn{11}{@{}l}{\textbf{GovReport}} \\
\textcolor{red}{\amormamba{}\amorheart{}} & 11.13 & 10.93 & 11.20 & 11.73 & 12.06 & 12.60 & 13.15 & 13.52 & 13.67 & 13.92 \\
\quad + distilled router & 11.14 & 10.91 & 11.17 & 11.69 & 12.04 & 12.60 & 13.17 & 13.55 & 13.73 & 13.99 \\
\textcolor{red}{\amorgdn{}\amorheart{}} & 10.88 & 10.63 & 10.76 & 11.02 & 11.08 & 11.36 & 11.58 & 11.67 & 11.73 & 11.87 \\
\quad + distilled router & 10.91 & 10.65 & 10.78 & 11.03 & 11.08 & 11.36 & 11.58 & 11.66 & 11.73 & 11.86 \\
\addlinespace[3pt]
\multicolumn{11}{@{}l}{\textbf{QMSum}} \\
\textcolor{red}{\amormamba{}\amorheart{}} & 14.73 & 15.82 & 18.18 & 19.87 & 22.18 & 25.54 & 27.84 & 27.84 & 29.80 & 34.10 \\
\quad + distilled router & 14.71 & 15.84 & 18.17 & 19.84 & 22.20 & 25.68 & 28.08 & 28.15 & 30.19 & 34.65 \\
\textcolor{red}{\amorgdn{}\amorheart{}} & 14.60 & 15.78 & 18.16 & 19.82 & 22.12 & 24.62 & 25.73 & 24.88 & 25.56 & 28.16 \\
\quad + distilled router & 14.61 & 15.81 & 18.19 & 19.83 & 22.16 & 24.71 & 25.87 & 25.05 & 25.73 & 28.40 \\
\addlinespace[3pt]
\multicolumn{11}{@{}l}{\textbf{NarrativeQA}} \\
\textcolor{red}{\amormamba{}\amorheart{}} & 62.36 & 61.73 & 63.07 & 64.81 & 60.75 & 58.34 & 57.36 & 56.95 & 56.68 & 56.76 \\
\quad + distilled router & 60.91 & 59.78 & 61.06 & 62.78 & 59.36 & 57.61 & 57.09 & 57.05 & 57.01 & 57.24 \\
\textcolor{red}{\amorgdn{}\amorheart{}} & 59.47 & 58.72 & 59.57 & 60.24 & 55.44 & 52.22 & 50.28 & 48.99 & 48.10 & 47.35 \\
\quad + distilled router & 58.97 & 57.84 & 58.36 & 58.76 & 54.10 & 51.17 & 49.61 & 48.60 & 47.93 & 47.32 \\
\addlinespace[3pt]
\multicolumn{11}{@{}l}{\textbf{Qasper}} \\
\textcolor{red}{\amormamba{}\amorheart{}} & 22.01 & 22.12 & 21.82 & 22.86 & 24.10 & 25.25 & 26.51 & 27.68 & 28.94 & 29.83 \\
\quad + distilled router & 21.15 & 21.28 & 21.14 & 22.31 & 23.67 & 24.93 & 26.33 & 27.60 & 28.99 & 29.97 \\
\textcolor{red}{\amorgdn{}\amorheart{}} & 21.15 & 21.21 & 20.82 & 21.68 & 22.66 & 23.42 & 24.17 & 24.91 & 25.65 & 26.26 \\
\quad + distilled router & 20.53 & 20.57 & 20.29 & 21.20 & 22.26 & 23.11 & 23.95 & 24.74 & 25.59 & 26.29 \\
\addlinespace[3pt]
\multicolumn{11}{@{}l}{\textbf{CodeParrot}} \\
\textcolor{red}{\amormamba{}\amorheart{}} & 89.17 & 101.90 & 113.85 & 161.26 & 218.54 & 313.57 & 417.63 & 482.96 & 561.35 & 627.21 \\
\quad + distilled router & 85.64 & 88.05 & 87.51 & 111.80 & 144.48 & 201.19 & 266.65 & 306.95 & 363.99 & 421.59 \\
\textcolor{red}{\amorgdn{}\amorheart{}} & 81.74 & 89.97 & 95.00 & 122.66 & 155.49 & 210.90 & 267.07 & 304.26 & 343.55 & 376.96 \\
\quad + distilled router & 79.02 & 81.59 & 80.79 & 99.06 & 121.05 & 156.79 & 193.35 & 215.94 & 242.84 & 269.15 \\
\addlinespace[3pt]
\multicolumn{11}{@{}l}{\textbf{PG-19}} \\
\textcolor{red}{\amormamba{}\amorheart{}} & 38.16 & 33.17 & 33.24 & 36.16 & 37.41 & 37.59 & 37.51 & 37.35 & 38.22 & 37.73 \\
\quad + distilled router & 34.14 & 31.13 & 32.24 & 35.64 & 37.16 & 37.54 & 37.60 & 37.52 & 38.45 & 38.07 \\
\textcolor{red}{\amorgdn{}\amorheart{}} & 37.14 & 31.90 & 31.94 & 34.50 & 35.26 & 35.01 & 34.62 & 34.18 & 34.63 & 34.00 \\
\quad + distilled router & 33.30 & 30.26 & 31.18 & 34.07 & 35.02 & 34.90 & 34.59 & 34.20 & 34.69 & 34.12 \\
\bottomrule
\end{tabular}}
\end{table*}

\begin{table*}[htbp]

\centering
\footnotesize
\setlength{\tabcolsep}{3pt}
\caption{\textbf{FineWeb-Edu length extrapolation routing comparison.} 1.5B.}
\label{tab:iclr-router-length-1p5b-fineweb}
\resizebox{\textwidth}{!}{%
\begin{tabular}{lrrrrrrrrrrrr}
\toprule
\textbf{Model} & \multicolumn{12}{c}{\textbf{Context length; PPL $\downarrow$}} \\
\cmidrule(lr){2-13}
 & 1K & 2K & 3K & 4K & 6K & 8K & 10K & 12K & 14K & 16K & 18K & 20K \\
\midrule
\multicolumn{13}{@{}l}{\textbf{FineWeb-Edu}} \\
\textcolor{red}{\amormamba{}\amorheart{}} & 14.64 & 12.88 & 13.08 & 13.30 & 13.23 & 13.30 & 14.11 & 14.03 & 14.26 & 14.33 & 14.50 & 14.42 \\
\quad + distilled router & 14.65 & 12.89 & 13.09 & 13.30 & 13.23 & 13.31 & 14.12 & 14.03 & 14.26 & 14.33 & 14.50 & 14.41 \\
\textcolor{red}{\amorgdn{}\amorheart{}} & 14.46 & 12.73 & 12.91 & 13.11 & 12.98 & 13.00 & 13.71 & 13.56 & 13.69 & 13.68 & 13.76 & 13.64 \\
\quad + distilled router & 14.47 & 12.74 & 12.91 & 13.11 & 12.98 & 13.00 & 13.71 & 13.56 & 13.69 & 13.67 & 13.76 & 13.63 \\
\bottomrule
\end{tabular}}
\end{table*}

\FloatBarrier
\endgroup
\Needspace{5\baselineskip}
\subsection{Routing-signal controls}
Table~\ref{tab:iclr-routing-controls} compares the native entropy gate and distilled router with a random router and zeroed attention outputs at 1.5B. The random router is an untrained predictor with a 256-unit hidden layer and an uncalibrated firing rate. Attention output zeroing sets the output projection $\mathbf W_O$ to zero in each of the $K=3$ \amor{} blocks, keeping all other weights fixed. Random routing lowers common-sense accuracy (Appendix~\ref{app:eval:common}) by about three points and three-task retrieval (SWDE, SQuAD-completion and FDA) by roughly two thirds. Most of the common-sense loss is on LAMBADA: \amormamba{} accuracy falls from 39.2\% to 13.2\%, and perplexity rises from 23.01 to 691.84. Attention output zeroing lowers the retrieval averages to 9.0\% and 11.4\% on \amormamba{} and \amorgdn{}.

\begin{table}[h]
  \centering
  \footnotesize
  \setlength{\tabcolsep}{5pt}
  \caption{\textbf{Routing-signal controls.} 1.5B.}
  \label{tab:iclr-routing-controls}
  \begingroup
  \begin{tabular}{lrrrr}
    \toprule
\textbf{Routing mechanism} & \multicolumn{2}{c}{\textcolor{red}{\amormamba{}\amorheart{}}} & \multicolumn{2}{c}{\textcolor{red}{\amorgdn{}\amorheart{}}} \\
    \cmidrule(lr){2-3}\cmidrule(lr){4-5}
& \textbf{Common-sense} & \textbf{Cloze retrieval} & \textbf{Common-sense} & \textbf{Cloze retrieval} \\
& avg acc (\%) $\uparrow$ & avg (\%) $\uparrow$ & avg acc (\%) $\uparrow$ & avg (\%) $\uparrow$ \\
    \midrule
\amor{} (native entropy gate) & 45.8 & 30.4 & 45.1 & 36.4 \\
\quad + distilled router & 45.8 & 30.5 & 45.0 & 36.7 \\
    \addlinespace[6pt]
Random router & 42.4 & 9.8 & 42.3 & 11.8 \\
Attention output zeroed & N/A & 9.0 & N/A & 11.4 \\
    \bottomrule
  \end{tabular}
  \endgroup
\end{table}

\FloatBarrier
\Needspace{5\baselineskip}
\section{Design Ablations}
\label{app:ablations}

We study six design ablations of \amor{}, each pretrained from scratch with its modified architecture. Within each scale, runs share the model width, FineWeb-Edu data, tokenizer, sequence length, training-token budget, effective batch size, AdamW settings, learning-rate schedule and warmup, gradient clipping, bfloat16 precision and training seed 42, following Section~\ref{sec:experiments} and Appendix~\ref{app:training}.

\textbf{Block design.} We compare canonical \amor{} with three variants on both recurrent backbones. \textbf{Symmetric block normalization} (\amor{}-symmetric) uses the backbone's final RMSNorm as Block~0's pre-norm and carries an unnormalized residual stream through all three \amor{} blocks. The \textbf{ungated post-hoc hybrid} (\amor{}-ungated) uses this symmetric layout with always-on causal attention, plain residual addition and Gaussian-initialized output projections. \textbf{\amor{}-MLP} adds an unconditional, pre-norm SwiGLU MLP after each \amor{} block, matching the backbone's MLP hidden dimension.

At 440M, canonical \amor{} has the highest eight-task common-sense average among these variants on both backbones (Table~\ref{tab:ablation_440m}). At 1.5B, symmetric block normalization slightly lowers FW-Edu PPL and has mixed common-sense results (Table~\ref{tab:common_sense_amor_h_1p5b}). It trails canonical \amor{} on four of six cloze retrieval tasks with Mamba2 and five with Gated DeltaNet (Table~\ref{tab:retrieval_amor_h_1p5b}).

\textbf{Block count.} We compare $K=3$ and \textbf{$\boldsymbol{K=1}$ \amor{} blocks} after the same recurrent backbone at 180M, 440M and 1.5B. Each block has its own entropy gate, attention projections and per-channel scaling vector $\boldsymbol{\alpha}$; $K=3$ is the canonical configuration presented in the main paper.

With the Mamba2 backbone, $K=1$ has a higher common-sense average at 180M and 440M, while $K=3$ leads at 1.5B. With Gated DeltaNet, $K=3$ leads at all three scales (Table~\ref{tab:vs_classic}).

\textbf{Attention-block placement.} Table~\ref{tab:iclr-placement} compares canonical \amor{}, which appends $K=3$ blocks after the recurrent backbone, with \textbf{\amor{}-serial} and \textbf{\amor{}-fused} at 440M. \amor{}-serial replaces recurrent layers at $\{6,12,18\}$ with \amor{} blocks. \amor{}-fused runs \amor{} gated attention blocks in parallel with the recurrent mixer at layers $\{0,12,23\}$, mirroring the fused hybrid baseline's placement in Section~\ref{sec:experiments}. Both evaluate entropy at intermediate depths.

Relative to canonical \amor{}, serial placement lowers the common-sense average by 0.90 points with Mamba2 and 0.56 points with Gated DeltaNet. Fused placement leaves the Mamba2 average unchanged and lowers the Gated DeltaNet average by 0.84 points (Table~\ref{tab:iclr-placement}).

\begingroup
\makeatletter
\setlength{\@fpsep}{12pt}
\setlength{\@dblfpsep}{12pt}
\makeatother
\begin{table*}[t]
\centering
\footnotesize
\setlength{\tabcolsep}{3pt}
\caption{\textbf{\amor{} block-design ablations: common-sense comparison.} 440M; in the format of Table~\ref{tab:common_sense}.}
\label{tab:ablation_440m}
\resizebox{\textwidth}{!}{%
\begin{tabular}{lccccccccc}
\toprule
\textbf{Model} & \textbf{LAMB} & \textbf{HSwag} & \textbf{PIQA} & \textbf{ARC-E} & \textbf{ARC-C} & \textbf{WinoG} & \textbf{OBQA} & \textbf{TQA mc2} & \textbf{Avg} \\
& acc $\uparrow$ & acc $\uparrow$ & acc $\uparrow$ & acc $\uparrow$ & acc $\uparrow$ & acc $\uparrow$ & acc $\uparrow$ & acc $\uparrow$ & acc $\uparrow$ \\
\midrule
\textcolor{red}{\amormamba{}\amorheart{}} & \textbf{29.3} & 31.4 & 65.9 & 55.4 & \textbf{24.7} & \textbf{52.4} & 21.0 & 36.4 & \textbf{39.6} \\
\amor{}-symmetric-Mamba2 & 28.3 & \underline{31.7} & 65.1 & \underline{56.5} & 22.1 & 49.2 & \textbf{23.0} & 35.7 & 38.9 \\
\amor{}-ungated-Mamba2 & \underline{28.9} & 31.5 & \underline{66.0} & 55.3 & 22.9 & 47.6 & \underline{21.6} & \underline{37.1} & 38.9 \\
\textsc{Amor}-MLP-Mamba2 & 27.4 & \textbf{31.8} & \textbf{66.2} & \textbf{56.7} & \underline{23.1} & \underline{51.1} & 19.2 & \textbf{39.2} & \underline{39.3} \\
\addlinespace[3pt]
\textcolor{red}{\amorgdn{}\amorheart{}} & \textbf{30.2} & \underline{31.8} & 65.0 & 56.0 & 24.6 & \textbf{51.3} & \textbf{23.8} & \underline{40.9} & \textbf{40.5} \\
\amor{}-symmetric-Gated DeltaNet & 27.9 & 31.7 & 65.0 & \textbf{56.9} & \textbf{25.1} & 50.5 & 20.0 & \textbf{42.1} & \underline{39.9} \\
\amor{}-ungated-Gated DeltaNet & \underline{28.6} & \underline{31.8} & \underline{65.8} & \underline{56.4} & \underline{24.7} & 49.6 & 20.0 & 39.4 & 39.5 \\
\textsc{Amor}-MLP-Gated DeltaNet & 27.4 & \textbf{31.9} & \textbf{66.0} & 55.9 & 24.6 & \underline{51.0} & \underline{20.2} & 39.3 & 39.5 \\
\bottomrule
\end{tabular}}
\end{table*}

\begin{table*}[t]
\centering
\footnotesize
\setlength{\tabcolsep}{3pt}
\caption{\textbf{Symmetric block normalization ablation: common-sense comparison.} 1.5B; in the format of Table~\ref{tab:common_sense}.}
\label{tab:common_sense_amor_h_1p5b}
\resizebox{\textwidth}{!}{%
\begin{tabular}{lccccccccccc}
\toprule
\textbf{Model} & \textbf{FW-Edu} & \textbf{LAMB} & \textbf{LAMB} & \textbf{HSwag} & \textbf{PIQA} & \textbf{ARC-E} & \textbf{ARC-C} & \textbf{WinoG} & \textbf{OBQA} & \textbf{TQA mc2} & \textbf{Avg} \\
 & PPL $\downarrow$ & PPL $\downarrow$ & acc $\uparrow$ & acc $\uparrow$ & acc $\uparrow$ & acc $\uparrow$ & acc $\uparrow$ & acc $\uparrow$ & acc $\uparrow$ & acc $\uparrow$ & acc $\uparrow$ \\
\midrule
\multicolumn{12}{@{}l}{\textbf{\small 1.5B}} \\
\textcolor{red}{\amormamba{}\amorheart{}} & 13.30 & 23.01 & 39.2 & 38.6 & 70.9 & 67.0 & 30.2 & 54.7 & 26.8 & 39.0 & 45.8 \\
\amor{}-symmetric-Mamba2 & 13.25 & 20.79 & 40.1 & 38.9 & 70.6 & 66.0 & 31.8 & 53.0 & 24.6 & 38.5 & 45.4 \\
\addlinespace[3pt]
\textcolor{red}{\amorgdn{}\amorheart{}} & 13.10 & 21.49 & 39.1 & 39.2 & 70.6 & 66.4 & 30.5 & 53.7 & 24.8 & 36.9 & 45.1 \\
\amor{}-symmetric-Gated DeltaNet & 13.09 & 22.64 & 38.2 & 39.1 & 71.4 & 67.4 & 32.3 & 55.4 & 24.4 & 35.5 & 45.5 \\
\bottomrule
\end{tabular}
}
\end{table*}

\begin{table*}[!t]
  \centering
  \footnotesize
  \setlength{\tabcolsep}{3pt}
  \caption{\textbf{Symmetric block normalization ablation: retrieval and S-NIAH comparison.} 1.5B; in the format of Table~\ref{tab:retrieval}.}
  \label{tab:retrieval_amor_h_1p5b}
  \resizebox{\textwidth}{!}{%
  \begin{tabular}{l ccccccccccccccc}
    \toprule
    \textbf{Model} & \textbf{SWDE}$\uparrow$ & \textbf{SQuAD}$\uparrow$ & \textbf{FDA}$\uparrow$ & \textbf{TQA}$\uparrow$ & \textbf{NQ}$\uparrow$ & \textbf{DROP}$\uparrow$ & \multicolumn{3}{c}{\textbf{S-NIAH-1}$\uparrow$} & \multicolumn{3}{c}{\textbf{S-NIAH-2}$\uparrow$} & \multicolumn{3}{c}{\textbf{S-NIAH-3}$\uparrow$} \\
    \cmidrule(lr){2-7}\cmidrule(lr){8-10}\cmidrule(lr){11-13}\cmidrule(lr){14-16}
    \textbf{Context length} & \multicolumn{6}{c}{2048} & 1024 & 2048 & 4096 & 1024 & 2048 & 4096 & 1024 & 2048 & 4096 \\
    \midrule
    \textcolor{red}{\amormamba{}\amorheart{}} & 33.8 & 36.0 & 21.2 & 40.4 & 10.2 & 21.1 & 100.0 & 95.8 & 44.4 & 100.0 & 99.6 & 47.2 & 70.8 & 83.6 & 25.6 \\
    \amor{}-symmetric-Mamba2 & 31.1 & 35.5 & 22.1 & 41.4 & \phantom{0}9.9 & 17.2 & 100.0 & 92.6 & 42.2 & 100.0 & 90.2 & 47.6 & 79.2 & 85.8 & 19.2 \\
    \addlinespace[3pt]
    \textcolor{red}{\amorgdn{}\amorheart{}} & 47.4 & 36.8 & 24.9 & 43.1 & 10.5 & 17.4 & 100.0 & 96.2 & 48.0 & 100.0 & 98.6 & 53.0 & 88.8 & 86.0 & 29.4 \\
    \amor{}-symmetric-Gated DeltaNet & 30.4 & 35.2 & \phantom{0}9.4 & 39.2 & 10.1 & 18.4 & 100.0 & 99.0 & 75.4 & 100.0 & 32.8 & 29.4 & 69.2 & 47.8 & 19.2 \\
    \bottomrule
  \end{tabular}}
\end{table*}

\begin{table*}[t]
\centering
\footnotesize
\setlength{\tabcolsep}{3pt}
\caption{\textbf{\amor{} block-count ablation: common-sense comparison of $K=3$ and $K=1$.} 180M, 440M and 1.5B; in the format of Table~\ref{tab:common_sense}.}
\label{tab:vs_classic}
\resizebox{\textwidth}{!}{%
\begin{tabular}{lccccccccc}
\toprule
\textbf{Variant} & \textbf{LAMB} & \textbf{HSwag} & \textbf{PIQA} & \textbf{ARC-E} & \textbf{ARC-C} & \textbf{WinoG} & \textbf{OBQA} & \textbf{TQA mc2} & \textbf{Avg} \\
& acc $\uparrow$ & acc $\uparrow$ & acc $\uparrow$ & acc $\uparrow$ & acc $\uparrow$ & acc $\uparrow$ & acc $\uparrow$ & acc $\uparrow$ & acc $\uparrow$ \\
\midrule
\multicolumn{10}{@{}l}{\textbf{\small 180M}} \\
\textcolor{red}{\amormamba{}\amorheart{} ($K{=}3$)} & 20.3 & 27.4 & 60.5 & 46.5 & 18.0 & 51.3 & 16.6 & 45.9 & 35.8 \\
\amormamba{} ($K{=}1$) & 19.7 & 27.5 & 60.7 & 47.6 & 18.7 & 51.9 & 18.4 & 44.4 & 36.1 \\
\addlinespace[3pt]
\textcolor{red}{\amorgdn{}\amorheart{} ($K{=}3$)} & 19.5 & 27.3 & 61.0 & 47.7 & 17.3 & 51.1 & 16.2 & 45.2 & 35.7 \\
\amorgdn{} ($K{=}1$) & 19.0 & 27.4 & 60.1 & 46.7 & 19.0 & 51.3 & 16.6 & 44.1 & 35.5 \\
\midrule
\multicolumn{10}{@{}l}{\textbf{\small 440M}} \\
\textcolor{red}{\amormamba{}\amorheart{} ($K{=}3$)} & 29.3 & 31.4 & 65.9 & 55.4 & 24.7 & 52.4 & 21.0 & 36.4 & 39.6 \\
\amormamba{} ($K{=}1$) & 29.2 & 31.6 & 65.6 & 56.2 & 23.9 & 50.4 & 21.6 & 39.5 & 39.7 \\
\addlinespace[3pt]
\textcolor{red}{\amorgdn{}\amorheart{} ($K{=}3$)} & 30.2 & 31.8 & 65.0 & 56.0 & 24.6 & 51.3 & 23.8 & 40.9 & 40.5 \\
\amorgdn{} ($K{=}1$) & 28.4 & 31.9 & 65.7 & 56.9 & 24.3 & 49.6 & 21.8 & 41.7 & 40.0 \\
\midrule
\multicolumn{10}{@{}l}{\textbf{\small 1.5B}} \\
\textcolor{red}{\amormamba{}\amorheart{} ($K{=}3$)} & 39.2 & 38.6 & 70.9 & 67.0 & 30.2 & 54.7 & 26.8 & 39.0 & 45.8 \\
\amormamba{} ($K{=}1$) & 36.7 & 38.9 & 70.2 & 67.6 & 30.6 & 53.4 & 23.8 & 34.6 & 44.5 \\
\addlinespace[3pt]
\textcolor{red}{\amorgdn{}\amorheart{} ($K{=}3$)} & 39.1 & 39.2 & 70.6 & 66.4 & 30.5 & 53.7 & 24.8 & 36.9 & 45.1 \\
\amorgdn{} ($K{=}1$) & 38.4 & 39.4 & 71.0 & 65.7 & 30.1 & 53.8 & 26.6 & 35.5 & 45.1 \\
\bottomrule
\end{tabular}}
\end{table*}

\FloatBarrier
\endgroup
\begin{table}[H]
\centering
\footnotesize
\setlength{\tabcolsep}{3pt}
\caption{\textbf{Attention-block placement ablation: common-sense comparison.} 440M; in the format of Table~\ref{tab:common_sense}.}
\label{tab:iclr-placement}
\resizebox{\textwidth}{!}{%
\begin{tabular}{lccccccccc}
\toprule
\textbf{Model} & \textbf{LAMB} & \textbf{HSwag} & \textbf{PIQA} & \textbf{ARC-E} & \textbf{ARC-C} & \textbf{WinoG} & \textbf{OBQA} & \textbf{TQA mc2} & \textbf{Avg} \\
& acc $\uparrow$ & acc $\uparrow$ & acc $\uparrow$ & acc $\uparrow$ & acc $\uparrow$ & acc $\uparrow$ & acc $\uparrow$ & acc $\uparrow$ & acc $\uparrow$ \\
\midrule
\textcolor{red}{\amormamba{}\amorheart{}} & 29.3 & 31.4 & 65.9 & 55.4 & 24.7 & 52.4 & 21.0 & 36.4 & 39.6 \\
\amor{}-serial-Mamba2 & 26.8 & 31.7 & 65.6 & 55.3 & 22.3 & 50.2 & 21.2 & 36.3 & 38.7 \\
\amor{}-fused-Mamba2 & 27.1 & 31.8 & 66.2 & 56.2 & 23.1 & 53.1 & 19.2 & 40.1 & 39.6 \\
\addlinespace[3pt]
\textcolor{red}{\amorgdn{}\amorheart{}} & 30.2 & 31.8 & 65.0 & 56.0 & 24.6 & 51.3 & 23.8 & 40.9 & 40.5 \\
\amor{}-serial-Gated DeltaNet & 27.7 & 31.8 & 65.7 & 55.8 & 24.0 & 49.5 & 22.8 & 41.9 & 39.9 \\
\amor{}-fused-Gated DeltaNet & 28.0 & 31.9 & 65.6 & 56.8 & 24.1 & 51.3 & 20.4 & 38.8 & 39.6 \\
\bottomrule
\end{tabular}}
\end{table}

\Needspace{5\baselineskip}
\section{Limitations}
\label{app:limitations}

We fix $\kappa=0.2$ across scales and backbones, motivated by an earlier small-budget 180M sweep under a fixed-offset rule. That sweep suggested firing on slightly fewer than half of positions, but does not establish a generalizable rule beyond the configurations studied here. The gate also operates on a progressively narrower entropy distribution: the \amormamba{} Block~0 median rises from 0.706 to 0.934 and its SD falls from 0.106 to 0.028 between 180M and 1.5B, with similar values for \amorgdn{} (Table~\ref{tab:gate_stats}). Whether entropy calibration improves routing in this regime remains open.

Entropy measures uncertainty in the predictive distribution, not correctness or recurrent-state capacity; a confident error can leave the gate closed. Runtime conclusions are specific to the measured kernels, hardware, batch sizes and workloads.

\Needspace{5\baselineskip}
\section{Model Weights}
\label{app:model-weights}

Model weights are available through the following links.

\begin{center}
{\small
\begin{tabular}{lll}
\toprule
Model & Scale & Link \\
\midrule
\amormamba{} & 180M & \url{https://huggingface.co/FlyinGodzilla/AMOR-Mamba2-180M} \\
\amorgdn{} & 180M & \url{https://huggingface.co/FlyinGodzilla/AMOR-GatedDeltaNet-180M} \\
\amormamba{} & 440M & \url{https://huggingface.co/FlyinGodzilla/AMOR-Mamba2-440M} \\
\amorgdn{} & 440M & \url{https://huggingface.co/FlyinGodzilla/AMOR-GatedDeltaNet-440M} \\
\amormamba{} & 1.5B & \url{https://huggingface.co/FlyinGodzilla/AMOR-Mamba2-1.5B} \\
\amorgdn{} & 1.5B & \url{https://huggingface.co/FlyinGodzilla/AMOR-GatedDeltaNet-1.5B} \\
\bottomrule
\end{tabular}
}
\end{center}

\end{document}